\PassOptionsToPackage{table}{xcolor}
\documentclass{article}

\usepackage{arxiv}
\usepackage[numbers,sort&compress]{natbib}
\usepackage{amsmath,amssymb}
\usepackage{booktabs}
\usepackage{array}
\usepackage{tabularx}
\usepackage{colortbl}
\usepackage{makecell}
\usepackage{threeparttable}
\usepackage{adjustbox}
\usepackage{graphicx}
\usepackage{wrapfig}
\usepackage{flafter}
\usepackage{float}
\usepackage{placeins}
\usepackage{microtype}
\usepackage{hyperref}
\usepackage{xurl}
\usepackage[utf8]{inputenc}
\usepackage[T1]{fontenc}
\usepackage[most]{tcolorbox}
\usepackage{listings}
\usepackage{enumitem}
\usepackage{tikz}
\usepackage{needspace}
\usepackage{eso-pic}
\usepackage{longtable}
\hypersetup{hidelinks}

\definecolor{EarthBlue}{HTML}{174A5B}
\definecolor{EarthTeal}{HTML}{2F746C}
\definecolor{EarthPale}{HTML}{EEF5F3}
\definecolor{EarthGray}{HTML}{F4F5F6}
\definecolor{EarthRule}{HTML}{BAC7C8}
\definecolor{EarthRed}{HTML}{9E3B33}
\colorlet{lastexamred}{EarthRed}
\colorlet{positionblue}{EarthTeal}

\makeatletter
\renewcommand{\section}{%
  \@startsection{section}{1}{\z@}%
    {-2.0ex \@plus -0.5ex \@minus -0.2ex}%
    {1.25ex \@plus 0.25ex \@minus 0.15ex}%
    {\large\sffamily\bfseries\color{EarthBlue}\raggedright}}
\renewcommand{\subsection}{%
  \@startsection{subsection}{2}{\z@}%
    {-1.65ex \@plus -0.4ex \@minus -0.2ex}%
    {0.68ex \@plus 0.15ex}%
    {\normalsize\sffamily\bfseries\color{EarthBlue}\raggedright}}
\renewcommand{\subsubsection}{%
  \@startsection{subsubsection}{3}{\z@}%
    {-1.35ex \@plus -0.35ex \@minus -0.15ex}%
    {0.45ex \@plus 0.12ex}%
    {\normalsize\sffamily\bfseries\color{EarthBlue}\raggedright}}
\renewcommand{\paragraph}{%
  \@startsection{paragraph}{4}{\z@}%
    {1.25ex \@plus 0.35ex \@minus 0.15ex}%
    {-0.82em}%
    {\normalsize\sffamily\bfseries\color{EarthBlue}}}
\makeatother

\fancypagestyle{earthversefirst}{%
  \fancyhf{}%
  \fancyhead[L]{%
    \includegraphics[height=18pt]{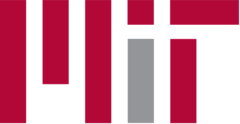}}%
  \fancyfoot[C]{\thepage}%
}
\fancypagestyle{earthversepaper}{%
  \fancyhf{}%
  \fancyhead[C]{\fontsize{6.8pt}{7.8pt}\selectfont\sffamily\bfseries\color{EarthBlue}%
    EarthVerse: Benchmarking Scientific Agents Across Dynamic Earth Systems and Natural Hazards}%
  \fancyfoot[C]{\thepage}%
}

\newcommand{\best}[1]{\textbf{\textcolor{EarthBlue}{#1}}}
\newcommand{\second}[1]{\textcolor{EarthTeal}{\underline{#1}}}
\newcommand{\gain}[1]{\textcolor{EarthTeal}{#1}}
\newcommand{\loss}[1]{\textcolor{EarthRed}{#1}}
\newcommand{\cmark}{\textcolor{EarthTeal}{\scalebox{1.35}{$\checkmark$}}}
\newcommand{\xmark}{\textcolor{EarthRule}{\scalebox{1.30}{$\times$}}}
\newcommand{\pmark}{%
  \tikz[baseline=-0.70ex]{%
    \draw[EarthBlue,line width=0.52pt] (0,0) circle (0.56ex);%
    \fill[EarthBlue] (0,-0.56ex) arc[start angle=-90,end angle=90,radius=0.56ex] -- cycle;}}
\newcommand{\projectlink}[3]{%
  \href{#1}{%
    \raisebox{-0.34ex}{\includegraphics[height=1.75ex]{#2}}%
    \hspace{0.42em}\textcolor{EarthBlue}{%
      \fontfamily{ppl}\selectfont\bfseries\fontsize{7.8pt}{8.8pt}\selectfont #3}}}

\newcommand{\tblhead}{\rowcolor{EarthPale}}
\newcommand{\keyrow}{\rowcolor{EarthPale}}
\newcommand{\keycell}[1]{\cellcolor{EarthPale}#1}
\newcommand{\tblpanel}[2]{\addlinespace[2.5pt]\multicolumn{#1}{@{}l}{\textit{\textcolor{EarthBlue}{#2}}}\\[1pt]}
\newcommand{\tabfontregular}{\fontsize{8pt}{9.2pt}\selectfont}
\newcommand{\tabfontcompact}{\fontsize{7.15pt}{8.25pt}\selectfont}
\newcommand{\tabfontdense}{\fontsize{6.7pt}{7.7pt}\selectfont}
\newtcolorbox{findings}{%
  enhanced, sharp corners,
  colback=lastexamred!4, colframe=lastexamred!70!black,
  boxrule=0pt, leftrule=3pt,
  left=7pt, right=7pt, top=5pt, bottom=5pt,
  before skip=6pt, after skip=8pt,
  fontupper={\fontsize{9.1pt}{11.0pt}\selectfont\sffamily\color{EarthRed!28!black}}
}
\newtcolorbox{positionbox}{%
  enhanced, sharp corners,
  colback=positionblue!5, colframe=positionblue!78!black,
  boxrule=0pt, leftrule=3pt,
  left=7pt, right=7pt, top=5pt, bottom=5pt,
  before skip=7pt, after skip=9pt,
  fontupper={\fontsize{9.1pt}{11.0pt}\selectfont\sffamily\color{EarthBlue!58!black}}
}
\newcommand{\findingbox}[3]{%
  \phantomsection\label{#1}%
  \begin{findings}
  {\sffamily\bfseries\textcolor{lastexamred!88!black}{Finding. #2.}} #3
  \end{findings}}
\newcommand{\positionstatement}[1]{%
  \begin{positionbox}
  {\sffamily\bfseries\textcolor{positionblue!88!black}{Position.}} #1
  \end{positionbox}}
\newtcolorbox{sourcepanel}[1]{
  enhanced jigsaw,breakable,parbox=false,
  boxrule=0.65pt,arc=1.2mm,
  pad at break*=1mm,lines before break=4,
  colback=white,colframe=CaseBlue!78!black,
  colbacktitle=CaseDeep,coltitle=white,
  fonttitle=\sffamily\small\bfseries,
  title={\textsf{PACKAGE SOURCES}\quad #1},
  title after break={\textsf{PACKAGE SOURCES}\quad #1\hfill\textit{continued}},
  left=1.8mm,right=1.8mm,top=1.2mm,bottom=1.2mm,
  before skip=5pt,after skip=6pt}
\newcommand{\sourcegroup}[2]{%
  \par\vspace{1.1mm}\noindent
  {\sffamily\small\bfseries\textcolor{#1!80!black}{#2}}\par
  \vspace{0.35mm}\noindent\textcolor{#1!35}{\rule{\linewidth}{0.45pt}}\par
  \vspace{0.5mm}%
  \noindent\begin{tabularx}{\linewidth}{@{}>{\raggedright\arraybackslash}p{0.29\linewidth}>{\raggedright\arraybackslash}p{0.27\linewidth}Y@{}}
  \scriptsize\bfseries\textcolor{CaseMuted}{Provider or product} &
  \scriptsize\bfseries\textcolor{CaseMuted}{Access form} &
  \scriptsize\bfseries\textcolor{CaseMuted}{Role in a package}\\[-1pt]
  \end{tabularx}\vspace{0.3mm}}
\newcommand{\sourceentry}[4][CaseBlue]{%
  \noindent\begin{tabularx}{\linewidth}{@{}>{\raggedright\arraybackslash}p{0.29\linewidth}>{\raggedright\arraybackslash}p{0.27\linewidth}Y@{}}
  \scriptsize\bfseries\textcolor{#1!78!black}{#2} &
  \scriptsize\textcolor{CasePurple!78!black}{#3} &
  \scriptsize #4\\
  \end{tabularx}\vspace{0.15mm}}

\newcolumntype{Y}{>{\raggedright\arraybackslash}X}
\newcolumntype{C}{>{\centering\arraybackslash}X}
\newcolumntype{R}{>{\raggedleft\arraybackslash}X}
\newcommand{\mainfigure}[2][]{%
  \includegraphics[#1]{figures/main_text/#2}}
\newcommand{\figurepanel}[3]{%
  \begin{minipage}[t]{#1}
  \centering
  \mainfigure[width=\linewidth]{#2}\par\vspace{1pt}
  {\scriptsize #3}
  \end{minipage}}

\definecolor{CaseBlue}{HTML}{285F74}
\definecolor{CaseDeep}{HTML}{173B49}
\definecolor{CaseTeal}{HTML}{2F7D73}
\definecolor{CaseGreen}{HTML}{4E7B5A}
\definecolor{CaseAmber}{HTML}{B17620}
\definecolor{CaseRose}{HTML}{A64B4B}
\definecolor{CasePurple}{HTML}{6C5A8D}
\definecolor{CaseInk}{HTML}{26383D}
\definecolor{CaseMuted}{HTML}{65757B}
\definecolor{CaseLine}{HTML}{B9C9CC}
\definecolor{CasePaleBlue}{HTML}{F1F7F9}
\definecolor{CasePaleTeal}{HTML}{F1F8F6}
\definecolor{CasePaleAmber}{HTML}{FCF7EC}
\definecolor{CasePaleRose}{HTML}{FCF2F2}
\definecolor{CasePalePurple}{HTML}{F6F3FA}
\definecolor{CaseCode}{HTML}{F7F9FA}
\newcommand{\appendixguideentry}[6]{%
  \noindent
  \begin{tabularx}{\linewidth}{@{}>{\centering\arraybackslash}m{0.058\linewidth}@{\hspace{0.9em}}Y@{\hspace{0.7em}}r@{}}
  \tikz[baseline=(letter.base)]{\node[
    rounded corners=1.2pt,
    draw=EarthBlue!62!black,
    fill=EarthBlue!9,
    minimum width=2.05em,
    minimum height=2.05em,
    inner sep=0pt] (letter)
    {\sffamily\bfseries\large\textcolor{EarthBlue!82!black}{#1}};} &
  \hyperref[#2]{\sffamily\bfseries\textcolor{CaseInk}{Appendix #1\enspace #3}} &
  \hyperref[#2]{\sffamily\bfseries\textcolor{EarthBlue!82!black}{\pageref*{#2}}}\\[1.5pt]
  & \small #4 & \\
  & \scriptsize\textcolor{CaseMuted}{#5} &
  \end{tabularx}\par
  \vspace{3.5pt}\noindent\textcolor{CaseLine}{\rule{\linewidth}{0.45pt}}\par\vspace{6pt}}

\newcommand{\scorechip}[1]{%
  \tikz[baseline=(n.base)]\node[fill=CaseGreen!13,draw=CaseGreen!65,
  rounded corners=2pt,inner xsep=4pt,inner ysep=1.6pt,
  font=\sffamily\scriptsize\bfseries,text=CaseGreen!55!black](n){Core #1};}
\newcommand{\failchip}[1]{%
  \tikz[baseline=(n.base)]\node[fill=CaseRose!12,draw=CaseRose!70,
  rounded corners=2pt,inner xsep=4pt,inner ysep=1.6pt,
  font=\sffamily\scriptsize\bfseries,text=CaseRose!75!black](n){#1};}
\newcommand{\systemchip}[2]{%
  \tikz[baseline=(n.base)]\node[fill=#1!11,draw=#1!58!black,
  rounded corners=2pt,inner xsep=4pt,inner ysep=1.4pt,
  font=\sffamily\scriptsize\bfseries,text=#1!70!black](n){#2};}
\newcommand{\familylabel}[2]{%
  \textcolor{#1!78!black}{\sffamily\scriptsize\bfseries #2}}
\newcommand{\toolchip}[2][CasePurple]{%
  \begingroup\setlength{\fboxsep}{1.4pt}%
  \colorbox{#1!10}{\strut\textcolor{#1!78!black}{%
  \texttt{\scriptsize\detokenize{#2}}}}\endgroup}
\newcommand{\stepbadge}[1]{%
  \tikz[baseline=(n.base)]\node[circle,fill=CaseBlue,text=white,
  minimum size=4.7mm,inner sep=0pt,font=\sffamily\tiny\bfseries](n){#1};}

\lstdefinestyle{earthlisting}{
  basicstyle=\ttfamily\scriptsize,
  stringstyle=\color{CaseDeep},
  breaklines=true,
  breakatwhitespace=false,
  columns=fullflexible,
  keepspaces=true,
  showstringspaces=false,
  upquote=true,
  aboveskip=0pt,
  belowskip=0pt,
  xleftmargin=1pt,
  frame=none,
  tabsize=2
}

\lstdefinestyle{earthjsonlisting}{
  style=earthlisting,
  basicstyle=\ttfamily\fontsize{6.8pt}{8.1pt}\selectfont,
  stringstyle=\color{CaseBlue!82!black},
  keywordstyle=\color{CaseTeal!78!black}\bfseries,
  morekeywords={true,false,null},
  literate=*
    {[}{{{\color{CasePurple}[}}}1
    {]}{{{\color{CasePurple}]}}}1
    {:}{{{\color{CaseMuted}:}}}1
    {,}{{{\color{CaseMuted},}}}1
    {0}{{{\color{CaseAmber}0}}}1
    {1}{{{\color{CaseAmber}1}}}1
    {2}{{{\color{CaseAmber}2}}}1
    {3}{{{\color{CaseAmber}3}}}1
    {4}{{{\color{CaseAmber}4}}}1
    {5}{{{\color{CaseAmber}5}}}1
    {6}{{{\color{CaseAmber}6}}}1
    {7}{{{\color{CaseAmber}7}}}1
    {8}{{{\color{CaseAmber}8}}}1
    {9}{{{\color{CaseAmber}9}}}1
}

\lstdefinestyle{earthpromptlisting}{
  style=earthlisting,
  basicstyle=\ttfamily\fontsize{7.2pt}{8.5pt}\selectfont,
  literate=
    {SYSTEM}{{{\color{CaseBlue}\bfseries SYSTEM}}}6
    {ROLE}{{{\color{CaseBlue}\bfseries ROLE}}}4
    {CONTEXT}{{{\color{CaseTeal}\bfseries CONTEXT}}}7
    {QUESTION}{{{\color{CaseAmber}\bfseries QUESTION}}}8
    {EVIDENCE}{{{\color{CaseAmber}\bfseries EVIDENCE}}}8
    {BOUNDARY}{{{\color{CaseAmber}\bfseries BOUNDARY}}}8
    {TOOLS}{{{\color{CasePurple}\bfseries TOOLS}}}5
    {INPUTS}{{{\color{CasePurple}\bfseries INPUTS}}}6
    {INPUT}{{{\color{CasePurple}\bfseries INPUT}}}5
    {RUNTIME}{{{\color{CasePurple}\bfseries RUNTIME}}}7
    {PAYLOAD}{{{\color{CasePurple}\bfseries PAYLOAD}}}7
    {TASK}{{{\color{CaseRose}\bfseries TASK}}}4
    {PROCEDURE}{{{\color{CaseRose}\bfseries PROCEDURE}}}9
    {INSTRUCTIONS}{{{\color{CaseRose}\bfseries INSTRUCTIONS}}}{12}
    {HARD}{{{\color{CaseRose}\bfseries HARD}}}4
    {RULES}{{{\color{CaseRose}\bfseries RULES}}}5
    {CONSTRAINTS}{{{\color{CaseRose}\bfseries CONSTRAINTS}}}{11}
    {FORBIDDEN}{{{\color{CaseRose}\bfseries FORBIDDEN}}}9
    {REQUIRED}{{{\color{CaseRose}\bfseries REQUIRED}}}8
    {COVERAGE}{{{\color{CaseRose}\bfseries COVERAGE}}}8
    {CALIBRATION}{{{\color{CaseTeal}\bfseries CALIBRATION}}}{11}
    {ACCEPTANCE}{{{\color{CaseTeal}\bfseries ACCEPTANCE}}}{10}
    {CHECKS}{{{\color{CaseTeal}\bfseries CHECKS}}}6
    {OUTPUT}{{{\color{CaseTeal}\bfseries OUTPUT}}}6
    {FORMAT}{{{\color{CaseTeal}\bfseries FORMAT}}}6
    {FIELDS}{{{\color{CaseTeal}\bfseries FIELDS}}}6
    {CRITIQUE}{{{\color{CaseAmber}\bfseries CRITIQUE}}}8
    {PRIORITIES}{{{\color{CaseAmber}\bfseries PRIORITIES}}}{10}
    {DECISION}{{{\color{CaseAmber}\bfseries DECISION}}}8
}

\tcbset{caseflow/.style={
  enhanced jigsaw,breakable,parbox=false,
  boxrule=0.65pt,arc=1.2mm,
  left=1.4mm,right=1.4mm,top=1.0mm,bottom=1.0mm,
  before skip=4pt,after skip=5pt,
  pad at break*=1mm,lines before break=4
}}

\newcommand{\promptdivider}[3]{%
  \par\vspace{0.9mm}\noindent
  \colorbox{#1!12}{\strut\textcolor{#1!78!black}{%
  \sffamily\scriptsize\bfseries #2}}\enspace
  \textcolor{#1!78!black}{\bfseries #3}\par
  \vspace{0.35mm}\noindent\textcolor{#1!28}{\rule{\linewidth}{0.35pt}}%
  \par\vspace{0.6mm}}

\newtcblisting{promptbox}[2][]{
  caseflow,listing only,listing engine=listings,
  colback=CasePaleBlue,colframe=CaseBlue,
  colbacktitle=CaseBlue,coltitle=white,
  fonttitle=\bfseries\small,
  title={\textsf{PROMPT}\quad #2},
  title after break={\textsf{PROMPT}\quad #2\hfill\textit{continued}},
  before skip=5pt,after skip=6pt,
  listing options={style=earthpromptlisting,escapeinside={(*@}{@*)}},#1
}

\newtcblisting{questionbox}[2][]{
  caseflow,listing only,listing engine=listings,
  colback=CasePaleAmber,colframe=CaseAmber,
  colbacktitle=CaseAmber,coltitle=white,
  fonttitle=\bfseries\small,
  title={\textsf{QUESTION}\quad #2},
  title after break={\textsf{QUESTION}\quad #2\hfill\textit{continued}},
  before skip=5pt,after skip=6pt,
  listing options={style=earthlisting,basicstyle=\rmfamily\fontsize{8.0pt}{9.6pt}\selectfont,
    escapeinside={(*@}{@*)}},#1
}

\newtcblisting{resultlisting}[2][]{
  caseflow,listing only,listing engine=listings,
  colback=CasePaleTeal,colframe=CaseTeal,
  colbacktitle=CaseTeal,coltitle=white,
  fonttitle=\bfseries\small,
  title={\textsf{RESULT}\quad #2},
  title after break={\textsf{RESULT}\quad #2\hfill\textit{continued}},
  listing options={style=earthjsonlisting},#1
}

\newtcolorbox{casebox}[2][]{
  caseflow,colback=white,colframe=CaseDeep,
  colbacktitle=CaseDeep,coltitle=white,fonttitle=\bfseries,
  title={\textsf{CASE}\quad #2},boxrule=0.8pt,arc=1.8mm,
  title after break={\textsf{CASE}\quad #2\hfill\textit{continued}},#1
}

\newtcolorbox{tracebox}[2][]{
  caseflow,colback=CaseCode,colframe=CaseTeal,
  colbacktitle=CaseTeal,coltitle=white,fonttitle=\bfseries\small,
  fontupper=\fontsize{8.6pt}{10.1pt}\selectfont,
  title={\textsf{TRACE}\quad #2},boxrule=0.7pt,arc=1.6mm,
  title after break={\textsf{TRACE}\quad #2\hfill\textit{continued}},#1
}

\newtcolorbox{notebox}[2][]{
  enhanced,breakable,colback=CasePalePurple,colframe=CasePurple,
  colbacktitle=CasePurple,coltitle=white,fonttitle=\bfseries\small,
  title={\textsf{NOTE}\quad #2},boxrule=0.6pt,arc=1.4mm,
  left=2mm,right=2mm,top=1.5mm,bottom=1.5mm,
  before skip=4pt,after skip=5pt,#1
}

\newtcolorbox{failurebox}[2][]{
  caseflow,colback=CasePaleRose,colframe=CaseRose,
  colbacktitle=CaseRose,coltitle=white,fonttitle=\bfseries,
  title={\textsf{FAILURE}\quad #2},boxrule=0.8pt,arc=1.8mm,
  title after break={\textsf{FAILURE}\quad #2\hfill\textit{continued}},#1
}

\newtcolorbox{casecard}[2][]{
  enhanced,breakable=false,colback=#2!2,colframe=#2!68!black,
  boxrule=0.65pt,arc=1.2mm,left=1.5mm,right=1.5mm,
  top=1mm,bottom=1mm,before skip=1.2pt,after skip=1.2pt,
  borderline west={1.8pt}{0pt}{#2!78!black},#1
}

\newtcolorbox{evidencebox}[3][]{
  caseflow,colback=#2!4,colframe=#2!78!black,
  colbacktitle=#2!78!black,coltitle=white,
  fonttitle=\bfseries\small,
  title={\textsf{#3}},
  title after break={\textsf{#3}\hfill\textit{continued}},#1
}

\newcommand{\summarycard}[8]{%
  \overviewrow{#1}{#2}{#3}{#4}{#5}{#6}{#7\enspace
  \textcolor{#1!70!black}{\sffamily\scriptsize\bfseries TRACE}\enspace #8}}

\newcommand{\promptkind}[2]{%
  \colorbox{#1!12}{\strut\textcolor{#1!78!black}{%
  \sffamily\scriptsize\bfseries #2}}}

\newcommand{\metric}[2]{%
  \mbox{\textcolor{CaseMuted}{\sffamily\scriptsize\bfseries #1}\nobreakspace
  \textcolor{CaseDeep}{\bfseries #2}}}

\newcommand{\verdict}[2][CaseGreen]{%
  \tcbox[on line,enhanced,boxrule=0.3pt,arc=0.8mm,boxsep=0pt,
  colback=#1!8,colframe=#1!40,
  left=0.7mm,right=0.7mm,top=0.28mm,bottom=0.28mm]{%
  \textcolor{#1!78!black}{\sffamily\scriptsize\bfseries #2}}}

\newcommand{\caseheading}[2]{%
  \subsection{\texorpdfstring{\textcolor{#1!78!black}{#2}}{#2}}}

\newcommand{\tracestep}[5][CasePurple]{%
  \noindent\begin{tabularx}{\linewidth}{@{}p{0.055\linewidth}p{0.235\linewidth}Y@{}}
  \stepbadge{#2} & \toolchip[#1]{#3} &
  \textcolor{#1!78!black}{\bfseries #4}\enspace #5\\
  \end{tabularx}\vspace{0.55mm}}

\newtcolorbox{promptgroup}[3][]{
  caseflow,colback=#2!2,colframe=#2!68!black,
  colbacktitle=#2!78!black,coltitle=white,
  fonttitle=\bfseries\small,
  title={\textsf{PROMPT SUITE}\quad #3},
  title after break={\textsf{PROMPT SUITE}\quad #3\hfill\textit{continued}},
  boxrule=0.72pt,arc=1.5mm,left=2mm,right=2mm,top=1.5mm,bottom=1.5mm,#1
}

\newtcolorbox{overviewbox}[1][]{
  caseflow,colback=white,colframe=CaseDeep!62,
  colbacktitle=CaseDeep,coltitle=white,
  fonttitle=\bfseries\small,
  title={\textsf{REPRESENTATIVE CASE SET}},
  title after break={\textsf{REPRESENTATIVE CASE SET}\hfill\textit{continued}},
  boxrule=0.72pt,arc=1.5mm,left=2mm,right=2mm,top=1.4mm,bottom=1.4mm,
  before skip=5pt,after skip=7pt,#1
}

\newtcolorbox{casepanel}[3][]{
  caseflow,colback=#2!1.5,colframe=#2!70!black,
  colbacktitle=#2!78!black,coltitle=white,
  fonttitle=\bfseries\small,
  fontupper=\fontsize{8.6pt}{10.1pt}\selectfont,
  title={\textsf{CASE}\quad #3},
  title after break={\textsf{CASE}\quad #3\hfill\textit{continued}},
  boxrule=0.72pt,arc=1.5mm,left=2mm,right=2mm,top=1.4mm,bottom=1.4mm,
  before skip=5pt,after skip=6pt,#1
}

\newcommand{\panelhead}[2]{%
  \Needspace{3\baselineskip}%
  \par\vspace{1.2mm}\noindent
  \textcolor{#1!78!black}{\sffamily\scriptsize\bfseries\MakeUppercase{#2}}\par
  \vspace{0.35mm}\noindent\textcolor{#1!30}{\rule{\linewidth}{0.35pt}}\par\vspace{0.7mm}}

\newcommand{\questionrole}[2]{%
  \tcbox[on line,enhanced,boxrule=0pt,arc=0.65mm,boxsep=0pt,
  colback=#1!11,colframe=#1!11,left=0.8mm,right=0.8mm,
  top=0.32mm,bottom=0.32mm]{%
  \textcolor{#1!82!black}{\sffamily\scriptsize\bfseries #2}}}

\newcommand{\questionvalue}[2]{%
  \textcolor{#1!82!black}{\bfseries #2}}

\newcommand{\questionitem}[3]{%
  \noindent\questionrole{#1}{#2}\enspace #3\par\vspace{0.65mm}}

\newcommand{\overviewrow}[7]{%
  \Needspace{5\baselineskip}%
  \noindent\begin{tabularx}{\linewidth}{@{}p{2.8mm}@{\hspace{1.8mm}}Y@{}}
  \tikz[baseline=-0.55ex]{\fill[#1!76!black] (0,0) rectangle (2.6mm,2.6mm);} &
  \familylabel{#1}{#2}\quad\systemchip{#1}{#3}\hfill
  \texttt{#4}\quad #5\\[-0.3mm]
  & \textcolor{#1!78!black}{\bfseries #6}\enspace #7\\
  \end{tabularx}\vspace{0.9mm}}

\renewcommand{\headeright}{}
\renewcommand{\undertitle}{}
\renewcommand{\shorttitle}{EarthVerse}
\date{}
\newtcolorbox{othervabstractbox}{%
  enhanced,sharp corners=downhill,arc=2mm,
  colback=EarthPale,colframe=EarthPale,boxrule=0pt,
  left=8mm,right=8mm,top=2mm,bottom=2mm,
  before skip=2pt,after skip=2pt,fontupper=\footnotesize}

\makeatletter
\renewcommand{\@maketitle}{%
  \vbox{%
    \hsize\textwidth
    \linewidth\hsize
    \vskip -0.04in
    \noindent\includegraphics[width=0.31\textwidth]{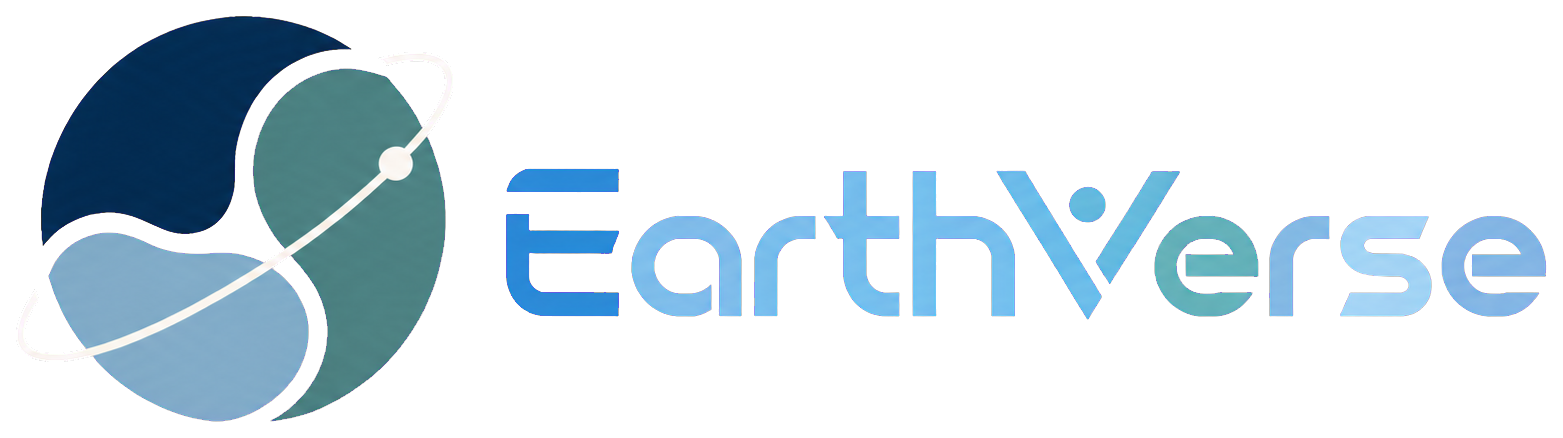}\par
    \vskip 0.035in
    \hrule height 2\p@
    \vskip 0.13in
    \centering
    {\fontsize{15.4pt}{18.2pt}\selectfont\scshape \@title\par}
    \vskip 0.13in
    \hrule height 2\p@
    \vskip 0.12in
    \textsc{\undertitle}\\[-1pt]
    \begin{tabular}[t]{c}\bfseries\rule{\z@}{13\p@}\@author\end{tabular}%
    \vskip 0.07in
  }%
}
\makeatother

\title{%
EarthVerse: Benchmarking Scientific Agents Across\\Dynamic Earth Systems and Natural Hazards}

\author{%
\resizebox{0.94\textwidth}{!}{\bfseries Zhiqing Cui$^{1}$\hspace{0.72em} Xinxiang Yin$^{2}$\hspace{0.72em} Yihong Tang$^{3}$\hspace{0.72em} Xinglang Zhang$^{4}$\hspace{0.72em} Yuanzhe Hu$^{5}$\hspace{0.72em} Siru Zhong$^{4}$\hspace{0.72em} Weidong Tang$^{6}$}\\[-0.6pt]
\resizebox{0.94\textwidth}{!}{\bfseries Yuxuan Liang$^{4}$\hspace{0.72em} Weijia Li$^{7}$\hspace{0.72em} Ming Jin$^{8}$\hspace{0.72em} Shirui Pan$^{8}$\hspace{0.72em} Yuhao Kang$^{9}$\hspace{0.72em} Dingyi Zhuang$^{10,\dagger}$\hspace{0.72em} Jinhua Zhao$^{10}$}\\[1.2pt]
\resizebox{0.94\textwidth}{!}{\normalfont $^{1}$NUIST\hspace{0.72em}$^{2}$HKU\hspace{0.72em}$^{3}$McGill\hspace{0.72em}$^{4}$HKUST(GZ)\hspace{0.72em}$^{5}$Georgia Tech\hspace{0.72em}$^{6}$NUS\hspace{0.72em}$^{7}$Tsinghua\hspace{0.72em}$^{8}$Griffith\hspace{0.72em}$^{9}$UT Austin\hspace{0.72em}$^{10}$MIT}\\[1pt]
{\makebox[0.94\textwidth][c]{\scriptsize\normalfont
\texttt{zhiqing@nuist.edu.cn}\hspace{1.5em}\texttt{dingyi@mit.edu}\hspace{1.5em}$^{\dagger}$Corresponding author}}\\[2.5pt]
{\makebox[0.94\textwidth][c]{\normalfont
\projectlink{https://huggingface.co/datasets/miracle10/EarthVerse}{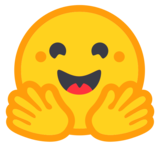}{Hugging Face}%
\hspace{2.1em}
\projectlink{https://github.com/CuiZHIQ/Earth-Verse}{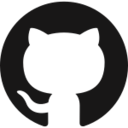}{GitHub}%
\hspace{2.1em}
\projectlink{https://cuizhiq.github.io/EarthVerse/}{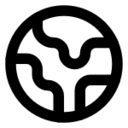}{Website}}}}

\begin{document}
\raggedbottom
\vspace*{-0.58in}
\maketitle

\begin{center}
\includegraphics[width=0.98\linewidth,height=0.82in,keepaspectratio,
  ]{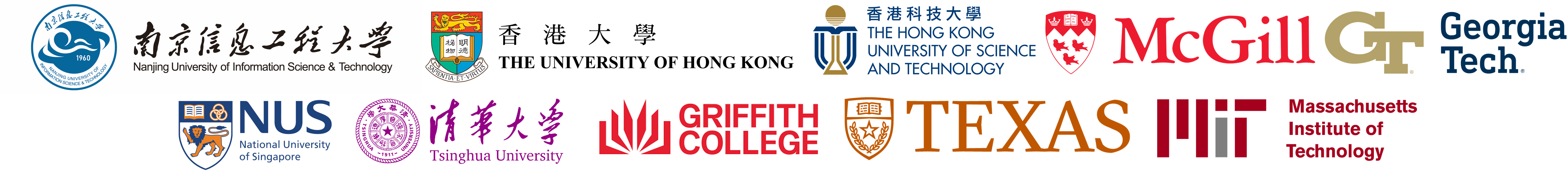}
\end{center}
\vspace{-0.50em}

\begin{othervabstractbox}
\begin{abstract}
Earth-system analysis reconstructs changing physical processes from observations that differ in source, scale, timing, and modality. Natural hazards make this work consequential because incomplete evidence can change estimates of severity, exposure, and mechanism. We introduce EarthVerse, a benchmark that evaluates scientific agents through package-scoped investigations. Its 405 reproducible tasks are grounded in 199 documented events and 19 hazard families. Agents inspect heterogeneous event packages, choose compatible evidence, execute transparent calculations, reconcile source differences, and preserve provenance in the final answer. We provide executable ground truth that decomposes each task into fine-grained answer units, together with task-specific rubrics that assess the supporting research process while allowing multiple valid paths. We evaluate 25 model and agent systems under a controlled tool-using protocol, then use controlled studies to locate failures in evidence access, tool selection, memory, reasoning, interaction, and scientific execution. Across systems, the best mean answer-unit accuracy is 84.65\%, while the highest Strict@95 is only 34.81\%. The gap shows that current agents often complete individual steps without maintaining a consistent chain across evidence, scales, units, calculations, and physical interpretation. EarthVerse provides a reproducible basis for measuring end-to-end scientific reliability in dynamic Earth systems.
\end{abstract}
\end{othervabstractbox}
\enlargethispage{1.1in}

\begin{figure}[H]
\centering
\vspace{-1.3em}
\makebox[\linewidth][c]{%
\begin{minipage}{0.90\linewidth}
\centering
\figurepanel{0.38\linewidth}{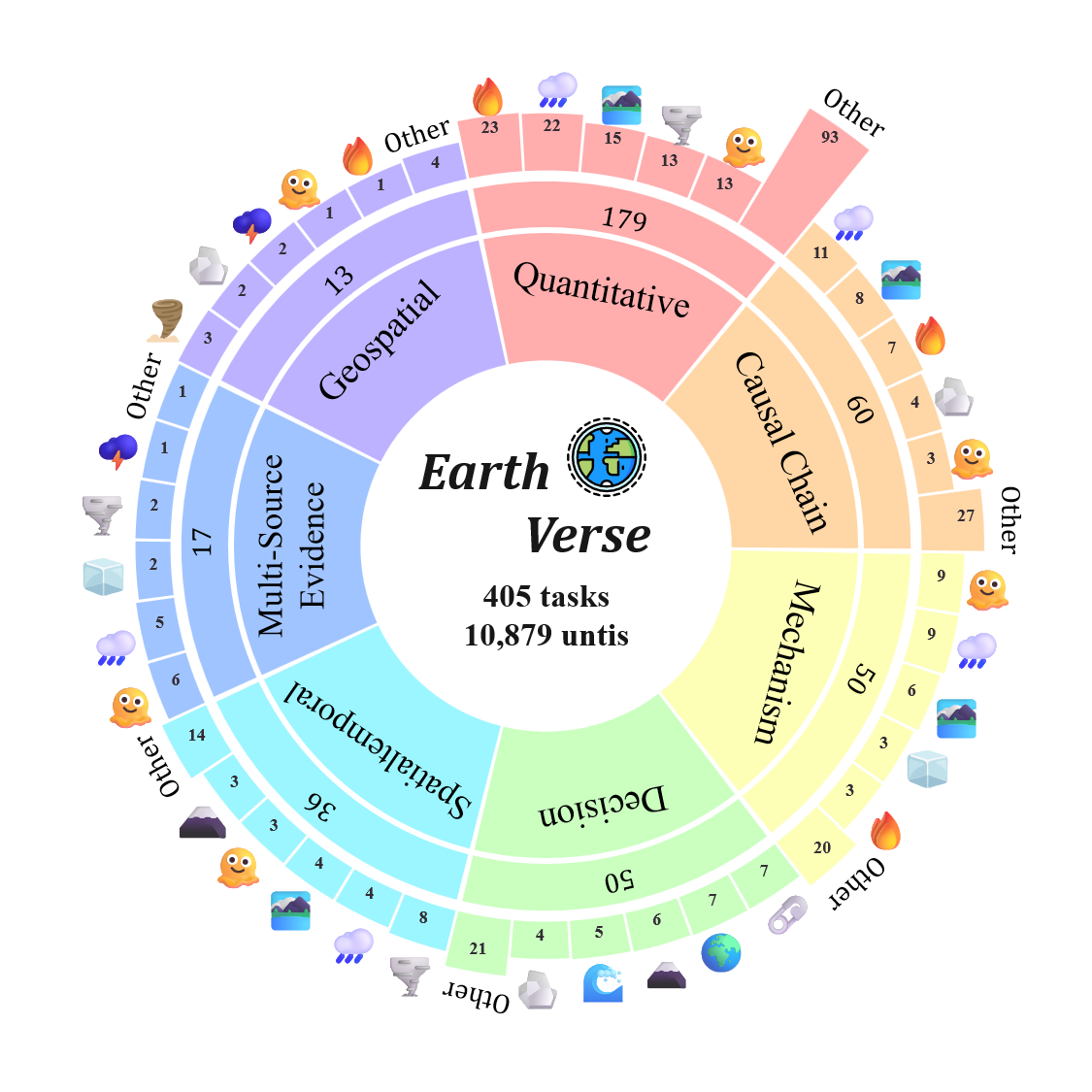}{\textbf{(a)} Scientific coverage.}%
\hfill
\figurepanel{0.60\linewidth}{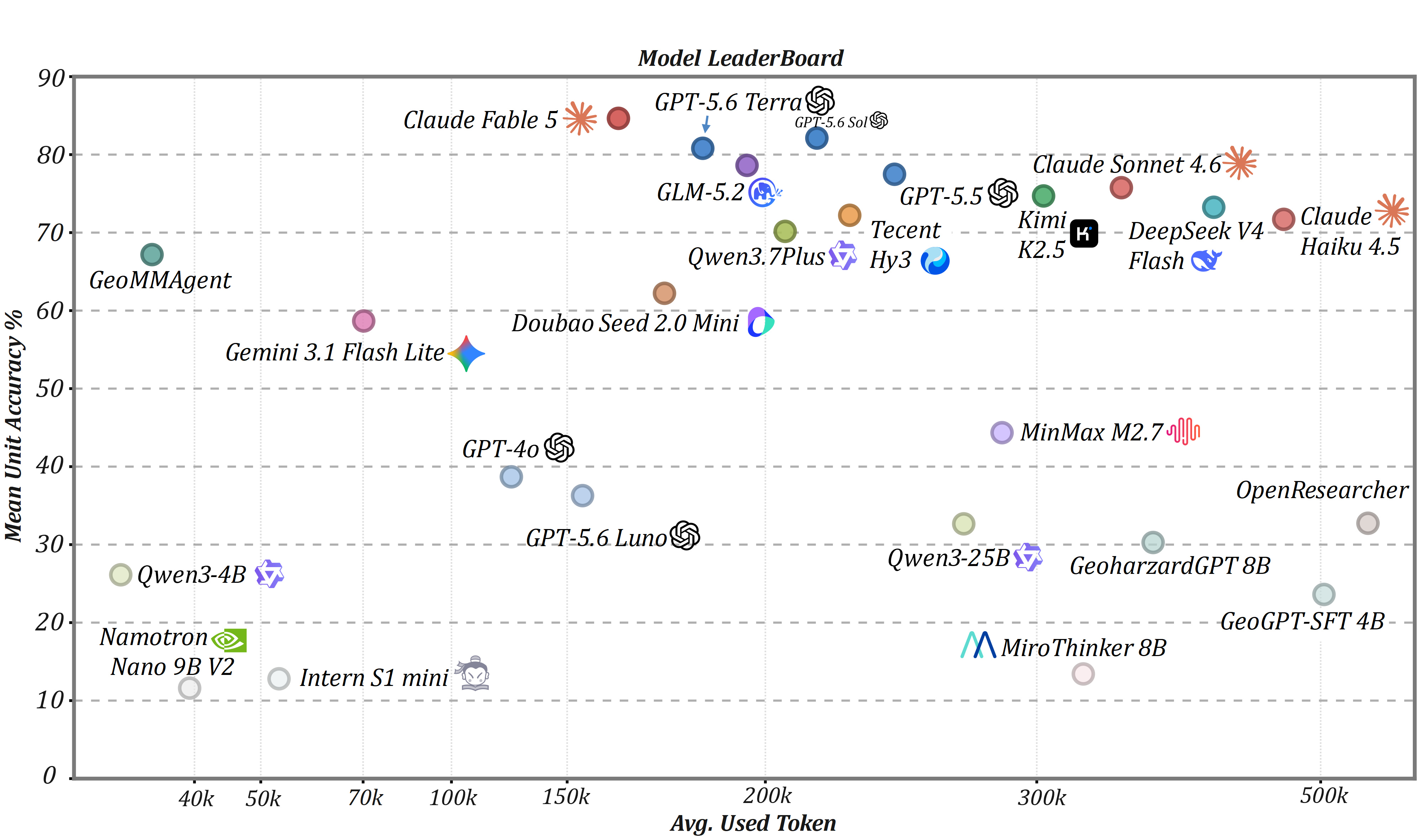}{\textbf{(b)} Leaderboard landscape.}%
\end{minipage}}
\vspace{-0.35em}
\caption{\textbf{Coverage and model landscape.} Task and answer-unit coverage across hazards and capabilities (left); mean answer-unit accuracy and token use across 25 systems (right).}
\label{fig:coverage-leaderboard}
\vspace{-0.7em}
\end{figure}
\clearpage
\setlength{\footskip}{16pt}
\pagestyle{earthversepaper}

\section{Introduction}
\label{sec:position}

Earth-system science asks how the atmosphere, land, ocean, ecosystems, and human activity shape one another. These processes govern water, food, health, infrastructure, and climate risk. Natural hazards make the question immediate: weather, climate, and water extremes have caused extensive mortality and economic loss, while compound events can propagate across sectors and regions~\citep{wmo2023atlas,wmo2025climate,ipcc2022impacts,zscheischler2018compound}. Timely analysis matters, but so does scientific traceability. A plausible explanation is not enough when it may inform disaster assessment or response.

Natural-hazard analysis applies Earth-system reasoning to decisions about event severity, exposure, physical mechanism, and response. It requires researchers to reconstruct an evolving process from observations collected for different purposes and at different scales.

The difficulty begins with observation. No single instrument records an evolving Earth system in full. Agency reports, stations, satellites, reanalyses, maps, and impact databases measure different variables over different footprints and time windows. Combining them requires scientific judgment before calculation: the analyst must determine which records describe the same process, which scale is appropriate, and what uncertainty remains. Work on machine learning for Earth systems and model--data integration has made these records easier to use, but has not removed the need to reconcile them~\citep{reichstein2019deep,markonis2021crossscale,gettelman2022modeldata,cui2026augur}.

Foundation models and tool-using agents now make a broader form of analysis possible. Systems such as Google Earth AI combine imagery, population, and environmental models through a reasoning agent, while scientific agents can inspect files, execute code, and revise a result after new observations arrive~\citep{bell2025earthai,yao2022react,chen2025scienceagentbench,hu2025survey,talemi2026agentic}. This changes the evaluation target. A system has to decide what evidence to seek, bind each value to the correct source and scale, and let tool feedback alter its account when the evidence disagrees.

Most Earth-science benchmarks begin after that decision has been made. They test knowledge, figure interpretation, remote sensing, or geospatial reasoning over a supplied passage, image, or named asset~\citep{zhang2025large,xu2025earthse,zhao2025msearth}. Newer multimodal and agent benchmarks broaden the interface, but generally continue to declare the observation or data layer in advance~\citep{xiao2026geommbench,li2025atmossci}. Frontier systems already approach or exceed published expert references on some of these formats. Such results do not show whether the same system can construct a compatible evidence base for an open question, carry calculations across several sources, and preserve support for every part of its conclusion.

\positionstatement{Earth-science analysis requires systems to maintain the correspondence between heterogeneous observations and scientific claims. Satellite products, station records, reanalyses, and event reports can support a shared physical explanation only when their scales, variable meanings, time windows, units, and transformations remain aligned. EarthVerse evaluates this evolving claim--evidence state: a single broken binding can make a locally correct calculation describe the wrong Earth process~\citep{du2025deepresearchbench,jiang2026deep}.}

EarthVerse contains 405 reproducible investigations grounded in 199 real disasters and extreme events. Each task opens an event collection assembled from 80 external source and access families. A typical event contains about 34 files, and solving a task requires evidence from more than seven distinct sources on average. Every file is connected to the event, so source choice remains a scientific judgment. The difficulty lies in deciding which records answer the question, whether their time windows and spatial support are compatible, and how much weight each source should carry. Working through these materials takes sustained interaction. Across the leaderboard, each task takes about 17 model rounds on average. A full evaluation of a frontier model costs more than \$2,500.

\begin{figure}[t]
\centering
\mainfigure[width=0.86\linewidth]{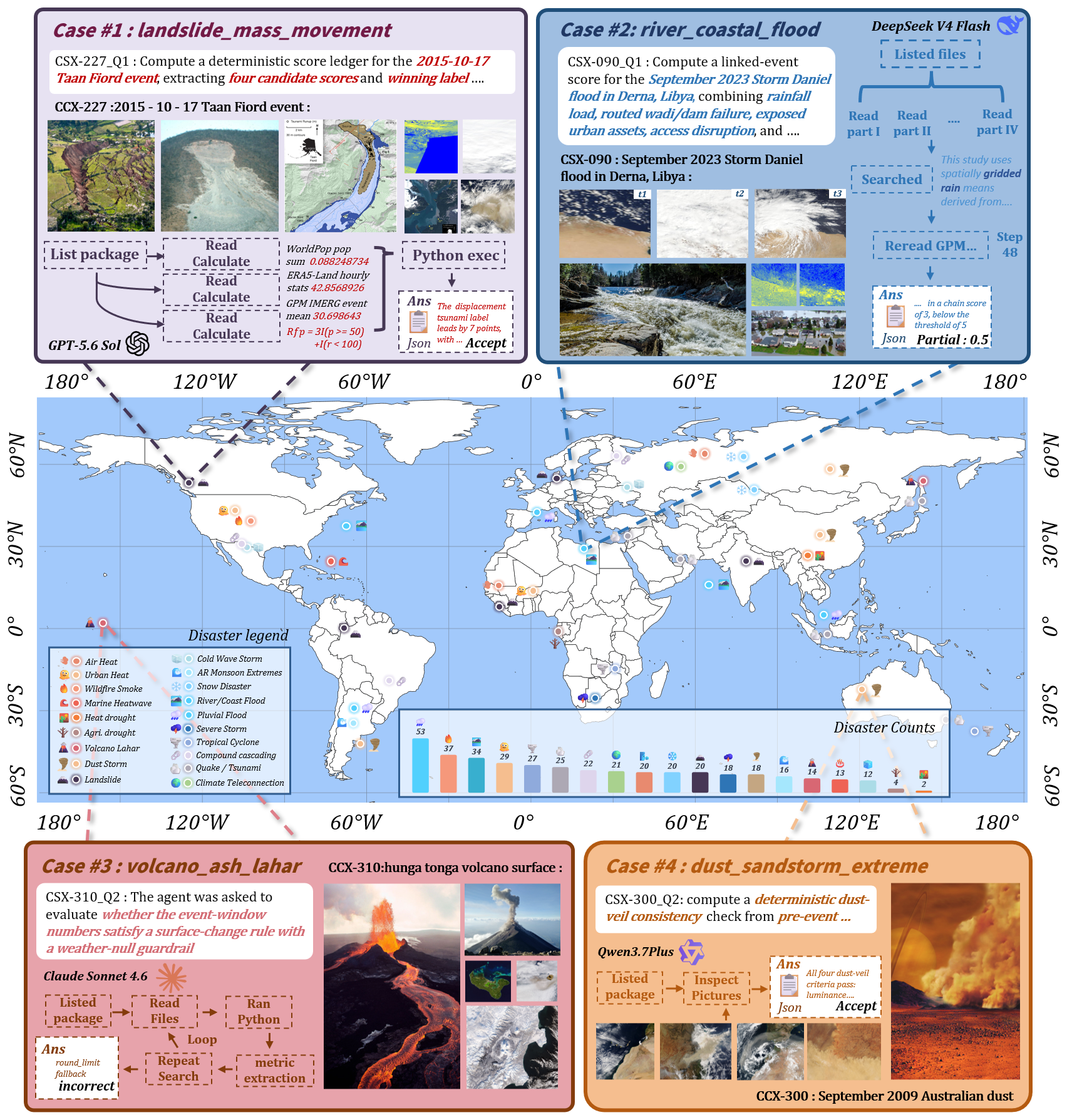}
\caption{\textbf{Benchmark overview.} Global event coverage and four representative investigations. Each case follows a different model trajectory, and performance is scored against the answer units defined for that task. In Case~2, the linked-event score is a deterministic composite derived from package evidence.}
\label{fig:event-investigations}
\end{figure}

Figure~\ref{fig:event-investigations} samples this breadth, pairing the global event collection with representative investigations that move from heterogeneous observations to an auditable conclusion.

EarthVerse uses transparent mathematics: aggregation, ratios, thresholds, spatial overlap, weighted indices, and bounded counterfactuals. The difficulty lies in combining these operations with compatible evidence. A wrong time window changes the total, which can alter both the comparison and the physical explanation. Tasks therefore combine process reconstruction, quantitative analysis, competing hypotheses, rankings, and open synthesis. Fine-grained answer units expose each obligation, while Strict@95 measures whether the resulting claim--evidence state is nearly complete.

The results show how demanding this composition is. Across 25 systems, the best mean answer-unit accuracy is 84.65\%, whereas the highest Strict@95 is only 34.81\%. Even the strongest system leaves roughly two thirds of investigations with at least one consequential error or omission. This is the benchmark's central reliability gap: an agent may recover most local facts and calculations while one unsupported source, mismatched scale, or missing mechanism breaks the account as a whole.

Figure~\ref{fig:coverage-leaderboard} places the benchmark's scientific coverage beside the resulting model landscape.

\paragraph{Contributions.}
We contribute \textbf{(i)} EarthVerse, a benchmark of reproducible multi-source investigations with executable answer units and task-specific process rubrics; \textbf{(ii)} a controlled evaluation of 25 model and agent systems that separates average scientific competence from reliable completion; and \textbf{(iii)} controlled experiments across evidence localization, tool selection, memory, reasoning, interaction modes, and professional scientific environments that identify the main failure points along the research chain.

\paragraph{Research questions.}
We organize the study around three questions. \textbf{RQ1:} Can current systems complete an auditable multi-source hazard investigation, and which missing links separate high average accuracy from reliable completion? \textbf{RQ2:} When does reasoning improve a scientific investigation, and which controls over evidence access, tools, memory, stopping, and execution keep the claim--evidence state open to correction? \textbf{RQ3:} How much of scientific-agent performance comes from interpreting supplied observations, and how much depends on constructing and maintaining the evidence base?

\FloatBarrier

\section{Related Work}
\label{sec:related-work}

\begin{wrapfigure}[18]{r}{0.44\linewidth}
\vspace{-6pt}
\centering
\mainfigure[width=\linewidth]{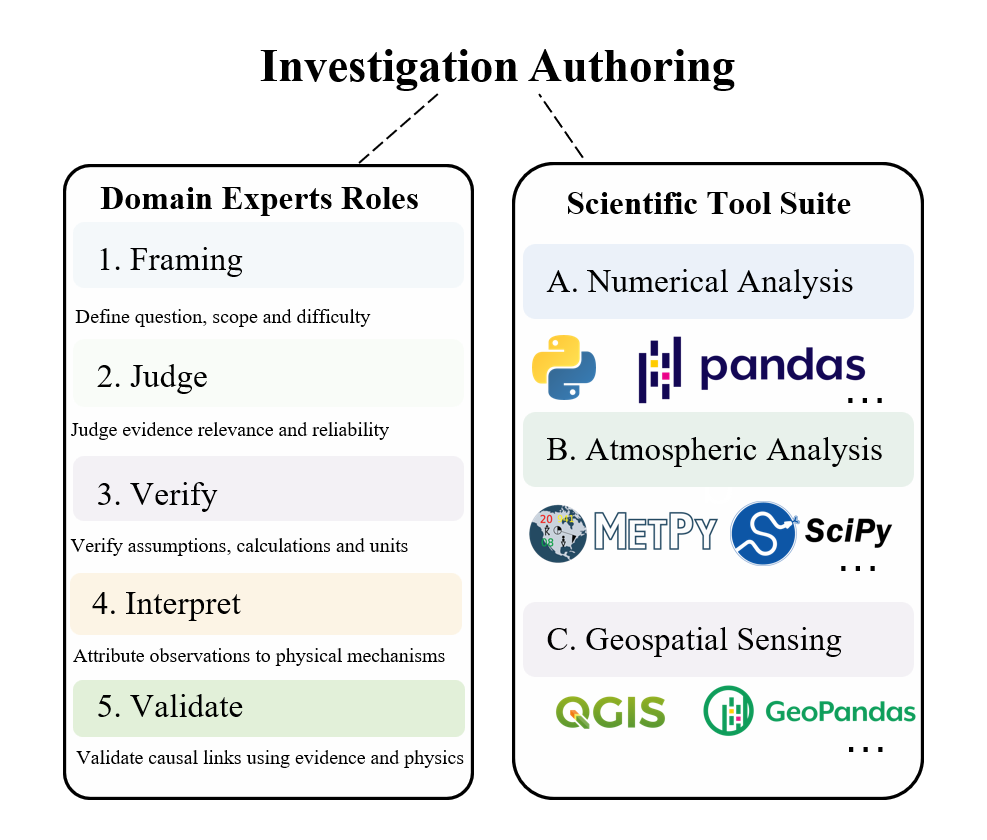}
\caption{Expert roles and scientific tools for investigation authoring.}
\label{fig:expert-tool-authoring}
\vspace{-5pt}
\end{wrapfigure}

\paragraph{Earth observation and scientific multimodality.}
Most Earth-science benchmarks isolate a particular form of interpretation: scientific figures, atmospheric processes, geospatial relations, or remote-sensing imagery~\citep{xu2025earthse,zhao2025msearth,xiao2026geommbench,li2025atmossci,danish2025geobench,wang2025xlrs}. Complementary work studies how representations transfer across sensors, regions, and downstream tasks~\citep{szwarcman2024prithvi,jakubik2025terramind,simumba2025geobench2,cui2026omniair}. These studies establish strong perception and forecasting settings, but the observation set is generally prepared before the model begins its analysis.

\paragraph{Disaster intelligence.}
Disaster benchmarks have followed a similar progression, from classification and damage mapping over supplied media to tool-supported emergency analysis~\citep{alam2018crisismmd,gupta2019xbd,rahnemoonfar2021floodnet}. DORA moves closer to an operational workflow through expert-authored tasks, typed tools, and replayable reference calls, while retaining a declared set of relevant observations and layers~\citep{wang2026can}.

\paragraph{Tool-using and scientific agents.}
Agent research supplies the missing interaction machinery. Early work established reasoning--action loops and learned API use~\citep{yao2022react,schick2023toolformer,qin2023toolllm}; later benchmarks placed agents in executable environments and scientific workflows involving code, data, and long-horizon orchestration~\citep{liu2023agentbench,mialon2023gaia,xie2024osworld,chen2025scienceagentbench,ma2024sciagent,shen2026sciagentgym}. BLADE and DiscoveryBench are especially relevant because they allow multiple valid analysis paths while keeping the resulting work evaluable~\citep{gu2024blade,majumder2025discoverybench}.

EarthVerse couples expert authoring with a scientific tool suite (Figure~\ref{fig:expert-tool-authoring}). Experts frame and verify each investigation; numerical, atmospheric, and geospatial tools support executable analysis. Evaluation then scores source choice, intermediate work, and physical interpretation separately from the final answer.

\paragraph{Geospatial and deep-research systems.}
Recent systems bring these strands together through remote-sensing tools, coordinated geospatial operations, iterative retrieval, or domain-specific pretraining~\citep{shabbir2025thinkgeo,feng2026earth,shabbir2026openearthagent,bao2026spatial,xiao2026geommbench,zheng2025deepresearcher,li2026openresearcher,miromind2025mirothinker,yu2026omnideepsearch,bai2025intern,zhang2024geogpt,ge2025geohazardgpt}. EarthVerse evaluates the combined research problem: whether an agent can choose compatible evidence, use scientific tools appropriately, and maintain a coherent account as an event analysis develops.

\makeatletter\WF@finale\makeatother

\section{The EarthVerse Benchmark}
\label{sec:benchmark}

EarthVerse contains 405 investigations grounded in 199 documented disasters and extreme events, covering 19 hazard families and 6,709 local files. Each task follows the rhythm of real post-event inquiry: establish what happened, identify the observations that can resolve the question, test the relevant quantities and mechanisms, reconcile disagreement, and state what the evidence supports. Packages preserve the complexity of the event without adding large collections of unrelated files merely to obstruct retrieval.

Multi-source misalignment is part of the scientific problem. Reports, stations, satellite products, reanalyses, exposure layers, and impact records may use different windows, footprints, variables, and units. A solver must inspect these differences before choosing evidence, then manage the dependency between tools, sources, calculations, and physical reasoning. Computation is frequent but deliberately transparent: aggregation, normalization, ratios, thresholds, lags, spatial overlap, weighted indices, and bounded counterfactuals. EarthVerse does not test recall of undocumented formulas or background-heavy differential equations. It tests whether simple, checkable operations can support a complex investigation.

\subsection{Design principles}
\label{sec:design-principles}

Four rules govern task construction.

\textbf{Event-scale scientific iteration.} The solver first identifies the event window, geographic extent, hazard evolution, evidence families, and major inconsistencies. It then alternates between targeted evidence access, calculation, and physical interpretation, revising the event picture when an observation conflicts with the current account.

\textbf{Evidence discovery.} The prompt names the event and scientific decision but not the relevant files. A released task requires 7.26 distinct sources on average, selected from a package with a median of 34 candidates. Across the release, evidence comes from dozens of distinct product and API families rather than one fixed archive. Tasks that reveal a path, filename, or answer-identifying quantity are rejected.

\textbf{Scientific checkability.} Conclusions are scored with their supporting values, units, source roles, and mechanism tests. The benchmark tests whether an operation is attached to the right evidence, not whether specialized mathematics can be reproduced from memory.

\textbf{Path flexibility under package grounding.} Report-first and time-series-first investigations can both receive full credit if they establish the required quantities and mechanism. Versioned packages keep the evidence and computation reproducible.

\subsection{Fidelity to investigative practice}
\label{sec:fidelity}

Post-event studies usually begin with an archive rather than a curated figure. Analysts compare agency bulletins, station records at different aggregations, satellite and reanalysis products, impact databases, and incomplete metadata. One source may refine the time window; another may challenge the assumed mechanism; a third may reveal a spatial mismatch. EarthVerse reproduces this investigative sequence in a fixed, replayable setting. The scientific question defines the decision, answer units specify what must be established, and the rubric scores source choice, calculation, revision, and causal argument. The benchmark stays close to disaster practice without relying on live feeds or unstated specialist knowledge.

\subsection{Task construction and answer contracts}

Each event contributes two tasks, with a third for seven flagship events. The public prompt states a scientific decision but does not identify the relevant files. Every task includes an expert solution, structured ground truth, a deterministic \texttt{compute\_gt.py} program, and a task-specific process rubric.

The task and evaluation flow is
\begin{equation}
\begin{aligned}
\left(q_i,P_i\right)
&\xrightarrow{\ \pi\ }
\tau_i=\left(a_1,o_1,\ldots,a_T,o_T\right)
\longrightarrow \hat y_i,\\
E_i
&=\left[f\!\left(\hat y_i,y_i^{\star}\right),
g\!\left(\tau_i;s_i,B_i\right)\right].
\end{aligned}
\label{eq:task-definition}
\end{equation}

\Needspace{0.38\textheight}
\begin{wrapfigure}[18]{r}{0.50\linewidth}
\vspace{-4pt}
\centering
\mainfigure[width=\linewidth]{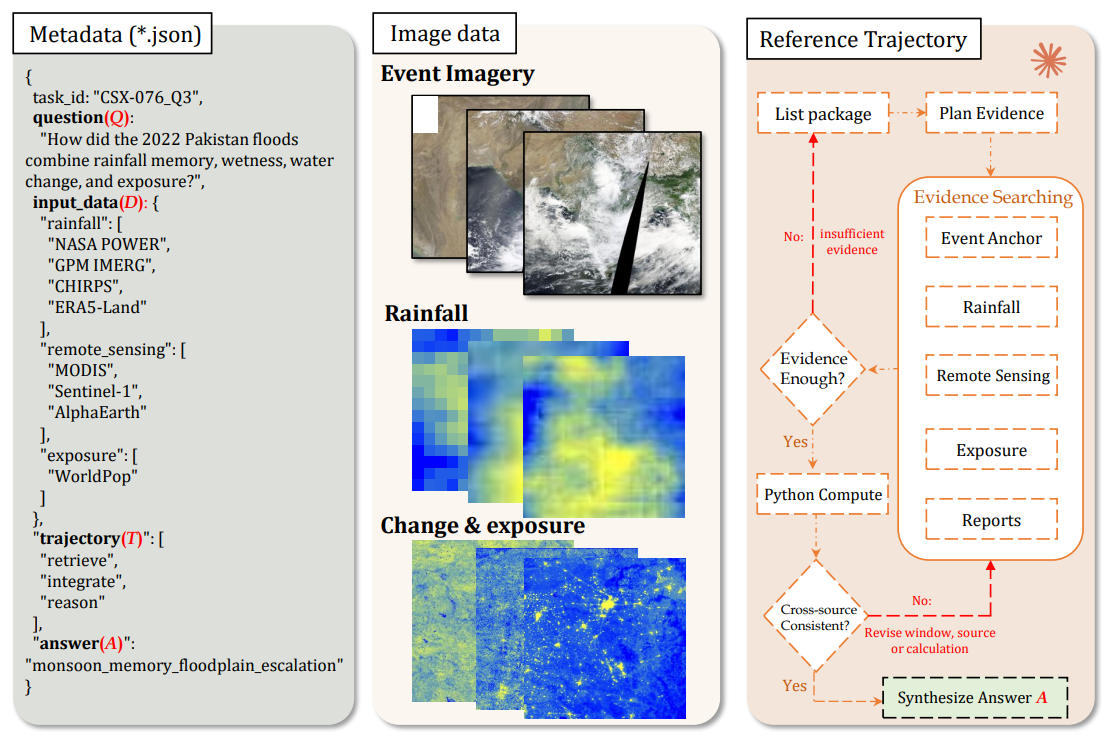}
\caption{\textbf{Task interface.} Metadata, observations, and an auditable research trajectory.}
\label{fig:task-evidence-trajectory}
\vspace{-4pt}
\end{wrapfigure}

For task $i$, the solver sees question $q_i$ and evidence collection $P_i$. Policy $\pi$ produces trajectory $\tau_i$ and answer $\hat y_i$. Function $f$ compares the answer with structured ground truth $y_i^{\star}$; $g$ evaluates the trajectory against expert analysis $s_i$ and rubric $B_i$ without prescribing a tool order. The release contains 10,879 answer units. We use units because a single choice cannot show whether a system recovered the correct window, ranking, quantity, mechanism, or qualification. Each unit is a nontrivial scientific obligation, and tasks can combine numerical results with orderings and open-ended conclusions. This remains discriminative when frontier models already perform extremely well on simpler multiple-choice and short-answer formats.

\paragraph{A worked example.}
One task examines the 2021 Pacific Northwest heat wave. The solver must distinguish a direct hourly record from a lower-resolution aggregate, identify the matching heat window, and compute heat and humidity statistics on the same temporal support. The arithmetic is elementary; the scientific work lies in choosing compatible observations and carrying that choice into the conclusion.

\makeatletter\WF@finale\makeatother

\subsection{Capabilities and research environment}
\label{sec:capability-environment}

\paragraph{Capability coverage.}
Each task receives one to three manually reviewed capability labels: physical-mechanism reasoning, spatiotemporal reconstruction, quantitative calculation, multi-source synthesis, causal-chain reasoning, ranking or decision, and remote-sensing or geospatial interpretation. Labels may overlap. Conditioned scores are diagnostic and do not change the overall metric.

Figure~\ref{fig:coverage-leaderboard}(a) shows that these capabilities recur across hazard families rather than forming separate task silos. Figure~\ref{fig:task-evidence-trajectory} illustrates how a solver links task metadata, observations, and its research trajectory.

\paragraph{Research tools.}
We selected six general operations that are sufficient to express every released workflow: discover files, read a source, search local text, inspect structured data, execute scoped Python, and finalize the answer. This common interface makes system comparisons consistent while leaving the scientific strategy open. The registry also supports custom tool registration and retains 170 reusable operations for reports, time series, vector and exposure analysis, raster and Earth observation, meteorology and climate, fire and geophysics, infrastructure, and output assembly. Structured returns separate execution failure from scientific disagreement.

\subsection{Quality assurance and benchmark comparison}
\label{sec:benchmark-comparison}

\begin{figure}[t]
\centering
\mainfigure[width=\linewidth]{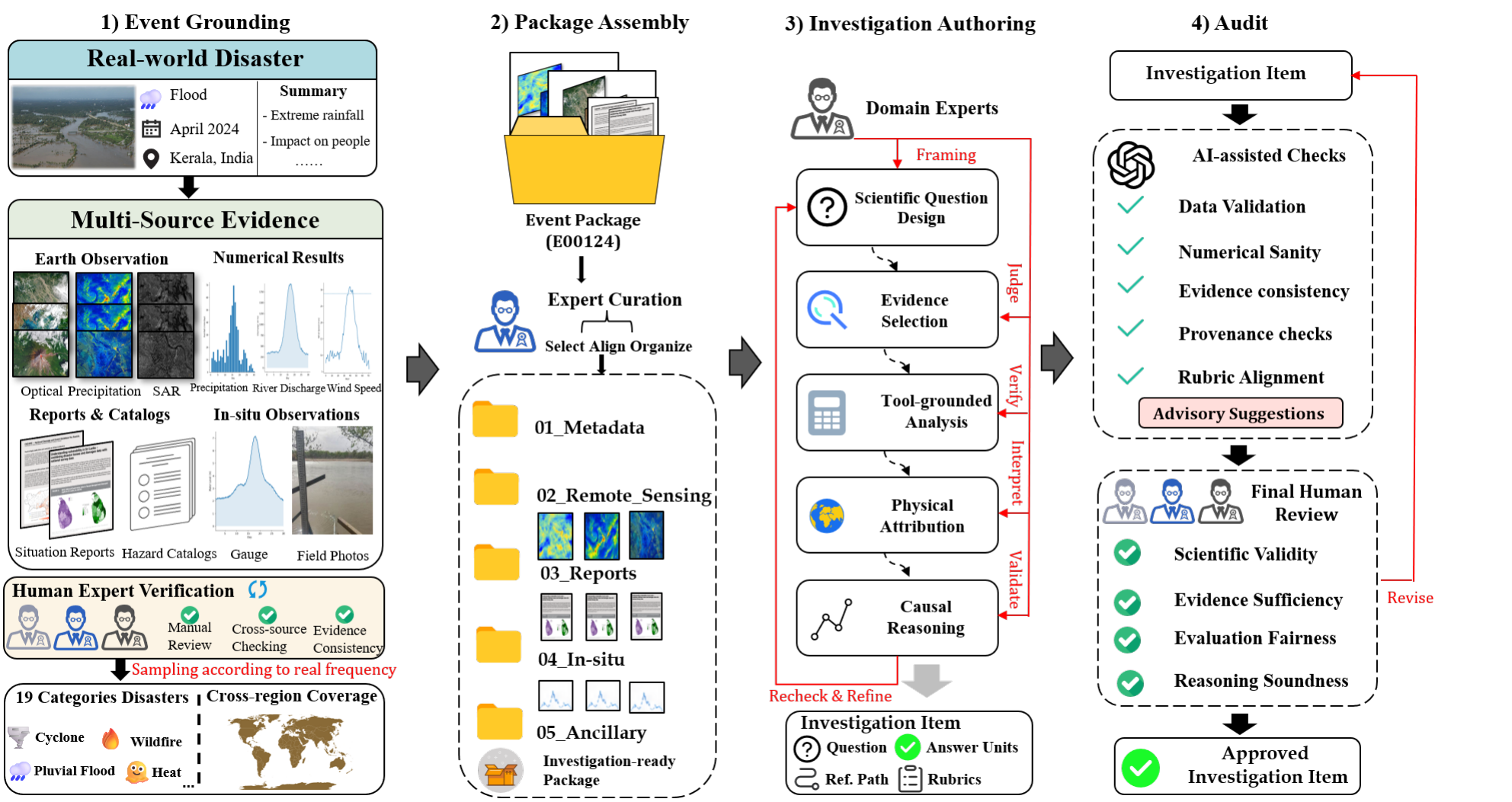}
\caption{\textbf{Benchmark construction.} Event grounding, package assembly, investigation authoring, and expert audit.}
\label{fig:benchmark-construction}
\end{figure}

\paragraph{Construction and review.}
Construction proceeds through event verification, package assembly, task authoring, executable validation, and release review (Figure~\ref{fig:benchmark-construction}). Automated tools support the process, but every released task is checked and corrected by people. Review covers event identity, source reliability, temporal and spatial fit, the public question, expert solution, ground truth, executable program, capability labels, and rubric. Package-local data must regenerate every deterministic value, and the visible assignment must match the hidden scoring target. Appendix~\ref{app:human-review} describes the human and agent review process; Appendix~\ref{app:construction} records the seven release gates.

A separate LLM pairwise comparison offers a second check on score validity. Its win--tie--loss pattern broadly follows the Core leaderboard across system families (Figure~\ref{fig:pairwise-preference}). Systems with higher Core are also preferred when complete answers and traces are compared directly, so the ranking is not driven solely by the number of satisfied scoring fields.

\paragraph{Position relative to prior benchmarks.}
Table~\ref{tab:benchmark-comparison} compares what the model is asked to do, what evidence it receives, how experts participate, and what is scored. EarthSE, MSEarth, and GeoMMBench usually provide one selected passage, figure, or image~\citep{xu2025earthse,zhao2025msearth,xiao2026geommbench}. ThinkGeo, Earth-Agent, OpenEarthAgent, and DORA add tools around declared imagery, assets, or layers~\citep{shabbir2025thinkgeo,feng2026earth,shabbir2026openearthagent,wang2026can}. EarthVerse instead asks the system to construct and document a multi-source evidence base whose contents depend on the disaster and the scientific question.

\begin{table}[!htbp]
\centering
\caption{\textbf{Where the scientific work begins.} The two text columns identify the starting evidence and analytical work; marks show full, partial, or absent/unreported coverage.}
\label{tab:benchmark-comparison}
\tabfontcompact
\renewcommand{\arraystretch}{1.12}
\setlength{\tabcolsep}{1.5pt}
\begin{tabularx}{0.99\linewidth}{@{}>{\raggedright\arraybackslash}p{0.14\linewidth} >{\raggedright\arraybackslash}p{0.12\linewidth} >{\raggedright\arraybackslash}p{0.135\linewidth} >{\centering\arraybackslash}p{0.055\linewidth} *{6}{C}@{}}
\toprule
\tblhead
Benchmark & Start evidence & Scientific work & Tools & \makecell{Find\\evidence} & \makecell{Multi-\\source} & Compute & \makecell{Unit\\score} & \makecell{Trace\\score} & \makecell{Human\\review} \\
\midrule
\href{https://arxiv.org/abs/2505.17139}{EarthSE} & Paper & Sci. QA & \xmark & \xmark & \xmark & \pmark & \xmark & \xmark & \pmark \\
\href{https://arxiv.org/abs/2505.20740}{MSEarth} & Figure & Figure QA & \xmark & \xmark & \xmark & \xmark & \xmark & \xmark & \pmark \\
\href{https://arxiv.org/abs/2604.08896}{GeoMMBench} & Image & Geo reasoning & \xmark & \xmark & \xmark & \xmark & \xmark & \xmark & \pmark \\
\href{https://arxiv.org/abs/2505.23752}{ThinkGeo} & RS image & Tool use & \cmark & \pmark & \pmark & \cmark & \xmark & \cmark & \pmark \\
\href{https://arxiv.org/abs/2509.23141}{Earth-Agent} & EO assets & EO planning & \cmark & \pmark & \cmark & \cmark & \xmark & \cmark & \pmark \\
\href{https://arxiv.org/abs/2602.17665}{OpenEarthAgent} & Imagery & Multimodal & \cmark & \pmark & \pmark & \cmark & \xmark & \cmark & \xmark \\
\href{https://arxiv.org/abs/2605.11633}{DORA} & Layers & Crisis analysis & \cmark & \pmark & \cmark & \cmark & \xmark & \cmark & \pmark \\
\addlinespace[2pt]
\keyrow \textbf{EarthVerse} & \textbf{Package} & \textbf{Event synthesis} & \cmark & \cmark & \cmark & \cmark & \cmark & \cmark & \cmark \\
\bottomrule
\end{tabularx}
\par\vspace{2pt}
\noindent\scriptsize \cmark\ full \quad \pmark\ partial \quad \xmark\ absent/unreported \hfill
\textit{EarthVerse scale:} 34 files/package; 7.26 sources and 26.86 units/task.
\end{table}

EarthVerse places the difficulty in scientific assembly rather than specialized mathematics. Fine-grained answer units state what must be established; the process rubric tests whether the chosen path supports those claims.

\section{Evaluation Protocol}
\label{sec:experiments}

EarthVerse fixes the scientific question, visible evidence boundary, execution budget, and answer contract across systems. During a run, the system selects evidence, performs calculations, records provenance, and submits a structured scientific answer through the same interface. The full inference configuration, model modalities, and controlled-study manifests are reported in Appendix~\ref{app:reproducibility}.

\subsection{Outcome, process, and reliability metrics}

For task $i$, let $U_i$ be the percentage of required answer units satisfied and $H_i$ the holistic answer judgment, both on a 0--100 scale. Answer correctness is $A_i=\min(H_i,U_i)$: a fluent answer cannot outrank the scientific obligations it actually completes. Let $P_i$ be the task-specific process-rubric score on the same scale. The primary score is
\begin{equation}
\label{eq:metrics}
\text{Combined Core}
=\frac{1}{2N}\sum_{i=1}^{N}\left(A_i+P_i\right).
\end{equation}
It therefore gives equal weight to endpoint correctness and the documented research process.

Answer units are necessary because one investigation may require a time window, several quantities, a source comparison, and a physical mechanism. A single holistic grade would hide which obligation failed and could let a fluent conclusion mask missing scientific work.

Mean unit accuracy averages $U_i$ across tasks and captures partial completion of required scientific claims and output fields; numerical units use task-specific tolerances. Strict@95 is the percentage of tasks with $U_i\geq95$ and measures near-complete task reliability rather than average progress. Capability-conditioned scores apply the same Core calculation to expert-tagged task slices. Trajectory statistics report interaction, evidence access, computation, latency, and token demand, but never enter the score. Solver failures receive zero; judge-service failures are rerun until a valid judgment is obtained.

\paragraph{Path pluralism and diagnostic validity.}
EarthVerse does not prescribe a tool order. Earth-Agent and DORA instead evaluate whether an agent follows a predefined tool or operational sequence. This is useful when a workflow has a canonical procedure, but complex multi-source investigations rarely admit a single defensible trajectory ground truth: report-first and time-series-first analyses may reach the same well-supported conclusion. EarthVerse therefore scores what the system establishes and whether the observed evidence and calculations support it. Tools and traces serve reasoning, review, and diagnosis; reproducing a prescribed path earns no credit by itself.

\subsection{Evaluated systems}
\label{sec:model-settings}

Table~\ref{tab:main-results} compares hosted systems, open-weight models, and research frameworks through one package interface. Some backbones are text-only, while others support native multimodal input; the shared leaderboard does not grant extra evidence to either group. All systems receive the same question, package boundary, answer contract, and execution budget. Appendix~\ref{app:model-interface} gives the complete roster, modality treatment, controller exceptions, and judge configuration.

\section{Main Evaluation}
\label{sec:main-results}

The results follow the three questions. The leaderboard and reliability diagnostics in Section~\ref{sec:overall-performance} answer RQ1. Reasoning, interaction, and evidence-path interventions in Section~\ref{sec:reasoning-evidence-control} answer RQ2. Section~\ref{sec:cross-benchmark} addresses RQ3 by comparing supplied-observation performance with evidence construction. Appendix~\ref{sec:findings} develops the resulting position.

\subsection{Overall system performance}
\label{sec:overall-performance}

RQ1 concerns completion at the investigation level. We begin by asking whether current systems can keep the whole account valid when most individual parts are correct.

\paragraph{Setting.}
Table~\ref{tab:main-results} reports all 25 systems under the package-scoped interactive protocol. The Harness column names the controller used to run each model. Core averages Answer correctness and Process-rubric score; Strict@95 counts tasks satisfying at least 95\% of answer units. Trajectory columns are per-task means and do not affect quality scores. Systems are grouped by family and ordered from lower to higher Core.

\begin{table}[!htbp]
\centering
\caption{\textbf{EarthVerse leaderboard.} Harness names the controller. Scores are percentages; failed, timed-out, or malformed runs remain zero, and batch reads may cover multiple files. Within each family, Core increases downward; \best{bold blue} and \second{underlined teal} mark the top two outcomes.}
\label{tab:main-results}
\tabfontdense
\setlength{\tabcolsep}{2.0pt}
\renewcommand{\arraystretch}{0.94}
\resizebox{\textwidth}{!}{%
\begin{tabular}{@{}l l *{12}{r}@{}}
\toprule
\multicolumn{2}{c}{System configuration} & \multicolumn{5}{c}{Outcome} & \multicolumn{7}{c}{Trajectory mean per task} \\
\cmidrule(lr){1-2}\cmidrule(lr){3-7}\cmidrule(l){8-14}
System & Harness & Answer & Process & Core & \makecell{Strict@95\\(\%)} & \makecell{Unit\\acc.} & Rounds & \makecell{Tool\\calls} & Reads & Files & Python & \makecell{Latency\\(s)} & \makecell{Tokens\\(k)} \\
\midrule
\tblpanel{14}{Agent frameworks (own controllers over the same package interface)}
OpenResearcher & OpenResearcher & 30.51 & 31.06 & 30.79 & 4.94 & 32.74 & 39.73 & 33.62 & 15.00 & 17.03 & 5.38 & 521 & 538.7 \\
GPT-5.5 & GeoMMAgent & 65.86 & 70.84 & 68.35 & 12.35 & 67.18 & 5.00 & 8.28 & 11.93 & 8.59 & 0.99 & 221 & 37.9 \\
\addlinespace[2pt]
\tblpanel{14}{Open-weight and Earth-specialized models}
Intern-S1-mini & EarthVerse & 10.78 & 12.67 & 11.72 & 0.25 & 11.60 & 4.99 & 3.43 & 1.80 & 2.39 & 0.07 & 57 & 39.4 \\
Nemotron Nano & EarthVerse & 12.30 & 14.01 & 13.15 & 0.25 & 12.80 & 2.59 & 2.07 & 1.26 & 1.77 & 0.15 & 279 & 53.2 \\
MiroThinker 8B & EarthVerse & 12.79 & 13.62 & 13.21 & 2.96 & 13.42 & 27.30 & 25.98 & 6.07 & 11.42 & 2.00 & 367 & 318.0 \\
GeoGPT & EarthVerse & 22.15 & 27.32 & 24.74 & 0.25 & 23.61 & 30.41 & 29.37 & 19.33 & 11.86 & 2.65 & 565 & 502.1 \\
Qwen3-4B & EarthVerse & 25.11 & 24.94 & 25.02 & 0.00 & 26.13 & 5.06 & 3.79 & 2.55 & 3.66 & 0.07 & 132 & 37.7 \\
GeohazardGPT & EarthVerse & 28.88 & 33.33 & 31.10 & 0.49 & 30.27 & 30.33 & 29.34 & 13.38 & 15.70 & 8.44 & 456 & 333.7 \\
Qwen3-235B & EarthVerse & 31.98 & 34.64 & 33.31 & 0.99 & 32.66 & 13.47 & 12.80 & 5.33 & 7.41 & 3.50 & 195 & 274.3 \\
\addlinespace[2pt]
\tblpanel{14}{Hosted general-purpose systems}
GPT-5.6 Luna & Codex & 35.68 & 35.63 & 35.66 & 11.11 & 36.28 & 6.58 & 5.39 & 2.37 & 4.29 & 1.05 & 189 & 156.9 \\
GPT-4o & EarthVerse & 37.35 & 41.11 & 39.23 & 0.49 & 38.69 & 18.81 & 8.95 & 4.04 & 6.84 & 0.69 & 55 & 127.4 \\
MiniMax M2.7 & EarthVerse & 43.66 & 45.77 & 44.71 & 7.41 & 44.35 & 17.27 & 16.08 & 7.11 & 8.94 & 2.22 & 205 & 285.8 \\
Gemini 3.1 Flash Lite & EarthVerse & 57.67 & 61.66 & 59.66 & 2.96 & 58.67 & 7.37 & 6.10 & 3.59 & 5.69 & 1.08 & 24 & 70.1 \\
Doubao Seed 2.0 Mini & EarthVerse & 61.14 & 64.84 & 62.99 & 5.93 & 62.22 & 13.71 & 12.70 & 7.21 & 8.86 & 1.08 & 113 & 175.1 \\
Qwen 3.7 Plus & EarthVerse & 69.33 & 71.37 & 70.35 & 18.27 & 70.16 & 17.15 & 15.72 & 6.51 & 11.25 & 3.31 & 313 & 204.3 \\
Claude Haiku 4.5 & EarthVerse & 70.76 & 73.70 & 72.23 & 14.07 & 71.70 & 21.31 & 19.86 & 10.08 & 11.72 & 3.96 & 324 & 477.0 \\
Tencent Hy3 & EarthVerse & 71.27 & 74.26 & 72.77 & 16.30 & 72.22 & 19.47 & 17.77 & 8.32 & 12.70 & 4.41 & 311 & 208.7 \\
DeepSeek V4 Flash & EarthVerse & 72.32 & 74.77 & 73.54 & 20.25 & 73.25 & 27.92 & 26.81 & 11.00 & 16.38 & 4.07 & 333 & 389.7 \\
Kimi K2.5 & EarthVerse & 73.89 & 76.76 & 75.32 & 14.57 & 74.70 & 24.25 & 22.67 & 10.01 & 16.00 & 4.79 & 439 & 302.9 \\
Claude Sonnet 4.6 & EarthVerse & 74.88 & 77.44 & 76.16 & 20.00 & 75.76 & 17.54 & 16.54 & 7.72 & 11.82 & 4.49 & 343 & 331.0 \\
GPT-5.5 & EarthVerse & 76.85 & 79.28 & 78.06 & 17.04 & 77.48 & 15.17 & 14.20 & 6.83 & 10.92 & 3.74 & 210 & 224.9 \\
GLM-5.2 & EarthVerse & 78.14 & 80.19 & 79.17 & 22.96 & 78.62 & 18.32 & 17.27 & 7.52 & 13.36 & 4.31 & 259 & 196.1 \\
GPT-5.6 Terra & Codex & 80.12 & 80.00 & 80.06 & \second{27.16} & 80.81 & 12.18 & 11.13 & 4.42 & 9.46 & 2.71 & 173 & 182.6 \\
GPT-5.6 Sol & Codex & \second{81.43} & \second{81.94} & \second{81.68} & \best{34.81} & \second{82.12} & 13.88 & 12.79 & 4.88 & 9.65 & 2.87 & 205 & 205.9 \\
\keyrow Claude Fable 5 & Claude Code & \best{83.99} & \best{85.95} & \best{84.97} & 25.93 & \best{84.65} & 14.45 & 13.45 & 5.92 & 10.02 & 2.61 & 247 & 161.6 \\
\bottomrule
\end{tabular}%
}
\end{table}

The landscape in Figure~\ref{fig:coverage-leaderboard}(b) shows that similar token budgets can produce sharply different answer-unit accuracy.

\paragraph{Mean competence versus conjunctive reliability.}
Claude Fable 5 leads on Core (84.97), while GPT-5.6 Sol leads on Strict@95 (34.81\%). Fable satisfies 84.65\% of answer units on average but reaches 95\% completion on only 25.93\% of tasks; Sol leaves fewer investigations with a consequential omission. Missing windows, provenance, alternative mechanisms, and unit conversions are coupled errors: a wrong window changes the calculation and its physical interpretation. Low Strict@95 rates show that current systems often know most of the analysis without keeping the whole analysis valid at once.

\findingbox{finding:conjunctive}{Reliability requires the whole evidence chain}{High average accuracy can hide one missing source, window, unit, calculation, or mechanism. In disaster analysis, that single break can invalidate an otherwise plausible account.}

\paragraph{Coupling reasoning with interaction.}
Most leaderboard systems use the EarthVerse harness, a common multi-round controller configured for evidence discovery, calculation, provenance, recovery, and structured finalization. Holding this scaffold fixed makes backbone comparisons fair, while framework rows retain their native controllers. Interaction volume still does not predict quality. OpenResearcher uses 39.73 rounds and 538.7k tokens per task, yet trails Claude Fable 5 by 54.18 Core points because weak sources and stale values survive into late revisions. GeoMMAgent is too brief in the opposite direction: five rounds leave evidence undiscovered and Strict@95 reaches only 12.35\%. Useful interaction updates the claim--evidence state and carries the revision into the answer.

\findingbox{finding:control}{Useful interaction changes the scientific state}{A tool call matters when its observation changes the chosen source, calculation, or physical account. Longer traces that preserve the same mistake add cost, not quality.}

\paragraph{Domain knowledge and frontier reasoning.}
Earth-specialized pretraining improves terminology and process recognition~\citep{bai2025intern,zhang2024geogpt,ge2025geohazardgpt}, but the evaluated open models remain 40--70 Core points behind frontier general systems. GeohazardGPT averages 30.33 rounds and 8.44 Python calls per task yet reaches 31.10 Core. The gap appears after recognition, in source selection, provenance, revision, and stopping. Domain training remains useful, but in these systems it has not yet produced a reliable research loop.

Capability slices in Appendix~\ref{app:capability} sharpen this result. Several systems rise above their own mean on calculation, spatiotemporal reconstruction, or remote sensing, yet fall on causal reasoning and ranking. Overall strength does not produce a uniform scientific profile.

\paragraph{Computational cost.}
Evaluating a frontier model costs more than \$2,500, while flash models remain below \$45. GPT-5.6 Terra retains 98\% of GPT-5.6 Sol's Core at less than half the cost. Cost depends on both model pricing and controller behavior, so neither trace length nor token count alone is an efficiency measure. Terra offers the strongest quality--cost balance among the top systems, whereas low-cost Flash models are useful for broad screening but do not match frontier reliability. Appendix~\ref{app:cost-accounting} reports the full comparison.

\FloatBarrier

\subsection{Reasoning, interaction, and evidence control}
\label{sec:reasoning-evidence-control}

\begin{wraptable}[25]{r}{0.585\textwidth}
\vspace{-9pt}
\centering
\caption{\textbf{Reasoning and evidence access.} GPT-5.5 across four access modes and five effort settings; diagnostics are per-task means.}
\label{tab:main-reasoning-access}
\label{tab:main-controlled-evidence}
\fontsize{6.5pt}{7.2pt}\selectfont
\setlength{\tabcolsep}{1.35pt}
\renewcommand{\arraystretch}{0.93}
\begin{tabular}{@{}l*{10}{r}@{}}
\toprule
& \multicolumn{4}{c}{Outcome} & \multicolumn{2}{c}{Interaction} & \multicolumn{2}{c}{Evidence} & \multicolumn{2}{c}{Resources} \\
\cmidrule(lr){2-5}\cmidrule(lr){6-7}\cmidrule(lr){8-9}\cmidrule(l){10-11}
Effort & Answer & Process & Core & \makecell{Unit\\acc.} & Rounds & Calls & Files & Refs & \makecell{Tokens\\(k)} & \makecell{Latency\\(s)} \\
\midrule
\tblpanel{11}{Compressed direct}
\keycell{none} & \keycell{\best{41.37}} & \keycell{\best{45.10}} & \keycell{\best{43.24}} & \keycell{\best{42.21}} & -- & -- & -- & -- & \keycell{13.9} & \keycell{55.6} \\
low & 24.99 & 29.32 & 27.16 & 25.43 & -- & -- & -- & -- & 14.9 & 120.5 \\
medium & 22.88 & 25.69 & 24.29 & 23.26 & -- & -- & -- & -- & 15.8 & 152.4 \\
high & 24.17 & 28.17 & 26.17 & 24.28 & -- & -- & -- & -- & 16.6 & 71.5 \\
xhigh & 16.67 & 16.67 & 16.67 & 16.67 & -- & -- & -- & -- & 17.6 & 88.1 \\
\tblpanel{11}{All-text direct}
\keycell{none} & \keycell{\best{27.66}} & \keycell{\best{31.22}} & \keycell{\best{29.44}} & \keycell{\best{28.27}} & -- & -- & -- & -- & \keycell{81.6} & \keycell{32.1} \\
low & 13.26 & 16.20 & 14.73 & 13.36 & -- & -- & -- & -- & 82.7 & 44.1 \\
medium & 9.79 & 12.18 & 10.99 & 10.05 & -- & -- & -- & -- & 83.5 & 61.0 \\
high & 8.80 & 9.00 & 8.90 & 8.82 & -- & -- & -- & -- & 85.4 & 78.5 \\
xhigh & 12.00 & 15.00 & 13.50 & 12.00 & -- & -- & -- & -- & 87.7 & 94.4 \\
\tblpanel{11}{EarthVerse interactive}
none & 48.89 & 53.70 & 51.30 & 49.46 & 9.68 & 8.68 & 7.00 & 9.94 & 96.9 & 140 \\
low & 56.97 & 60.05 & 58.51 & 57.53 & 14.70 & 13.70 & 9.44 & 14.70 & 162.4 & 259 \\
medium & 61.45 & 64.80 & 63.13 & 62.35 & 17.20 & 16.20 & 10.71 & 18.12 & 206.4 & 424 \\
\keyrow high & \best{72.41} & \best{75.19} & \best{73.80} & \best{72.59} & 14.96 & 13.96 & 9.92 & 17.19 & 174.3 & 571 \\
xhigh & 69.66 & 67.50 & 68.58 & 69.83 & 11.25 & 10.25 & 10.25 & 16.50 & 100.0 & 674 \\
\tblpanel{11}{Interactive with final review}
none & 53.70 & 56.48 & 55.09 & 54.06 & 11.62 & 9.62 & 7.42 & 10.46 & 130.7 & 108 \\
low & 61.93 & 65.50 & 63.72 & 62.78 & 18.18 & 16.18 & 10.60 & 17.02 & 227.0 & 253 \\
medium & 67.16 & 69.22 & 68.19 & 67.91 & 20.81 & 18.81 & 11.88 & 21.02 & 266.3 & 552 \\
high & 69.82 & 74.71 & 72.27 & 70.73 & 20.63 & 18.63 & 12.13 & 21.10 & 255.1 & 1,105 \\
\keyrow xhigh & \best{84.36} & \best{88.21} & \best{86.29} & \best{84.40} & 17.57 & 15.57 & 10.14 & 18.29 & 217.7 & 1,493 \\
\bottomrule
\end{tabular}
\vspace{-35pt}
\end{wraptable}

For RQ2, we separate additional deliberation from interventions that let the system revise its claim--evidence state. Tables~\ref{tab:main-reasoning-access} and~\ref{tab:main-evidence-interventions} examine complementary parts of that loop: how evidence can be used during reasoning, and what the system can locate or act on.

Table~\ref{tab:main-reasoning-access} shows a sharp split between fixed and editable evidence. With compressed or all-text direct input, the best result comes from no additional reasoning; high effort lowers Core from 43.24 to 26.17 and from 29.44 to 8.90, respectively. Once the system can retrieve evidence and act on what it finds, Core rises from 51.30 with no extra effort to 73.80 at high effort. Final review changes the best operating point again: xhigh reaches 86.29, compared with 68.58 without review. Reasoning helps when it can trigger retrieval, recomputation, or correction. Deliberating longer over the same snapshot often deepens the initial mistake.

Resource use reinforces this distinction. High effort in the interactive setting reaches 73.80 Core with 14.96 rounds and 174.3k tokens; medium effort uses more rounds and tokens but reaches only 63.13. Additional steps help when they change the evidence set, repair a calculation, or resolve a conflict that reaches the final answer. Trace length alone says little about the quality of the investigation.

Final review also has a conditional effect. At xhigh, it raises Core from 68.58 to 86.29, but at high effort it changes 73.80 to 72.27. Review is most useful when the investigation still contains evidence dependencies or calculations that can be checked. Once the claim--evidence state is coherent, another pass may add cost without improving the answer. This favors targeted verification over a fixed rule that every task needs more reasoning.

\makeatletter\WF@finale\makeatother

To locate the remaining bottlenecks, Table~\ref{tab:main-evidence-interventions} uses answer-free oracles that remove one obstacle at a time: they reveal the relevant files, evidence roles, useful tool classes, or a high-level plan, but never the target answer. The same paired tasks vary tool presentation, package evidence, and access to meteorological operations. The meteorological conditions compare a fixed compact suite with a cue-routed subset; tool use remains optional, so this probe asks whether specialist access alone improves the investigation.

\begin{table}[!htbp]
\centering
\caption{\textbf{Evidence-path interventions.} Complete paired outcome, effect, and capability grid relative to the Top-20 router. Intervals are paired 95\% bootstrap CIs.}
\label{tab:main-evidence-interventions}
\label{tab:bottleneck-robustness}
\label{tab:intervention-ci}
\label{tab:oracle-capability}
\fontsize{6.25pt}{7.2pt}\selectfont
\setlength{\tabcolsep}{0.74pt}
\renewcommand{\arraystretch}{0.98}
\begin{adjustbox}{max width=\textwidth,center}
\begin{tabular}{@{}l *{4}{r} c *{7}{r}@{}}
\toprule
& \multicolumn{3}{c}{Outcome} & \multicolumn{2}{c}{Paired effect} & \multicolumn{7}{c}{Core by capability} \\
\cmidrule(lr){2-4}\cmidrule(lr){5-6}\cmidrule(l){7-13}
Intervention & Answer & Process & Core & \makecell{$\Delta$\\Core} & 95\% CI & \makecell{Physical\\mech.} & \makecell{Spatio-\\temporal} & \makecell{Quant.\\calc.} & \makecell{Multi-\\source} & \makecell{Causal\\chain} & \makecell{Rank\\sort} & \makecell{RS /\\geo.} \\
\midrule
\keyrow Top-20 semantic-router baseline & 66.90 & 65.88 & 66.39 & 0.00 & -- & 65.15 & 64.40 & 73.43 & 90.00 & 65.89 & 50.84 & 53.94 \\
\addlinespace[1pt]\midrule
\tblpanel{13}{Answer-free information oracles}
Relevant-file oracle & \best{79.46} & \best{82.75} & \best{81.11} & \gain{\textbf{+14.72}} & [+5.48, +24.69] & 80.61 & 91.47 & 87.57 & 90.45 & 76.53 & 60.70 & 72.09 \\
Evidence-map oracle & 76.96 & 78.50 & 77.73 & \gain{+11.34} & [+2.34, +21.25] & 76.69 & 90.50 & 83.83 & 97.50 & 73.35 & 58.16 & 61.25 \\
Tool-subset oracle & 68.53 & 68.62 & 68.58 & \gain{+2.19} & [$-$3.55, +7.28] & 67.32 & 65.67 & 72.60 & 92.50 & 69.54 & 59.14 & 59.75 \\
Planning oracle & 65.19 & 66.50 & 65.84 & \loss{$-$0.54} & [$-$7.07, +4.35] & 64.31 & 57.10 & 71.24 & 95.00 & 67.93 & 54.76 & 60.25 \\
\addlinespace[1pt]\midrule
\tblpanel{13}{Tool-interface presentation}
Hierarchical catalog & 66.76 & 67.62 & 67.19 & \gain{+0.80} & [$-$9.18, +11.86] & 65.81 & 66.72 & 67.83 & 93.50 & 70.78 & 56.60 & 61.67 \\
Descriptive tool names & 66.88 & 66.75 & 66.82 & \gain{+0.43} & [$-$3.55, +4.05] & 65.07 & 61.21 & 72.92 & 100.00 & 66.98 & 57.59 & 55.23 \\
Dynamic descriptions & 61.39 & 61.75 & 61.57 & \loss{$-$4.82} & [$-$15.02, +4.33] & 59.81 & 61.56 & 63.81 & 95.00 & 61.27 & 54.58 & 60.75 \\
Opaque tool names & 56.72 & 59.50 & 58.11 & \loss{$-$8.28} & [$-$23.57, +6.33] & 56.16 & 48.88 & 58.54 & 95.22 & 59.76 & 54.01 & 80.00 \\
\addlinespace[1pt]\midrule
\tblpanel{13}{Evidence perturbations}
Equivalent unit change & 67.61 & 71.12 & 69.37 & \gain{+2.98} & [$-$9.75, +15.53] & 68.02 & 66.93 & 71.44 & 95.00 & 71.29 & 55.47 & 76.38 \\
Added distractor files & 65.91 & 67.75 & 66.83 & \gain{+0.44} & [$-$6.80, +7.27] & 65.46 & 60.47 & 72.15 & 92.95 & 65.16 & 59.98 & 70.62 \\
Missing evidence & 63.97 & 64.75 & 64.36 & \loss{$-$2.03} & [$-$12.05, +7.08] & 62.49 & 60.31 & 67.82 & 100.00 & 66.63 & 54.74 & 55.12 \\
Truncated evidence & 63.17 & 63.38 & 63.27 & \loss{$-$3.12} & [$-$14.13, +7.47] & 61.78 & 55.67 & 70.53 & 91.70 & 64.24 & 41.52 & 68.92 \\
Conflicting evidence & 63.03 & 63.12 & 63.08 & \loss{$-$3.31} & [$-$12.05, +2.97] & 61.50 & 67.97 & 66.63 & 92.95 & 60.46 & 55.19 & 54.50 \\
\addlinespace[1pt]\midrule
\tblpanel{13}{Optional meteorological access}
Compact meteorology suite & 62.69 & 65.38 & 64.03 & \loss{$-$2.36} & [$-$11.92, +5.89] & 62.14 & 61.97 & 66.67 & 100.00 & 65.22 & 56.15 & 58.47 \\
Cue-routed meteorology & 57.22 & 59.88 & 58.55 & \loss{$-$7.84} & [$-$19.40, +3.93] & 56.74 & 58.87 & 59.25 & 92.95 & 60.28 & 50.45 & 57.25 \\
\bottomrule
\end{tabular}
\end{adjustbox}
\end{table}

\findingbox{finding:editable-evidence-state}{Reasoning needs an editable claim--evidence state}{Additional reasoning helps when it can change the investigation. Interactive retrieval and final review let the system replace a weak source, recompute a quantity, and carry the correction into the final answer. With a fixed evidence snapshot, the same effort can reinforce an early error.}

Table~\ref{tab:main-evidence-interventions} identifies what changes that investigation. Relevant-file localization and the evidence map raise Core by 14.72 and 11.34 points, and they are the only interventions whose confidence intervals exclude zero. A curated tool subset adds 2.19 points, while the planning oracle slightly lowers performance. Renaming tools, normalizing units, and exposing optional meteorological tools also fail to reproduce the localization gain. More tools or more procedural guidance are therefore not enough. The agent must find the observations that determine the scientific object before it can reason reliably about that object.

The capability columns make the mechanism more precise. Relevant-file localization lifts spatiotemporal Core from 64.40 to 91.47, physical-mechanism Core from 65.15 to 80.61, and quantitative-calculation Core from 73.43 to 87.57. Multi-source performance changes little because the baseline already covers many sources; the gain comes from selecting and aligning the right ones. Missing, truncated, or conflicting evidence lowers overall quality, although the paired intervals remain wide. In this study, evidence localization is the clearest reproducible bottleneck, and downstream reasoning quality depends on whether that step succeeds.

\FloatBarrier

\subsection{Cross-benchmark transfer and benchmark adaptation}
\label{sec:cross-benchmark}

RQ3 separates the ability to interpret a supplied observation from the longer task of constructing and maintaining the evidence base.

\begin{figure}[H]
\centering
\includegraphics[width=0.94\linewidth]{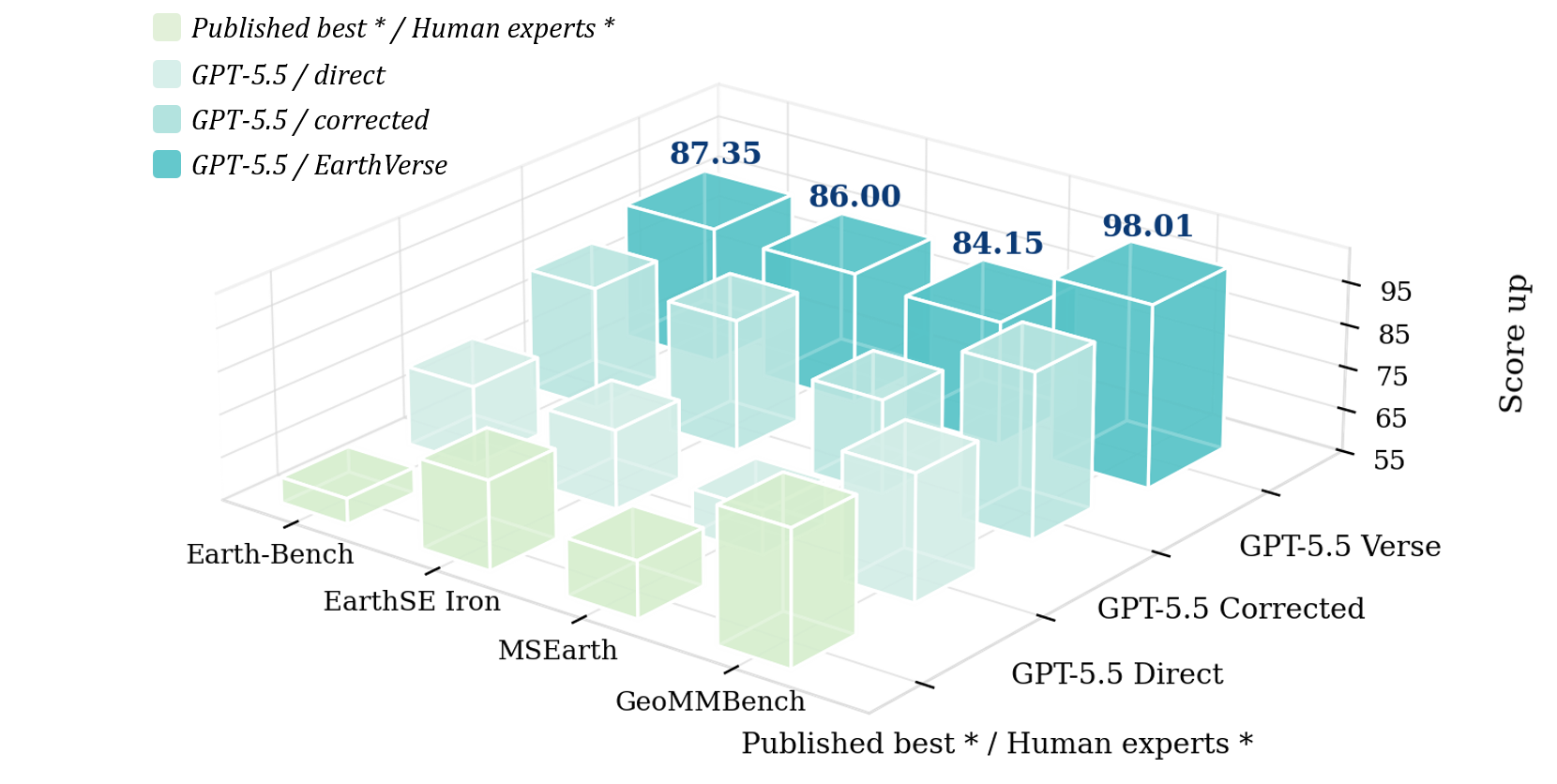}
\caption{\textbf{Cross-benchmark protocol sensitivity.} Benchmark-native scores across three GPT-5.5 stages and published references.}
\label{fig:cross-benchmark-results}
\end{figure}

We evaluate GPT-5.5 under the public protocols of Earth-Bench, EarthSE, MSEarth, and GeoMMBench. Each comparison begins with a direct response, then repairs clearly erroneous benchmark items or judgments, and finally applies the structured EarthVerse procedure. This procedure is answer-free. We summarize recurring question types across the benchmark and provide a general guide for identifying the relevant variables, choosing the necessary calculation, and checking whether the conclusion follows from the supplied observation. The guide contains no item answer, target label, or item-specific evidence. Figure~\ref{fig:cross-benchmark-results} summarizes a benchmark-native score for each stage and includes the relevant published reference~\citep{zhao2025msearth,xiao2026geommbench}. Because the benchmarks define different metrics, the figure supports comparisons within each benchmark, not rankings across the four groups. Complete native metrics and source-paper rows accompany the release.

The broad pattern is consistent, although the source of the gain differs by benchmark. On Earth-Bench, direct GPT-5.5 already clears the published line and improves further when the answer contract is made explicit. EarthSE is less uniform: multiple-choice and free-response results are already high, while fill-in accuracy rises from 20.29 to 61.90 after schema and answer-granularity corrections. MSEarth and GeoMMBench show the clearest recovery from interface mismatch. Direct GPT-5.5 trails the displayed reference, whereas the corrected and EarthVerse stages close or reverse the gap.

The improvement comes from making the problem legible to the model, not from revealing a solution or teaching new Earth-science knowledge during evaluation. Two changes are especially clear: EarthSE fill-in accuracy rises from 20.29 to 61.90, while GeoMMBench accuracy rises from 84.52 to 98.01, above both the published GeoMMAgent result of 88.40 and the reported human reference of 86.50. Under their native metrics, the tested frontier model handles these supplied-observation tasks well once the question structure is explicit.

These results concern tasks in which a passage, figure, or named observation has already been selected and the required output is clear. Current LLMs and agents are less reliable when they must decide which sources matter, align records with different support, verify calculations, preserve provenance, and revise downstream conclusions. This difference explains why adapted performance can be high while EarthVerse Strict@95 remains below 35\%.

\findingbox{finding:saturation}{Long-horizon evidence construction remains unresolved}{Answer-free structural guidance brings a frontier model to or above published reference levels on supplied-observation benchmarks, including the displayed human lines. EarthVerse Strict@95 remains below 35\% because the agent must construct, check, and revise the evidence base itself.}

\FloatBarrier

\section{Conclusion}

\textbf{RQ1: scientific reliability depends on the complete claim--evidence state.} Current systems often recover much of an analysis, yet one unsupported source, scale mismatch, or missing mechanism can change the result. Progress depends on preserving the links among claims, sources, scales, and calculations throughout the investigation.

\textbf{RQ2: reasoning helps when control keeps the claim--evidence state editable.} More effort over a fixed serialization does not recover interactive performance. Evidence localization and targeted review help when they let the system replace a weak source, recompute a quantity, and carry the correction into its final account. The controller must preserve claim--source bindings, check method applicability, and stop once the required evidence is complete.

\textbf{RQ3: interpreting evidence is easier than constructing it.} Within benchmarks that supply the observation, structured adaptation reaches or exceeds published reference lines. EarthVerse remains substantially harder because the system must decide what belongs in the analysis and keep the selected evidence compatible through calculation and synthesis.

EarthVerse contributes an executable benchmark and controlled diagnostics for multi-source hazard research. The leaderboard and interventions point to the same limitation: agents often recover valid local facts but fail to carry source changes and corrected quantities into the final account. By tracing that failure across evidence, computation, and reporting, EarthVerse connects model performance to the reliability of the resulting scientific analysis.

\section*{Author Contributions}
Zhiqing Cui led the project and was responsible for benchmark conceptualization, methodology, data curation, experimentation, and writing the original manuscript. 
Xinxiang Yin and Xinglang Zhang developed the figures and visual presentation. 
Yuanzhe Hu and Weidong Tang coordinated review and quality control.
Siru Zhong contributed to conceptual development and scientific discussion. 
Yihong Tang contributed methodological guidance, benchmark conceptualization, and scientific discussion. Dingyi Zhuang provided senior supervision, methodological guidance, and funding acquisition. Jinhua Zhao provided overall supervision, and project leadership. Yuxuan Liang, Weijia Li, Ming Jin, Shirui Pan, and Yuhao Kang contributed scientific discussion and critical review. All authors reviewed and approved the manuscript.

\bibliographystyle{unsrtnat}
\bibliography{references}

\clearpage
\appendix

\phantomsection
\section*{Appendix Guide}
\label{app:contents}

\noindent{\color{CaseDeep}\rule{\linewidth}{0.9pt}}\par\vspace{6pt}
\vspace{6pt}

\appendixguideentry{A}{sec:controlled-studies}{Controlled Studies}
{How reasoning, evidence access, memory, tools, and execution conditions change research quality.}
{Answer mode \textbullet\ reasoning effort \textbullet\ routing \textbullet\ context \textbullet\ robustness \textbullet\ scientific environments}{CaseBlue}

\appendixguideentry{B}{app:discussion}{Discussion, Broader Impact, and Limitations}
{What the results imply for scientific agents, disaster analysis, oversight, and responsible use.}
{Research position \textbullet\ broader impact \textbullet\ system design \textbullet\ limitations}{CaseTeal}

\appendixguideentry{C}{app:reproducibility}{Experimental Setup and Reproducibility}
{How models, modalities, inference, scoring, study manifests, and release decisions are controlled.}
{System roster \textbullet\ modality interface \textbullet\ main configuration \textbullet\ sampling \textbullet\ audit}{CaseGreen}

\appendixguideentry{D}{app:full-results}{Extended Diagnostics}
{Where capability, trajectory, error-path, resource, intervention, and tool-compliance analyses are reported.}
{Capability scores \textbullet\ error paths \textbullet\ costs \textbullet\ ablations \textbullet\ uncertainty}{CaseAmber}

\appendixguideentry{E}{app:casebook}{Prompts and Execution Contracts}
{What the solver, verifier, judge, and benchmark-construction reviewers are instructed to do.}
{Runtime visibility \textbullet\ solver contract \textbullet\ final review \textbullet\ judging \textbullet\ construction review}{CasePurple}

\appendixguideentry{F}{app:casebook-overview}{Representative Case Set}
{Which systems and tasks were selected to represent the benchmark's main scientific workflows and failure modes.}
{Five successful cases \textbullet\ one partial failure \textbullet\ system family \textbullet\ task scope \textbullet\ score}{CaseRose}

\appendixguideentry{G}{app:casebook-trajectories}{Detailed Research Trajectories}
{How selected systems discover evidence, calculate results, recover from errors, and finalize an answer.}
{Full task \textbullet\ ordered trace \textbullet\ computed ledger \textbullet\ evidence reconciliation \textbullet\ failure analysis}{CaseBlue}

\vfill
\noindent{\sffamily\small\bfseries\textcolor{CaseDeep}{Reading routes}}\par\vspace{3pt}
\noindent\begin{tabularx}{\linewidth}{@{}*{3}{>{\raggedright\arraybackslash}X}@{}}
\small\textbf{Experiments}\enspace\textcolor{CaseBlue}{A + D} &
\small\textbf{Reproduction}\enspace\textcolor{CaseGreen}{C + E} &
\small\textbf{Interpretation}\enspace\textcolor{CaseRose}{B + F + G}
\end{tabularx}
\par\vspace{2pt}
\noindent{\color{CaseDeep}\rule{\linewidth}{0.9pt}}

\clearpage

\section{Controlled Studies}
\label{sec:controlled-studies}

RQ2 asks when reasoning improves an investigation and which controls preserve or break scientific continuity. We therefore vary answer mode, reasoning effort, evidence localization, tool routing, context, stopping, and professional scientific execution. Unless noted, every condition uses GPT-5.5 under the protocol in Section~\ref{sec:model-settings}, with identical task manifests within each experiment.

\subsection{Reasoning effort and answer mode}
\label{sec:reasoning-ablation}

We cross four answer modes with five reasoning settings on a fixed, difficulty-stratified manifest. Compressed direct uses a bounded evidence summary, while all-text direct serializes every readable file. Both interactive conditions use the same EarthVerse harness; the review variant adds one evidence-and-calculation audit before finalization. The compact view below retains the outcome columns so that each reasoning curve can be compared without the denser interaction and resource diagnostics in Table~\ref{tab:main-reasoning-access}.

\begin{table}[!htbp]
\centering
\caption{\textbf{Outcome view of the reasoning sweep.} GPT-5.5; \best{bold blue} and shading mark the best setting in each access mode.}
\label{tab:reasoning-modes}
\tabfontcompact
\setlength{\tabcolsep}{3.1pt}
\renewcommand{\arraystretch}{1.04}
\makebox[\linewidth][c]{%
\begin{minipage}[t]{0.44\linewidth}
\sffamily\bfseries\textcolor{EarthBlue}{Compressed direct (selected evidence)}\par\vspace{1pt}
\begin{tabular}{@{}lrrrr@{}}
\toprule
\tblhead Effort & Answer & Process & Core & \makecell{Unit\\acc.} \\
\midrule
\keyrow none & \best{41.37} & \best{45.10} & \best{43.24} & \best{42.21} \\
low & 24.99 & 29.32 & 27.16 & 25.43 \\
medium & 22.88 & 25.69 & 24.29 & 23.26 \\
high & 24.17 & 28.17 & 26.17 & 24.28 \\
xhigh & 16.67 & 16.67 & 16.67 & 16.67 \\
\bottomrule
\end{tabular}
\end{minipage}
\hspace{13pt}
\begin{minipage}[t]{0.44\linewidth}
\sffamily\bfseries\textcolor{EarthBlue}{All-text direct (budgeted evidence)}\par\vspace{1pt}
\begin{tabular}{@{}lrrrr@{}}
\toprule
\tblhead Effort & Answer & Process & Core & \makecell{Unit\\acc.} \\
\midrule
\keyrow none & \best{27.66} & \best{31.22} & \best{29.44} & \best{28.27} \\
low & 13.26 & 16.20 & 14.73 & 13.36 \\
medium & 9.79 & 12.18 & 10.99 & 10.05 \\
high & 8.80 & 9.00 & 8.90 & 8.82 \\
xhigh & 12.00 & 15.00 & 13.50 & 12.00 \\
\bottomrule
\end{tabular}
\end{minipage}
}

\vspace{5pt}
\makebox[\linewidth][c]{%
\begin{minipage}[t]{0.44\linewidth}
\sffamily\bfseries\textcolor{EarthTeal}{EarthVerse interactive}\par\vspace{1pt}
\begin{tabular}{@{}lrrrr@{}}
\toprule
\tblhead Effort & Answer & Process & Core & \makecell{Unit\\acc.} \\
\midrule
none & 48.89 & 53.70 & 51.30 & 49.46 \\
low & 56.97 & 60.05 & 58.51 & 57.53 \\
medium & 61.45 & 64.80 & 63.13 & 62.35 \\
\keyrow high & \best{72.41} & \best{75.19} & \best{73.80} & \best{72.59} \\
xhigh & 69.66 & 67.50 & 68.58 & 69.83 \\
\bottomrule
\end{tabular}
\end{minipage}
\hspace{13pt}
\begin{minipage}[t]{0.44\linewidth}
\sffamily\bfseries\textcolor{EarthTeal}{EarthVerse interactive + review}\par\vspace{1pt}
\begin{tabular}{@{}lrrrr@{}}
\toprule
\tblhead Effort & Answer & Process & Core & \makecell{Unit\\acc.} \\
\midrule
none & 53.70 & 56.48 & 55.09 & 54.06 \\
low & 61.93 & 65.50 & 63.72 & 62.78 \\
medium & 67.16 & 69.22 & 68.19 & 67.91 \\
high & 69.82 & 74.71 & 72.27 & 70.73 \\
\keyrow xhigh & \best{84.36} & \best{88.21} & \best{86.29} & \best{84.40} \\
\bottomrule
\end{tabular}
\end{minipage}
}
\end{table}

\subsection{Tools, context, and interaction budget}
\label{sec:system-ablation}

On one fixed paired manifest, we vary catalog size, semantic routing, tool composition, context retention, and interaction budget. Every condition uses the default medium reasoning setting and the same answer and process scores. Appendix~\ref{app:system-diagnostics} reports trajectory diagnostics.

\begin{table}[!b]
\centering
\caption{\textbf{System ablations.} GPT-5.5 at medium reasoning; \best{bold blue} marks panel bests.}
\label{tab:system-design-ablation}
\label{tab:system-capability}
\tabfontdense
\setlength{\tabcolsep}{2.0pt}
\renewcommand{\arraystretch}{1.04}
\begin{adjustbox}{max width=0.96\textwidth,center}
\begin{tabular}{@{}l *{12}{r}@{}}
\toprule
& \multicolumn{3}{c}{Outcome} & \multicolumn{1}{c}{Coverage} & \multicolumn{7}{c}{Core by capability} & \multicolumn{1}{c}{Resource} \\
\cmidrule(lr){2-4}\cmidrule(lr){5-5}\cmidrule(lr){6-12}\cmidrule(l){13-13}
Condition & Answer & Process & Core & \makecell{Unit\\acc.} & \makecell{Physical\\mech.} & \makecell{Spatio-\\temporal} & \makecell{Quant.\\calc.} & \makecell{Multi-\\source} & \makecell{Causal\\chain} & Ranking & \makecell{RS /\\geo.} & Tokens \\
\midrule
\tblpanel{13}{(a) Tool catalog: nested subsets of 170 tools}
Fixed tools (6) & \best{64.34} & \best{65.62} & \best{64.98} & \best{64.69} & 63.27 & 66.22 & 70.34 & 97.50 & 63.35 & 51.78 & 58.12 & 229,557 \\
Fixed tools (10) & 60.22 & 57.10 & 58.66 & 60.35 & 56.74 & 58.36 & 61.04 & 95.22 & 55.77 & 58.32 & 61.25 & 194,047 \\
Fixed tools (20) & 60.58 & 59.38 & 59.98 & 60.93 & 58.13 & 70.50 & 61.67 & 95.00 & 54.90 & 56.50 & 55.62 & 232,912 \\
Fixed tools (40) & 58.55 & 61.00 & 59.77 & 59.43 & 58.16 & 47.81 & 62.03 & 90.45 & 63.63 & 59.57 & 54.12 & 232,993 \\
Fixed tools (70) & 58.76 & 61.25 & 60.01 & 59.19 & 58.54 & 55.68 & 63.89 & 87.95 & 60.14 & 54.21 & 54.62 & 233,759 \\
Fixed tools (100) & 63.96 & 63.12 & 63.54 & 64.11 & 61.81 & 58.93 & 65.84 & 96.37 & 65.79 & 60.82 & 51.00 & 278,936 \\
Full catalog (170) & 54.48 & 55.88 & 55.18 & 55.46 & 55.58 & 55.72 & 54.96 & 47.50 & 52.75 & 63.15 & 55.00 & 273,216 \\
\addlinespace[2pt]\midrule
\tblpanel{13}{(b) Routing and tool composition}
Top-10 semantic router & 64.12 & \best{66.25} & 65.19 & 64.55 & 64.37 & 60.64 & 67.93 & 80.74 & 67.50 & 57.20 & 59.17 & 216,227 \\
Top-20 semantic router & \best{66.90} & 65.88 & \best{66.39} & \best{67.14} & 65.15 & 64.40 & 73.43 & 90.00 & 65.89 & 50.84 & 53.94 & 180,010 \\
Workflow-profile router & 56.05 & 56.75 & 56.40 & 56.47 & 54.74 & 52.63 & 58.60 & 87.95 & 55.39 & 55.00 & 60.62 & 220,232 \\
Expert-only suite & 59.28 & 60.88 & 60.08 & 59.50 & 58.48 & 61.08 & 61.71 & 90.45 & 57.79 & 61.08 & 57.62 & 295,490 \\
General + expert suite & 59.07 & 63.00 & 61.04 & 59.95 & 59.25 & 66.01 & 63.21 & 95.00 & 61.34 & 49.27 & 51.38 & 195,092 \\
Atmospheric Python suite & 59.48 & 60.12 & 59.80 & 59.67 & 58.26 & 59.64 & 60.69 & 89.20 & 60.94 & 58.31 & 49.25 & 198,802 \\
Atmospheric Fortran suite & 57.46 & 57.88 & 57.67 & 57.51 & 55.81 & 51.05 & 61.19 & 92.95 & 57.35 & 53.88 & 60.75 & 205,262 \\
\addlinespace[2pt]\midrule
\tblpanel{13}{(c) Context retained before summarization}
Aggressive compression & 48.75 & 51.62 & 50.19 & 49.00 & 48.07 & 57.55 & 53.73 & 90.45 & 44.65 & 45.44 & 47.00 & 197,038 \\
Short retained context & 56.31 & 56.38 & 56.34 & 56.52 & 54.16 & 57.30 & 55.54 & 97.72 & 57.85 & 60.08 & 42.25 & 209,827 \\
Long retained context & \best{64.32} & \best{63.38} & \best{63.85} & \best{64.75} & 62.21 & 59.43 & 68.29 & 95.00 & 64.72 & 58.79 & 48.25 & 227,536 \\
Full retained history & 58.90 & 58.25 & 58.58 & 59.42 & 56.98 & 55.39 & 62.15 & 89.00 & 57.25 & 52.59 & 61.97 & 187,700 \\
\addlinespace[2pt]\midrule
\tblpanel{13}{(d) EarthVerse-harness turn budget under one wall clock}
12-turn budget & 44.84 & 49.50 & 47.17 & 44.95 & 44.65 & 45.36 & 46.27 & 95.00 & 47.73 & 57.50 & 35.13 & 106,263 \\
24-turn budget & 54.10 & 52.88 & 53.49 & 54.32 & 51.31 & 57.88 & 55.46 & 95.00 & 50.93 & 49.13 & 50.92 & 161,404 \\
96-turn budget & 59.72 & 60.12 & 59.92 & 60.16 & 57.94 & 62.76 & 60.53 & 97.50 & 62.10 & 54.08 & 44.38 & 182,798 \\
Wall-clock limited & \best{62.65} & \best{64.50} & \best{63.57} & \best{62.91} & 62.16 & 62.72 & 66.66 & 90.45 & 67.11 & 56.91 & 33.38 & 221,180 \\
\bottomrule
\end{tabular}
\end{adjustbox}
\end{table}

\subsubsection{Tool-catalog scale}
Nested fixed catalogs expose increasingly broad subsets of the same 170-tool registry (panel (a) of Table~\ref{tab:system-design-ablation}). The six-tool condition contains package discovery, reading, search, tabular inspection, Python execution, and finalization; each larger catalog adds domain tools without changing the task prompt.

The full catalog loses 9.80 Core points relative to the six-tool foundation while using 19\% more tokens. The decline is not monotonic, which rules out a simple capacity penalty. Extra tools add nearby analytical paths with different assumptions about variables, units, aggregation, and applicability. When the controller cannot retain why it chose one path, later outputs from another tool can enter the same ledger as if they were comparable. Catalog breadth then creates scientific ambiguity. Reporting a tool count without the navigation policy misses this failure mode.

\subsubsection{Semantic routing and scientific composition}
The routing study selects tools from task text and package metadata before interaction (panel (b) of Table~\ref{tab:system-design-ablation}). Top-$k$ routing exposes the highest-scoring individual tools, whereas workflow routing first selects a domain workflow. A separate composition study compares an expert-only suite, a mixed general/expert suite, and atmospheric Python and Fortran suites.

Top-20 routing recovers 11.21 Core points over the full catalog with 34\% fewer tokens. Its advantage comes from narrowing the next choice without deciding the investigation in advance. Workflow routing is more brittle because it can commit to a domain before the evidence warrants that commitment; the expert-only suite has the opposite problem, lacking ordinary file and calculation operations that connect specialist output to a scored claim. Routing should act as a revisable prior: preserve a small general core, introduce specialist methods when their required variables appear, and allow the evidence to overturn the initial route.

\subsubsection{Context retention}
We vary the amount of tool history preserved verbatim before older content is summarized (panel (c) of Table~\ref{tab:system-design-ablation}). Aggressive compression keeps a narrow recent window; short and long policies progressively retain more observations; full history disables summarization within the tested budget.

Long bounded retention gains 13.66 Core points over aggressive compression and 5.27 over full replay. Too little history forces the agent to rediscover files and often severs intermediate values from their provenance. Full replay retains rejected hypotheses and superseded numbers alongside active values, so discarded information can reappear in later calculations. Scientific memory therefore needs a compact claim--evidence state that binds claims to sources, units, temporal and spatial support, and unresolved conflicts while archiving replaced values.

\findingbox{finding:memory}{Scientific memory is selective}{Long bounded retention beats both aggressive compression and full replay. The useful state is a compact ledger of active claims, sources, units, and unresolved conflicts, not a verbatim transcript.}

\subsubsection{Interaction budget}
The final study changes only the EarthVerse-harness turn limit. The unbounded condition still uses the common wall-clock budget, so it tests productive stopping rather than unlimited execution.

Twelve turns usually end before discovery, calculation, and verification can all occur. Additional turns improve mean quality, but the gain depends on what those turns accomplish. Extra interaction helps when it closes a named evidence gap and hurts when the agent reopens a settled calculation or follows an unproductive branch. Stopping is therefore a coverage decision: the system should stop when required claims are supported, units and provenance are checked, and material contradictions have been resolved or disclosed.

\FloatBarrier

\subsection{Oracle access, discoverability, and evidence robustness}
\label{sec:oracle-robustness}

These conditions reuse one paired task set and the Top-20 router baseline. Table~\ref{tab:bottleneck-robustness} reports the outcomes; Appendix~\ref{app:oracle-diagnostics} retains trajectory and resource diagnostics.

\subsubsection{Oracle bottleneck localization}
Each oracle removes one information bottleneck while leaving the model, budget, tools, and scoring unchanged. It is a diagnostic upper bound rather than a proposed deployment setting, and none of the conditions reveals the target answer. The relevant-file oracle identifies only the package-relative files used by the reference analysis. The tool-subset oracle names useful tool classes; the planning oracle supplies a short answer-free plan; and the evidence-map oracle adds source roles and intermediate quantities without stating the conclusion.

\subsubsection{Tool naming and discoverability}
We hold tool implementations fixed and alter only how the catalog is presented (panel (b) of Table~\ref{tab:bottleneck-robustness}). Descriptive names add intended-use and exclusion cues; hierarchical discovery chooses a domain before exposing tools; dynamic descriptions retrieve candidate tools at each step; opaque names remove semantic hints.

\subsubsection{Evidence perturbation and robustness}
Perturbations preserve the question and answer schema while modifying the package evidence (panel (c) of Table~\ref{tab:bottleneck-robustness}): irrelevant files are added, a secondary source is made inconsistent, one supporting file is removed, long text is truncated, or units are changed while preserving physical equivalence.

\subsubsection{Compact meteorological access}
This probe tests access rather than compulsory execution. One condition exposes the same compact shortlist of meteorological operations to every task; the other uses task and package cues to expose at most three. The agent may ignore them in either condition. This separates tool availability from the professional-environment study below, where repeated specialist use is required.

\FloatBarrier

\subsection{Professional meteorological environments}
\label{sec:meteo}

This study asks a stricter question: what happens when a task must be completed through a named scientific environment? Each condition requires repeated use of one of 13 libraries or runtimes under the same protocol. Adapters expose only implemented operations and record whether calls succeed, allowing the results to separate useful scientific fit from technically successful but unsuitable method use.

\begin{table}[!htbp]
\centering
\caption{\textbf{Required scientific environments.} GPT-5.5; $\Delta$Core is relative to the Top-20 router.}
\label{tab:meteo-professional}
\tabfontregular
\renewcommand{\arraystretch}{1.04}
\setlength{\tabcolsep}{4.2pt}
\begin{adjustbox}{max width=0.90\textwidth,center}
\begin{tabular}{@{}l *{5}{r}@{}}
\toprule
& \multicolumn{3}{c}{Outcome} & \multicolumn{1}{c}{Change} & \multicolumn{1}{c}{Coverage} \\
\cmidrule(lr){2-4}\cmidrule(lr){5-5}\cmidrule(l){6-6}
Environment & Answer & Process & Core & \makecell{$\Delta$\\Core} & \makecell{Unit\\acc.} \\
\midrule
\keyrow Top-20 router (ref.) & 66.90 & 65.88 & 66.39 & 0.00 & 67.14 \\
\addlinespace[2pt]\midrule
\tblpanel{6}{(a) Python scientific libraries}
MetPy & 68.69 & 70.38 & 69.53 & \gain{+3.14} & 68.80 \\
xclim & 67.77 & 69.62 & 68.70 & \gain{+2.31} & 67.95 \\
thermofeel & 66.78 & 68.75 & 67.76 & \gain{+1.37} & 67.47 \\
PyET & 65.34 & 67.38 & 66.36 & \loss{$-$0.03} & 65.89 \\
climate-indices & 64.50 & 66.75 & 65.62 & \loss{$-$0.77} & 65.58 \\
xskillscore & 63.37 & 64.12 & 63.75 & \loss{$-$2.64} & 63.90 \\
pyextremes & 59.02 & 60.00 & 59.51 & \loss{$-$6.88} & 59.27 \\
\addlinespace[2pt]\midrule
\tblpanel{6}{(b) Compiled and domain-language environments}
Julia & \best{71.77} & 70.25 & \best{71.01} & \gain{\textbf{+4.62}} & \best{71.99} \\
R climate & 68.55 & \best{71.38} & 69.96 & \gain{+3.57} & 69.44 \\
CDO & 66.91 & 69.50 & 68.20 & \gain{+1.81} & 67.35 \\
NCO & 67.39 & 65.00 & 66.19 & \loss{$-$0.20} & 67.49 \\
NCL & 65.03 & 65.00 & 65.01 & \loss{$-$1.38} & 65.27 \\
Fortran & 64.46 & 64.25 & 64.35 & \loss{$-$2.04} & 64.52 \\
\bottomrule
\end{tabular}
\end{adjustbox}
\end{table}

Julia, R climate, MetPy, xclim, and CDO improve Core because their implemented operations match common package structures such as time series, gridded fields, and atmospheric profiles. Julia gives the largest gain (+4.62), but the ranking is less informative than the conditions under which the tools help. They compress many low-level numerical steps into an inspectable transformation without hiding the variables or aggregation. That advantage disappears when the package lacks the temporal support or paired variables assumed by the method.

Required use exposes a failure that optional-tool studies can miss. Pyextremes loses 6.88 Core points because many event packages do not contain a long, homogeneous series suitable for return-level estimation; forecast-verification methods face the same mismatch when paired forecasts and observations are absent. Under a hard requirement, the agent sometimes bends the available data toward the method instead of rejecting the method. A scientific interface needs an explicit refusal path, backed by minimum record length, sampling assumptions, missing-data rules, and a statement of what the output can support.

\findingbox{finding:applicability}{Execution success is not scientific validity}{A specialist method helps only when its required variables and sampling support exist. Otherwise, a successful call can produce a precise result that the package cannot justify.}

\FloatBarrier
\section{Discussion, Broader Impact, and Limitations}
\label{app:discussion}

\subsection{Research position}
\label{sec:findings}

EarthVerse evaluates a complete research loop over a bounded event archive. The agent must decide what counts as evidence, transform it, and revise an event-scale explanation while retaining enough provenance for another analyst to replay the work. This is more constrained than open-ended research, but it begins earlier than supplied-observation QA and asks more of the controller than a tool-execution benchmark.

Strong systems often interpret a selected observation correctly. Reliability falls when they must choose the record, its scale, the compatible variables, and the point at which the analysis is complete. Many failures originate in \emph{evidential binding}: a plausible value is attached to the wrong aggregation, time window, unit, or source role, and later reasoning treats the mismatch as established fact. This explains why stronger reasoning, more tools, and longer traces can all fail to help. They operate on the state the controller has preserved. If that state no longer records what a value means, additional reasoning only makes the wrong account more elaborate.

Complex Earth-system investigations rarely have a unique valid research path. EarthVerse therefore prescribes neither a tool order nor a reference trajectory. Different paths are valid when they recover the required windows, quantities, source roles, and mechanisms, and preserve support for the resulting claims~\citep{gu2024blade,majumder2025discoverybench}.

\subsection{Broader impact}

\paragraph{Scientific practice.}
Assistants can already reduce the mechanical burden of first-pass hazard analysis: inventorying an archive, locating candidate records, harmonizing units, repeating calculations, and surfacing source disagreement. The experiments also show where this assistance becomes risky. A fluent system may complete each local operation while carrying a scale mismatch through the full analysis. Experts still need to judge comparability, uncertainty, and consequences; package-local provenance gives them something concrete to inspect.

\paragraph{Reliability in high-stakes settings.}
High average unit accuracy can hide one missing source, time window, calculation, or mechanism. In disaster analysis, that single omission may change estimated severity, exposed population, or the causal account used to justify action. The benchmark does not certify operational readiness. It makes these omissions visible before a result is treated as operational evidence.

\paragraph{Access and evaluation practice.}
A local package reduces dependence on changing APIs and lets researchers inspect the same evidence. Evaluating frontier systems across hundreds of long trajectories is still expensive and may exclude smaller groups. Cost should be reported alongside quality, and releases should include compact diagnostic subsets, task-level scores, and traces. Subset results should not be presented as full-benchmark estimates.

\subsection{System design implications}
\label{sec:implications}

For Earth-system analysis, a trajectory is useful only when it preserves the support for each claim. A record of calls without source bindings cannot show whether the final explanation follows from the evidence.

\paragraph{Claim--evidence state.}
A transcript records chronology but does not identify which claims and values remain active. The controller needs a claim--source graph in which every derived value retains its source, unit, time window, transformation, and relation to competing explanations. Corrections should replace active values without erasing the history that explains the change.

\paragraph{Scientific operations.}
Scientific tools need to expose applicability as part of their output. Record length, sampling support, missing-data rules, and estimator assumptions determine whether a computed value can support the claim. Training that rewards successful calls alone encourages the model to force data into an available method; justified refusal should receive credit when the inputs are inadequate.

\paragraph{Verification and oversight.}
Final review should operate on claims rather than prose. For each required conclusion, the system should expose coverage, dimensional checks, source conflicts, and the mechanism tests it rejected. Experts can then inspect the points where judgment entered the analysis and decide whether the result is fit for use. A polished report without this record is not auditable.

\subsection{Limitations}
\label{sec:limitations}

\paragraph{Coverage and ecological validity.}
EarthVerse is a controlled approximation of research. Its 199 packages cover many natural hazards, but not the full range of Earth-system science, long-term attribution, or live forecasting. Package isolation improves reproducibility and blocks web leakage, at the cost of open-ended discovery and institutional data-quality judgments. Masked products also prevent tests of absolute geolocation.

\paragraph{Ground truth and judge validity.}
Structured ground truth necessarily chooses one decomposition of a scientific answer. Process scores also rely on an LLM judge that compares each trace with a reference solution. Separate answer and process blocks reduce score leakage but cannot remove judge error. Future releases should measure agreement with domain experts on a stratified subset and retain criterion-level decisions.

\paragraph{Data provenance and maintenance.}
The assembled event packages, task formulations, answer units, and derived targets were created for EarthVerse and were not available online before release, which reduces the chance of exact benchmark memorization. Some underlying historical events and source documents may still have appeared in pretraining. Package isolation, hidden scoring artifacts, and package-specific calculations make shortcut answering harder.

\paragraph{Statistical scope of controlled studies.}
The controlled studies use paired subsets, and several intervals in Appendix~\ref{app:oracle-diagnostics} include zero. Large contrasts, such as the relevant-file oracle, support architectural hypotheses. Smaller changes need replication across more packages and backbones. Cross-benchmark revision measures recoverability under feedback, not zero-shot superiority.

\FloatBarrier
\section{Experimental Setup and Reproducibility}
\label{app:reproducibility}

This section collects implementation details that are necessary for reproduction but not for interpreting the main findings. It records the evaluated systems, the common evidence interface, the leaderboard configuration, controlled-study manifests, scoring isolation, and retained audit artifacts.

\subsection{Systems, controllers, and model modality}
\label{app:model-interface}

The leaderboard contains 25 complete systems. Hosted endpoints include the OpenAI GPT-4o, GPT-5.5, and GPT-5.6 tiers~\citep{openai2024gpt4o,openai2026gpt55,openai2026gpt56}; Anthropic Claude tiers~\citep{anthropic2026systemcards}; Gemini~\citep{google2026gemini31}; MiniMax M2.7~\citep{minimax2026m27}; Seed 2.0~\citep{bytedance2026seed2}; Qwen3.7~\citep{qwen2026qwen37}; Tencent Hy3~\citep{tencent2026hy3}; DeepSeek-V4~\citep{deepseek2026v4}; Kimi K2.5~\citep{kimi2026k25}; and GLM-5.2~\citep{zai2026glm52}. The open-weight group includes Qwen3 at 4B and 235B scales~\citep{yang2025qwen3}, Nemotron Nano~\citep{nvidia2025nemotron}, MiroThinker 8B~\citep{miromind2025mirothinker}, GeohazardGPT, GeoGPT, and Intern-S1-mini~\citep{ge2025geohazardgpt,zhang2024geogpt,bai2025intern}.

Most systems run in the EarthVerse harness, which manages package discovery, tool use, context, recovery, and finalization. GeoMMAgent uses its adapted five-stage coordinator with GPT-5.5 as the worker model in every stage; OpenResearcher retains its research-loop controller~\citep{xiao2026geommbench,li2026openresearcher}. These rows therefore describe a model--controller system, not a bare backbone.

The roster mixes text-only models with natively multimodal endpoints. Direct image interpretation is required by only a small fraction of EarthVerse tasks; most remote-sensing evidence is represented by product metadata, raster statistics, change indices, or other auditable tool outputs. Raw pixels are not selectively sent to models with vision support. This keeps the evidence interface comparable, although the resulting scores should not be read as a standalone test of visual perception.

\subsection{Shared experimental configuration}
\label{app:inference-config}

Every run exposes a scientific question and one evidence root while keeping the solution, rubric, ground truth, scoring code, prior outputs, and web evidence hidden. Reads, writes, and Python execution stay inside the active evidence collection. Reports and tables are returned as local text or structured summaries; image and raster operations return recorded metadata and measurements.

The EarthVerse harness is the common controller for the standard model rows. To study the attainable system-level ceiling, we also evaluate the GPT-5.6 family in Codex and Claude Fable 5 in Claude Code, the strongest model-native harnesses available for those systems. GeoMMAgent pairs GPT-5.5 with its five-stage coordinator, while OpenResearcher keeps its own controller. The Harness column in Table~\ref{tab:main-results} makes these differences explicit.

The common execution limit is 48 turns, 1,680 seconds for research, and 1,800 seconds overall. The controller retains 16 recent messages and a 24,000-character research state; individual tool observations are capped at 12,000 characters. Package-scoped Python has a 60-second call limit. Up to two schema repairs are allowed, with 120 seconds reserved for finalization and a separate 120-second final-call timeout. We retain provider-native reasoning and generation controls when available, use endpoint defaults otherwise, and do not impose one temperature across providers. The judge is a fixed \texttt{gpt-4.1} endpoint, and randomized procedures use seed 2026. Endpoint access dates and local model revisions are retained in the release registry.

\subsection{Package evidence sources and access interfaces}
\label{app:package-sources}

The 199 package manifests contain 6,397 provenance records, 835 raw \texttt{source\_name} labels, and 6,709 local files from 107 external host domains. Many raw labels are event-specific filenames, response formats, endpoint variants, or repeated queries to the same service rather than distinct sources. We therefore group aliases at the product/API-family level while keeping distinct providers, sensors, product versions, and access semantics separate. This yields 80 external source and access families. We also list nine package-local derivative families used directly in computation; they do not count as additional upstream sources.

\begin{sourcepanel}{Distinct source and access families across 199 event packages}
\sourcegroup{CaseBlue}{Event discovery, reports, and authoritative anchors \hfill 29 families}
\sourceentry{GDACS}{event-list REST/GeoJSON API; RSS archive}{Cross-hazard event discovery, dates, alert levels, and identifiers.}
\sourceentry{NASA EONET v3}{REST search by date, category, or keyword}{Wildfire, flood, storm, and general natural-event catalog records.}
\sourceentry{ReliefWeb}{public update and report search}{Humanitarian situation reports, response updates, and impact narratives.}
\sourceentry{HDX}{CKAN \texttt{package\_search} API}{Candidate humanitarian datasets and country- or event-specific impact layers.}
\sourceentry{GDELT}{event-window article search}{Secondary news discovery used only when the package retains the result.}
\sourceentry{Wikipedia}{REST page-summary and event-search interfaces}{Event-name cross-checks and compact encyclopedic context.}
\sourceentry{Wikidata}{entity-search API}{Structured event and place identifiers used for identity checks.}
\sourceentry{NASA Earth Observatory}{event articles and image pages}{Locked event anchors and satellite-based physical narratives.}
\sourceentry{NASA Scientific Visualization Studio}{visualization pages and media API}{Official imagery, animations, and explanatory visual products.}
\sourceentry{NASA Disasters and Applied Sciences}{activation and event pages}{Satellite-based disaster-response context and mapped observations.}
\sourceentry{World Meteorological Organization}{reports, topic pages, and bulletins}{Authoritative global weather, climate, and extreme-event summaries.}
\sourceentry{World Meteorological Centre Beijing}{event reports and operational bulletins}{Regional meteorological summaries and event chronology.}
\sourceentry{NOAA Climate.gov}{event trackers and climate reports}{Reviewed US climate-event narratives and physical context.}
\sourceentry{US National Weather Service and WPC}{forecast discussions, storm summaries, and precipitation pages}{Operational weather chronology, rainfall totals, and warning context.}
\sourceentry{NOAA NESDIS and SSD}{satellite product and event pages}{Operational satellite imagery and environmental monitoring context.}
\sourceentry{Copernicus EMS}{activation and rapid-mapping archives}{Event confirmation and links to emergency-mapping products.}
\sourceentry{UN OCHA}{situation reports and humanitarian updates}{Reported impacts, needs, access constraints, and response activity.}
\sourceentry{UNOSAT}{satellite-assessment reports and maps}{Mapped damage, flood, and exposure evidence.}
\sourceentry{UN country teams and UN Geneva}{country and event updates}{Local consequence and response narratives retained as event evidence.}
\sourceentry{IFRC}{emergency appeals and operational updates}{Humanitarian consequences, needs, and Red Cross response records.}
\sourceentry{AHA Centre}{regional disaster updates and reports}{ASEAN event impacts and coordinated response context.}
\sourceentry{World Health Organization}{health-emergency reports}{Mortality, morbidity, service disruption, and public-health consequences.}
\sourceentry{UNICEF}{humanitarian situation reports}{Impacts on children, water, health, education, and relief delivery.}
\sourceentry{FAO}{food, agriculture, and livelihood reports}{Crop, livestock, food-security, and rural-impact evidence.}
\sourceentry{World Bank}{assessment and recovery documents}{Infrastructure, economic loss, and reconstruction context.}
\sourceentry{World Weather Attribution}{event-study pages and reports}{Attribution context and documented physical or societal drivers.}
\sourceentry{South Asian operational agencies}{IMD/RSMC New Delhi, DHM Nepal, and DMC Sri Lanka}{Official cyclone, rainfall, flood, and warning records.}
\sourceentry{National meteorological services}{BoM, HKO, UK Met Office, MetService, and SAWS}{National weather reports, warnings, and climate summaries.}
\sourceentry{Scientific institutes}{ESSL, ICIMOD, NGI/NGU, ICL, OSTI}{Hail, glacier, landslide, heat, and other specialist event accounts.}

\sourcegroup{CaseTeal}{Meteorology, climate, hydrology, ocean, and air quality \hfill 13 families}
\sourceentry[CaseTeal]{ERA5-Land}{hourly reanalysis; package-local aggregates/stacks}{Temperature, precipitation, wind, and land-surface background.}
\sourceentry[CaseTeal]{CHIRPS v2}{daily precipitation archive}{Event accumulation, wet spells, climatology, and rainfall anomalies.}
\sourceentry[CaseTeal]{NASA GPM IMERG V07}{Earthdata CMR discovery and precipitation granules}{Satellite precipitation totals, timing, and spatial structure.}
\sourceentry[CaseTeal]{Open-Meteo Archive}{historical hourly/daily REST API}{Point temperature, humidity, apparent temperature, rain, snow, and wind.}
\sourceentry[CaseTeal]{Open-Meteo Air Quality}{historical air-quality REST API}{PM$_{2.5}$, PM$_{10}$, aerosol, and related atmospheric fields.}
\sourceentry[CaseTeal]{NASA POWER}{daily point REST API}{Temperature, corrected precipitation, and 10-m wind cross-checks.}
\sourceentry[CaseTeal]{NOAA PSL climate indices}{download tables for ONI, SOI, and related modes}{Large-scale climate-state and teleconnection context.}
\sourceentry[CaseTeal]{NOAA CPC ONI}{ASCII and tabular index archive}{Operational Oceanic Ni\~no Index values and episode timing.}
\sourceentry[CaseTeal]{IRI and Australian BoM}{ENSO monitoring archives}{Independent ENSO phase and coupled-ocean background.}
\sourceentry[CaseTeal]{GPCC}{precipitation-anomaly products}{Gauge-based anomaly context alongside CHIRPS.}
\sourceentry[CaseTeal]{NOAA OISST v2 high resolution}{PSL THREDDS/ERDDAP catalog}{Sea-surface temperature and marine-heatwave context.}
\sourceentry[CaseTeal]{Indian Ocean Dipole index}{DMI archive}{Indian Ocean climate-mode state for regional events.}
\sourceentry[CaseTeal]{NOAA ocean services}{tide gauges and Ocean Prediction Center analyses}{Coastal water levels, surge context, and marine storm structure.}

\sourcegroup{CasePurple}{Remote sensing and geospatial context \hfill 12 families}
\sourceentry[CasePurple]{NASA Worldview/GIBS}{snapshot API and MODIS/VIIRS browse imagery}{Pre-event and in-event true-color or thematic visual context.}
\sourceentry[CasePurple]{MODIS/VIIRS vegetation products}{Worldview or archive previews}{Vegetation condition and drought-stress context.}
\sourceentry[CasePurple]{MODIS/VIIRS MAIAC AOD}{aerosol and smoke previews}{Smoke transport and aerosol loading context.}
\sourceentry[CasePurple]{Sentinel-1 GRD}{VV pre/post imagery and change products}{Flood, deformation, and surface-change comparison.}
\sourceentry[CasePurple]{Sentinel-2 SR Harmonized}{surface reflectance and dNBR derivatives}{Burn severity and optical before/after consistency.}
\sourceentry[CasePurple]{Google Satellite Embedding / AlphaEarth}{annual 64-band embeddings}{Annual state vectors, cosine-change statistics, and change rasters.}
\sourceentry[CasePurple]{Copernicus EMS Rapid Mapping}{EMSR/GFM maps and archive products}{Mapped flood, fire, earthquake, and exposure footprints.}
\sourceentry[CasePurple]{Microsoft Planetary Computer}{STAC metadata search}{AOI/date discovery for registered Earth-observation assets.}
\sourceentry[CasePurple]{Natural Earth}{Admin-0 and physical basemap files}{Low-resolution geopolitical and physical reference layers.}
\sourceentry[CasePurple]{geoBoundaries}{administrative-boundary API}{Open administrative geometries for package AOIs.}
\sourceentry[CasePurple]{GHSL}{built-up and settlement layers}{Urban extent and settlement structure.}
\sourceentry[CasePurple]{Copernicus DEM and SRTM}{elevation and derived slope}{Terrain constraints for flood, landslide, and glacier analyses.}

\sourcegroup{CaseAmber}{Exposure, infrastructure, and human impact \hfill 9 families}
\sourceentry[CaseAmber]{WorldPop GP 100 m}{population grids and AOI summaries}{Population exposure, density, and resampled 1-km context.}
\sourceentry[CaseAmber]{OpenStreetMap Nominatim}{geocoding REST API}{Event location, fallback coordinates, and AOI checks.}
\sourceentry[CaseAmber]{OpenStreetMap Overpass}{bounded feature-query API}{Roads, buildings, services, shelters, and critical amenities.}
\sourceentry[CaseAmber]{OpenStreetMap Geofabrik}{regional extracts}{Broader road and infrastructure context where bounded queries are insufficient.}
\sourceentry[CaseAmber]{FEWS NET and IPC}{food-security products and reports}{Drought exposure, food-security phases, and response pressure.}
\sourceentry[CaseAmber]{NYC Open Data}{Socrata 311 service-request API}{Flood complaints and city-service stress during urban rainfall.}
\sourceentry[CaseAmber]{Harris County and Houston open data}{flood-warning and 311 interfaces}{Gauge conditions, complaints, and local flood impacts.}
\sourceentry[CaseAmber]{Hong Kong public feeds}{warnings, incident, and transport records}{Local disruption and response evidence for typhoon events.}
\sourceentry[CaseAmber]{Brazil CEMADEN and civil defence}{alerts and municipal reports}{Rainfall-triggered hazard warnings and local consequences.}

\sourcegroup{CaseRose}{Hazard-specific scientific and operational sources \hfill 17 families}
\sourceentry[CaseRose]{USGS ComCat}{earthquake-event REST API}{Origin time, magnitude, depth, location, and event products.}
\sourceentry[CaseRose]{USGS ShakeMap and PAGER}{event-product feeds}{Instrumental intensity, shaking footprint, and loss context.}
\sourceentry[CaseRose]{USGS Did You Feel It?}{community-intensity product}{Reported shaking distribution and populated-area impacts.}
\sourceentry[CaseRose]{NOAA NCEI/WDS}{Global Historical Tsunami Database}{Tsunami source, run-up, and historical event records.}
\sourceentry[CaseRose]{NOAA NHC}{Tropical Cyclone Reports archive}{Track, intensity, rainfall, surge, and impact summaries.}
\sourceentry[CaseRose]{NOAA HRD H*Wind}{surface-wind analysis archive}{Storm wind structure and peak-wind calibration.}
\sourceentry[CaseRose]{NOAA Coral Reef Watch}{5-km thermal-stress products}{HotSpot, Degree Heating Week, and bleaching-risk context.}
\sourceentry[CaseRose]{Copernicus GloFAS/GFM/EFAS}{flood-monitoring and event products}{River-flood extent, timing, and European flood context.}
\sourceentry[CaseRose]{Dartmouth Flood Observatory}{Global Flood Monitor/archive}{Independent flood occurrence and extent context.}
\sourceentry[CaseRose]{Smithsonian GVP}{volcano reports and event pages}{Eruption chronology, volcano status, and physical interpretation.}
\sourceentry[CaseRose]{Volcanic Ash Advisory Centres}{operational ash advisories}{Ash extent, altitude, movement, and aviation context.}
\sourceentry[CaseRose]{GLIMS and Randolph Glacier Inventory}{glacier outlines and catalog interfaces}{Glacier geometry and inventory context.}
\sourceentry[CaseRose]{NSIDC and ICIMOD}{glacier and cryosphere reports}{Regional ice change and glacier-hazard evidence.}
\sourceentry[CaseRose]{NASA Global Landslide Catalog / COOLR}{landslide catalog search}{Occurrence records and landslide-event cross-checks.}
\sourceentry[CaseRose]{USGS landslide releases}{event pages, publications, and data releases}{Site evidence for landslides and landslide-generated tsunamis.}
\sourceentry[CaseRose]{ESSL and IEM/NWS archives}{severe-weather summaries and Local Storm Reports}{Hail, wind, tornado, and convective-event observations.}
\sourceentry[CaseRose]{Australian fire and smoke products}{official extent and smoke records}{Fire perimeter, smoke, and air-quality context.}

\sourcegroup{CaseGreen}{Package-local derivatives retained for computation \hfill 9 derived families}
\sourceentry[CaseGreen]{Compact event AOI}{event-bounding-box derivative}{Common spatial support for package layers.}
\sourceentry[CaseGreen]{ERA5-Land aggregate and stack}{hourly statistics and raster bundle}{Compact numerical access without replacing the upstream provenance.}
\sourceentry[CaseGreen]{GPM and CHIRPS event accumulations}{derived totals and AOI rasters}{Comparable precipitation ledgers across satellite and gauge products.}
\sourceentry[CaseGreen]{WorldPop summaries}{AOI sum and 1-km resample}{Deterministic population exposure inputs.}
\sourceentry[CaseGreen]{Sentinel-1 change}{pre/post statistics and raster}{Auditable radar change metrics.}
\sourceentry[CaseGreen]{Sentinel-2 dNBR}{summary and raster}{Auditable burn-severity calculation.}
\sourceentry[CaseGreen]{Satellite-embedding change}{annual statistics and raster}{Normalized annual surface-change evidence.}
\sourceentry[CaseGreen]{Bounded OSM slice}{package-local JSON}{Stable infrastructure and service counts.}
\sourceentry[CaseGreen]{Locked event anchor}{package-local report copy}{Fixed event identity, time window, and provenance for replay.}
\end{sourcepanel}

The inventory is exhaustive at this level of grouping, although each event uses only a hazard- and availability-specific subset. The 80 external rows cover distinguishable upstream sources and access methods; the nine derived rows record transformations reused across packages. Per-package manifests retain the raw label, exact URL, acquisition status, package-relative output, byte count, and checksum for record-level provenance.

\subsection{Scoring, isolation, and audit protocol}
\label{app:scoring-semantics}
\label{app:reproducibility-boundaries}

The judge receives the task, structured ground truth, final answer, reference process, rubric, and trajectory. It scores required answer units separately from evidence selection, calculation, source reconciliation, and mechanism analysis. Combined Core is the mean of these blocks, and Strict@95 records near-complete tasks. Capability columns use overlapping expert tags as diagnostic slices, not additive categories. Controlled comparisons use paired manifests and task-level bootstrap resampling.

Solvers may list, read, search, and compute only within the active event package. The reference solution, rubric, structured ground truth, and ground-truth program remain hidden. Tools record each path and operation before shortening observations, which preserves provenance through context compression. Finalization receives the accumulated research state and required schema but no answer-bearing hint. Released traces cannot reconstruct provider-hidden reasoning, proprietary routing, or undocumented cache accounting.

\subsection{Controlled-study sampling design}
\label{app:sampling-design}

The reasoning-effort study uses a fixed manifest containing historically difficult tasks and a fixed-seed random draw from the rest of the release, preferring distinct events. Tool, routing, context, and interaction studies use a smaller paired manifest drawn from it. Oracle and robustness conditions reuse the same task IDs and Top-20 router outputs. Paired 95\% intervals use 5,000 bootstrap resamples of task-level score differences with seed 2026.

Professional-environment conditions use GPT-5.5 with the default medium reasoning setting and 48 turns. Core libraries retain the paired manifest; input-dependent environments draw a difficulty-balanced compatible manifest with seed 2026. Their compliance contract requires at least two successful, distinct specialist calls, targets three, and counts at most four. These paired ablations are diagnostic and should not be read as exact full-leaderboard effects.

\paragraph{Artifacts retained for audit.}
\begin{itemize}[leftmargin=*,itemsep=2pt,topsep=3pt]
\item \textbf{Task identity.} Version, manifest, package checksum, and question reproduce the scientific input.
\item \textbf{System identity.} Endpoint revision, controller, tool profile, and context policy reproduce the executable system.
\item \textbf{Research trace.} Actions, arguments, observations, errors, and timing reconstruct the investigation, including failed calls.
\item \textbf{Submission.} Raw text, parsed object, repair attempts, and final status keep malformed or missing answers in the denominator.
\item \textbf{Reference and judgment.} Ground truth, solution, computation, rubric, judge prompt, and criterion decisions support answer-unit and aggregate recomputation.
\item \textbf{Paired comparison.} Task IDs, intervention, seed, and bootstrap resamples identify the controlled change.
\end{itemize}

\subsection{Human evaluation and agent-assisted review}
\label{app:human-review}

Human review begins when a task is conceived and continues through release. The team included six doctoral-level experts in Earth science and twenty undergraduates in related fields. Doctoral reviewers set the scientific direction, checked physical validity, and approved the final scoring specification. Undergraduate reviewers followed a common protocol for source and metadata checks, package replay, calculation verification, and secondary review of answer units. Reviewers contributed more than 50 hours per person on average.

\paragraph{Task direction and scoring checkpoints.}
For every task, a doctoral-level reviewer first analyzed the event and specified the scientific question, hazard process, spatial and temporal support, admissible evidence, required calculations, and answer fields. LLM agents then explored the package for plausible source combinations, alternative calculations, ambiguities, and likely failure modes. Reviewers used these passes to refine the reference analysis and define the final answer units and process checks. Agents proposed candidates; human experts decided which checks were scientifically necessary, supported by the package, and suitable for scoring.

\paragraph{Joint verification and correction.}
Every candidate task underwent agent replay followed by human inspection. Reviewers confirmed that package evidence could answer the public question, that \texttt{compute\_gt.py} regenerated the structured targets, and that the solution, answer units, and rubric referred to the same quantities and physical claims. Any mismatch returned the task to construction. A small number of sources had minor spatial offsets between the data footprint and the documented disaster location. Human experts removed the affected answer units rather than score a quantity the package could not support.

\paragraph{Iteration scale.}
For difficult cases, constructing and reviewing a single question consumed more than 20 million tokens across repeated agent passes. These passes covered evidence exploration, draft construction, calculation checks, replay, discrepancy analysis, and revision. Human reviewers used them to align the scientific question, evidence, expert solution, deterministic program, answer units, and process rubric before release.

\subsection{Benchmark construction and release gates}
\label{app:construction}

Each construction stage produces an auditable artifact and an explicit release decision. A failed gate returns the task to the responsible stage.

\paragraph{Seven release gates.}
\begin{enumerate}[leftmargin=*,itemsep=2pt,topsep=3pt]
\item \textbf{Event anchoring:} verify identity, hazard type, time window, spatial scale, and event--source agreement.
\item \textbf{Package assembly:} build a typed relative-path manifest and confirm that the required evidence roles are readable.
\item \textbf{Task authoring:} require package-local resolution without revealing a file path or answer-bearing value.
\item \textbf{Reference analysis:} align claims, units, mechanisms, and defensible alternatives; trace every load-bearing claim.
\item \textbf{Executable grounding:} regenerate structured ground truth deterministically and pass schema and tolerance checks.
\item \textbf{Rubric design:} map the task obligations to five to eight criteria totaling exactly 20 points.
\item \textbf{Final review:} align the question, solution, ground truth, program, and rubric before release approval.
\end{enumerate}

\FloatBarrier
\section{Extended Diagnostics}
\label{app:full-results}

Table~\ref{tab:main-capability} breaks Combined Core down by seven overlapping capability tags. Because a task may carry several tags, the columns are diagnostic views rather than additive partitions.

The strongest systems lead across nearly every tag, but their residual errors concentrate in the connections between results. Among the top five, causal-chain scores trail quantitative calculation by 4.9--7.9 points. A single defensible data path may be enough to compute a total or ratio. An evolving hazard account is harder to stabilize: rainfall, exposure, damage, and response evidence must refer to compatible windows and scales, and the proposed mechanism must survive their disagreement. GeoMMAgent breaks the otherwise stable ordering. Its short, structured search works well on tabular evidence but often bypasses raster intermediates, leaving remote sensing at 58.82. Capability scores thus capture the organization of an investigation as much as the knowledge of its backbone.

\subsection{Capability-conditioned performance}
\label{app:capability}
\begingroup
\tabfontregular
\setlength{\tabcolsep}{3.0pt}
\renewcommand{\arraystretch}{1.00}
\setlength{\LTleft}{\fill}
\setlength{\LTright}{\fill}
\begin{longtable}{@{}l *{7}{r}@{}}
\caption{\textbf{Core by capability.} Tags overlap; system groups follow Table~\ref{tab:main-results}. \best{Bold blue} marks column bests.}
\label{tab:main-capability}\\
\toprule
& \multicolumn{3}{c}{Earth-process analysis} & \multicolumn{4}{c}{Evidence synthesis} \\
\cmidrule(lr){2-4}\cmidrule(l){5-8}
System & \makecell{Physical\\mech.} & \makecell{Spatio-\\temporal} & \makecell{Quant.\\calc.} & \makecell{Multi-\\source} & \makecell{Causal\\chain} & Ranking & \makecell{RS /\\geo.} \\
\midrule
\endfirsthead
\multicolumn{8}{r}{\textit{Table~\thetable\ continued}}\\[-1pt]
\toprule
System & \makecell{Physical\\mech.} & \makecell{Spatio-\\temporal} & \makecell{Quant.\\calc.} & \makecell{Multi-\\source} & \makecell{Causal\\chain} & Ranking & \makecell{RS /\\geo.} \\
\midrule
\endhead
\midrule
\multicolumn{8}{r}{\textit{continued on next page}}\\
\endfoot
\bottomrule
\endlastfoot
\tblpanel{8}{Agent frameworks}
GeoMMAgent & 67.72 & 72.94 & 68.85 & 70.56 & 66.15 & 67.00 & 58.82 \\
OpenResearcher & 30.46 & 35.24 & 31.35 & 34.63 & 25.35 & 30.41 & 33.18 \\
\addlinespace[2pt]
\tblpanel{8}{Open-weight and Earth-specialized models}
Qwen3-235B & 31.90 & 40.35 & 34.03 & 35.51 & 29.00 & 29.29 & 32.32 \\
GeohazardGPT & 31.77 & 31.31 & 28.04 & 30.42 & 32.12 & 39.04 & 29.16 \\
Qwen3-4B & 24.63 & 22.71 & 24.70 & 24.32 & 25.89 & 24.06 & 32.61 \\
GeoGPT & 25.34 & 25.70 & 24.66 & 23.14 & 23.63 & 26.28 & 22.77 \\
MiroThinker 8B & 13.51 & 14.39 & 14.68 & 14.52 & 8.61 & 6.76 & 21.96 \\
Nemotron Nano & 11.79 & 14.41 & 12.30 & 17.89 & 12.42 & 13.26 & 15.48 \\
Intern-S1-mini & 12.04 & 10.70 & 11.99 & 10.45 & 10.57 & 13.21 & 16.50 \\
\addlinespace[2pt]
\tblpanel{8}{Hosted general-purpose systems}
Claude Fable 5 & \best{85.15} & \best{85.51} & \best{87.14} & \best{83.45} & \best{82.20} & \best{80.98} & \best{89.99} \\
GPT-5.6 Sol & 81.81 & 83.31 & 85.42 & 82.13 & 77.54 & 71.63 & 88.50 \\
GPT-5.6 Terra & 80.83 & 79.83 & 82.96 & 79.06 & 76.26 & 76.12 & 86.27 \\
GLM-5.2 & 80.25 & 77.30 & 80.24 & 76.85 & 80.08 & 78.34 & 79.27 \\
GPT-5.5 & 77.21 & 78.38 & 81.91 & 76.79 & 74.33 & 75.76 & 83.77 \\
Claude Sonnet 4.6 & 75.49 & 73.60 & 78.36 & 77.92 & 73.43 & 79.73 & 75.93 \\
Kimi K2.5 & 75.07 & 79.10 & 77.81 & 75.34 & 70.48 & 70.82 & 77.94 \\
DeepSeek V4 Flash & 73.63 & 76.33 & 75.16 & 72.59 & 68.91 & 72.33 & 77.75 \\
Tencent Hy3 & 72.67 & 75.92 & 73.71 & 70.32 & 69.72 & 70.23 & 79.41 \\
Claude Haiku 4.5 & 69.60 & 72.44 & 76.09 & 73.50 & 67.73 & 66.76 & 79.66 \\
Qwen 3.7 Plus & 70.41 & 72.40 & 71.13 & 69.27 & 67.19 & 68.08 & 78.26 \\
Doubao Seed 2.0 Mini & 63.24 & 66.43 & 62.67 & 59.30 & 60.42 & 63.51 & 70.67 \\
Gemini 3.1 Flash Lite & 60.18 & 61.40 & 60.38 & 56.99 & 56.37 & 58.35 & 68.49 \\
MiniMax M2.7 & 42.76 & 52.91 & 48.80 & 49.53 & 39.99 & 24.21 & 42.28 \\
GPT-4o & 39.97 & 38.86 & 38.77 & 34.93 & 37.56 & 43.70 & 48.30 \\
GPT-5.6 Luna & 34.74 & 42.91 & 38.58 & 36.48 & 31.77 & 21.17 & 38.21 \\
\end{longtable}
\endgroup

\subsection{Preference structure and diagnostic error paths}
\label{app:error-atlas}

\begin{figure}[H]
  \centering
  \vspace{-4pt}
  \includegraphics[width=0.74\textwidth]{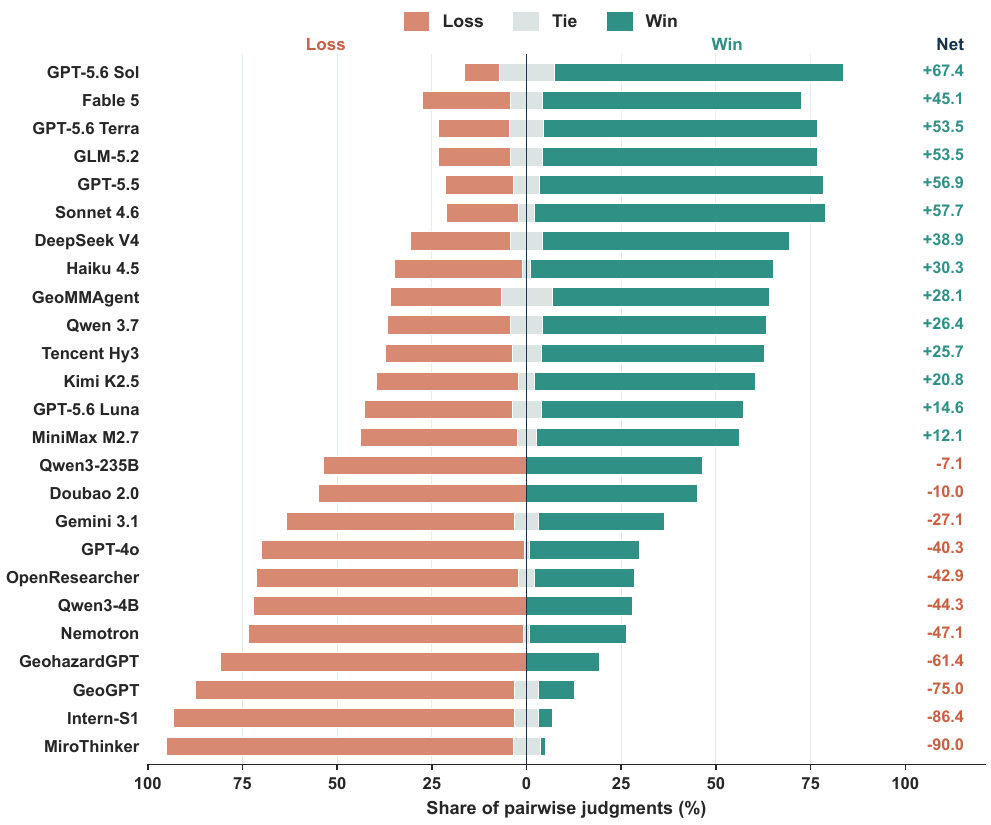}
  \vspace{-3pt}
  \caption{Pairwise win, tie, and loss shares. Net is win share minus loss share.}
  \label{fig:pairwise-preference}
\end{figure}

Pairwise judgments broadly recover the leaderboard order while exposing uncertainty hidden by Core alone. Figure~\ref{fig:pairwise-preference} separates wins, losses, and ties in opponent-balanced comparisons. Leaders win consistently, adjacent frontier systems remain close, and ties are uncommon. The drop beyond the leading group agrees with low Strict@95: partial success is common, while consistently complete research remains rare.

\positionstatement{Reliability in dynamic Earth systems depends on revising the claim--evidence state as observations change. A system must detect conflicts, isolate local errors, and carry corrections into derived quantities, causal interpretations, and the final answer. Systems differ not only in how often they err, but also in whether those errors are found, contained, and repaired. Longer traces, more tool calls, and higher execution success do not measure that capacity~\citep{chen2025scienceagentbench,du2025deepresearchbench}.}

\FloatBarrier

\Needspace{0.42\textheight}
\begin{wrapfigure}[15]{r}{0.54\textwidth}
  \centering
  \vspace{-2pt}
  \includegraphics[width=\linewidth]{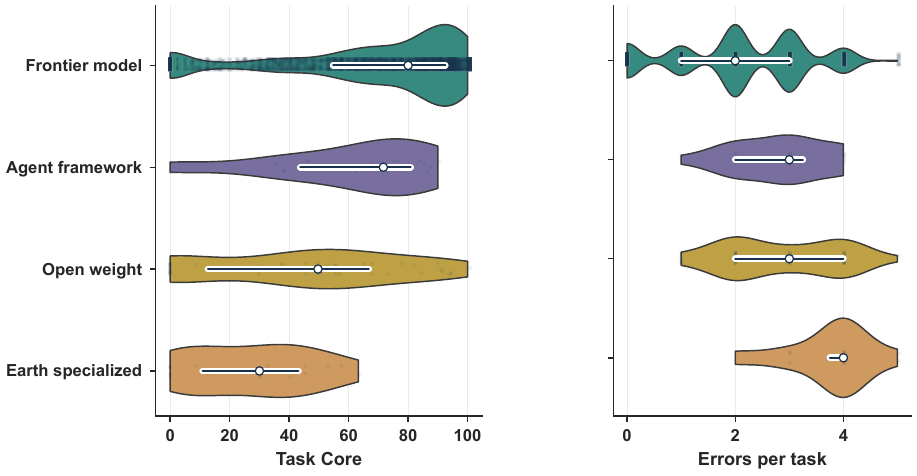}
  \caption{Task Core and diagnosed errors by system family.}
  \label{fig:outcome-distributions}
  \vspace{-5pt}
\end{wrapfigure}
Figure~\ref{fig:outcome-distributions} shows the same separation at task level. Frontier models have a dense high-score mode but retain a long tail toward zero. Agent frameworks cluster at moderately high scores, while open-weight and Earth-specialized systems shift downward and rarely approach complete solutions. Error counts overlap far more than Core scores. Strong systems still make mistakes; their advantage lies in catching or containing them before the final answer. Error frequency therefore describes the research process, while Core records the damage that remains.

\begin{figure}[!htbp]
  \centering
  \includegraphics[width=0.76\textwidth]{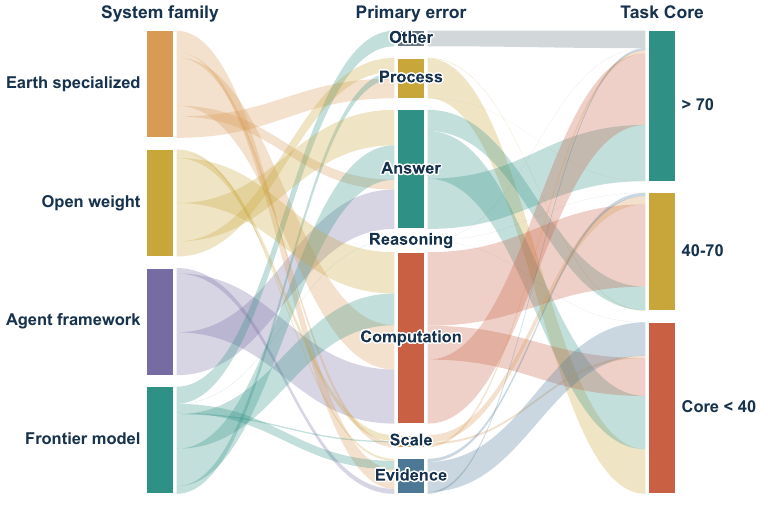}
  \caption{System family, primary error family, and task-level Core band.}
  \label{fig:failure-flow}
\end{figure}

The family-level flow in Figure~\ref{fig:failure-flow} sharpens this result. Computation and answer errors occur in every family, yet their consequences differ. Frontier systems more often keep affected tasks in the 40--70 or above-70 bands; open-weight and Earth-specialized systems more often fall below 40. Recovery, cross-source checking, and final synthesis determine whether a local mistake remains local or spreads through the answer.

\makeatletter\WF@finale\makeatother

\FloatBarrier

\begin{figure}[H]
  \centering
  \includegraphics[width=\textwidth]{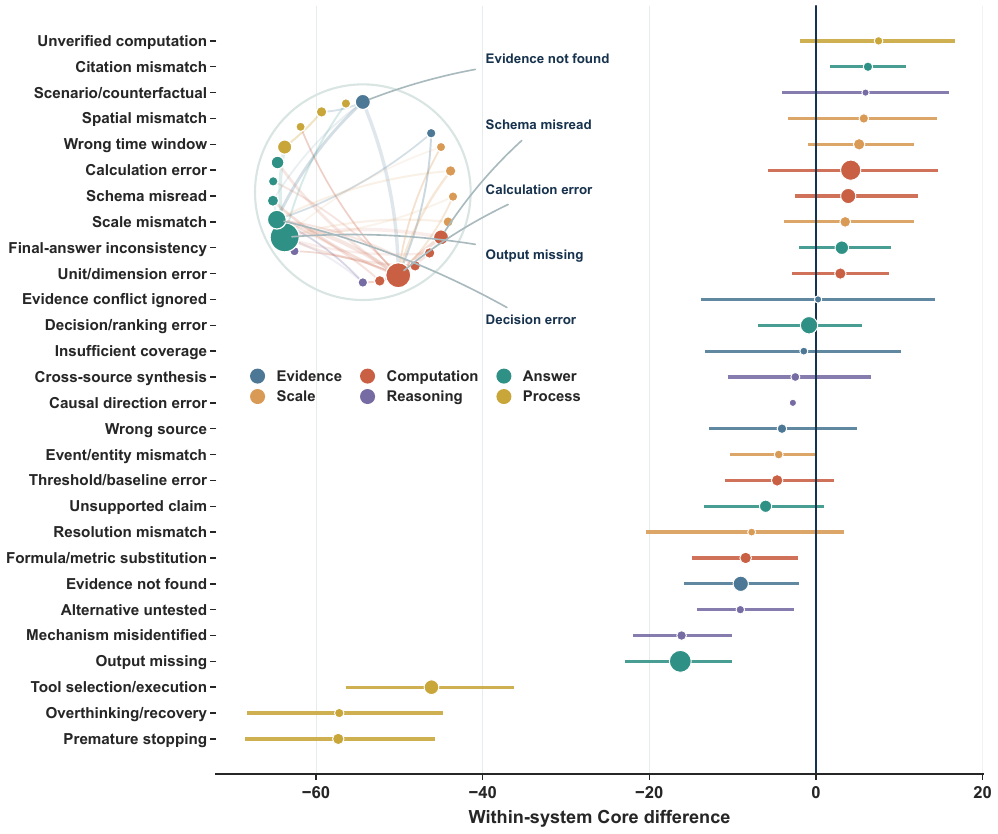}
  \caption{Within-system Core differences and the strongest error co-occurrences.}
  \label{fig:error-effects}
\end{figure}

\noindent Figure~\ref{fig:error-effects} examines the same pattern within systems. The largest Core losses accompany premature stopping, failed recovery, tool failure, and missing required outputs---failures that interrupt the research loop itself. Numerical and schema errors are common, but their average effect is smaller because some runs catch or absorb them before finalization. The inset connects missing evidence, output omissions, schema mistakes, and calculation errors into one cluster. These are often successive stages of a failed investigation, not independent defects.

\FloatBarrier
\subsection{Reasoning modes and resource accounting}
\label{app:reasoning-resources}

Tables~\ref{tab:reasoning-diagnostics}--\ref{tab:evaluation-cost} report trajectory, latency, token use, and evaluation cost. They separate interactive retrieval from direct serialization and model activity from successful evidence acquisition.

\subsubsection{Interactive trajectory diagnostics}
\label{app:reasoning-process}

Higher reasoning effort changes the treatment of evidence more than the amount collected. At xhigh, the EarthVerse harness opens about as many files as at medium effort and uses fewer rounds, yet takes almost twice as long. The additional work occurs between observations: the model compares interpretations, revisits calculations, and decides whether its account is complete. Explicit review intensifies this pattern. From harness-high to review-xhigh, output nearly doubles while file reads barely move. Some of that deliberation repairs omissions; some only restates a settled path. Rounds, tokens, and latency therefore measure different kinds of work, and none is a reliable stand-in for scientific progress.

\begin{table}[H]
\centering
\caption{\textbf{Interactive trajectory diagnostics.} Per-task means; call success is the share of tool calls that return normally.}
\label{tab:reasoning-diagnostics}
\tabfontcompact
\renewcommand{\arraystretch}{1.10}
\setlength{\tabcolsep}{1.1pt}

\begin{tabular}{@{}l l *{11}{r}@{}}
\toprule
& & \multicolumn{3}{c}{Execution} & \multicolumn{3}{c}{Evidence} & \multicolumn{5}{c}{Resources} \\
\cmidrule(lr){3-5}\cmidrule(lr){6-8}\cmidrule(l){9-13}
Mode & Effort & Rounds & \makecell{Tool\\calls} & \makecell{Success\\(\%)} & Reads & Files & \makecell{Evidence\\refs.} & \makecell{Input\\(k)} & \makecell{Output\\(k)} & \makecell{Total\\(k)} & \makecell{Cached\\(k)} & \makecell{Latency\\(s)} \\
\midrule
EarthVerse & none & 9.68 & 8.68 & 92.08 & 3.84 & 7.00 & 9.94 & 94.3 & 2.6 & 96.9 & 10.6 & 140 \\
& low & 14.70 & 13.70 & 93.67 & 5.64 & 9.44 & 14.70 & 157.4 & 5.1 & 162.4 & 15.4 & 259 \\
& medium & 17.20 & 16.20 & 95.39 & 6.84 & 10.71 & 18.12 & 198.6 & 7.9 & 206.4 & 17.9 & 424 \\
& high & 14.96 & 13.96 & 97.19 & 5.15 & 9.92 & 17.19 & 164.8 & 9.5 & 174.3 & 13.2 & 571 \\
& xhigh & 11.25 & 10.25 & 100.00 & 4.75 & 10.25 & 16.50 & 90.6 & 9.4 & 100.0 & 15.9 & 674 \\
\midrule
Review & none & 11.62 & 9.62 & 94.26 & 4.96 & 7.42 & 10.46 & 126.9 & 3.8 & 130.7 & 18.8 & 108 \\
& low & 18.18 & 16.18 & 94.94 & 6.84 & 10.60 & 17.02 & 219.0 & 8.0 & 227.0 & 25.1 & 253 \\
& medium & 20.81 & 18.81 & 94.78 & 6.56 & 11.88 & 21.02 & 253.8 & 12.5 & 266.3 & 25.2 & 552 \\
& high & 20.63 & 18.63 & 95.37 & 6.50 & 12.13 & 21.10 & 239.2 & 15.9 & 255.1 & 22.2 & 1,105 \\
& xhigh & 17.57 & 15.57 & 99.29 & 5.86 & 10.14 & 18.29 & 198.9 & 18.8 & 217.7 & 25.6 & 1,493 \\
\bottomrule
\end{tabular}%

\end{table}

\subsubsection{Direct package-serialization diagnostics}
\label{app:direct-resources}

Direct conditions receive a fixed package serialization before generation. Compressed direct includes more files in far fewer tokens than all-text, yet neither mode lets the model return to a source after discovering that a window, unit, or variable was wrong. That restriction changes the task. The model can reason about the supplied snapshot, but it cannot conduct the next measurement implied by its own reasoning. Larger inputs and longer outputs fail to rescue these baselines because the evidence schedule is fixed before the analysis begins. The missing operation is revision of the observation set, not another pass over the same text.

\begin{table}[H]
\centering
\caption{\textbf{Direct-serialization resources.} Direct modes inject a fixed file set before generation; values are per-task means.}
\label{tab:direct-resources}
\tabfontregular
\setlength{\tabcolsep}{3.5pt}
\renewcommand{\arraystretch}{1.10}
\begin{tabular}{@{}l l r r r r r r@{}}
\toprule
\tblhead
Mode & Effort & \makecell{Injected\\files} & \makecell{Evidence\\refs.} & Input & Output & Total & \makecell{Latency\\(s)} \\
\midrule
Compressed & none & 17.92 & 24.56 & 13,016 & 878 & 13,894 & 55.61 \\
& low & 17.92 & 25.31 & 13,033 & 1,857 & 14,890 & 120.45 \\
& medium & 17.92 & 25.72 & 12,838 & 3,004 & 15,842 & 152.36 \\
& high & 17.93 & 26.73 & 12,958 & 3,630 & 16,588 & 71.47 \\
& xhigh & 17.67 & 25.00 & 12,878 & 4,726 & 17,604 & 88.14 \\
\midrule
All text & none & 7.57 & 13.89 & 80,842 & 751 & 81,593 & 32.06 \\
& low & 7.68 & 14.00 & 80,886 & 1,854 & 82,739 & 44.10 \\
& medium & 7.46 & 13.90 & 80,941 & 2,557 & 83,498 & 61.02 \\
& high & 8.73 & 15.13 & 81,623 & 3,735 & 85,359 & 78.49 \\
& xhigh & 8.00 & 10.00 & 83,655 & 4,013 & 87,668 & 94.38 \\
\bottomrule
\end{tabular}
\end{table}

\subsubsection{Evaluation cost}
\label{app:cost-accounting}

Table~\ref{tab:evaluation-cost} reports the cost of the evaluated runs for models with a comparable USD list price.

\begin{table}[H]
\centering
\caption{\textbf{Evaluation cost.} Rows descend by total cost.}
\label{tab:evaluation-cost}
\fontsize{8.4pt}{9.5pt}\selectfont
\setlength{\tabcolsep}{6.0pt}
\renewcommand{\arraystretch}{1.16}
\begin{tabular}{@{}l r r r r@{}}
\toprule
\tblhead
System & Core & \makecell{Tokens/task\\(k)} & \makecell{Price\\(USD/M)} & \makecell{Cost\\(USD)} \\
\midrule
Claude Fable 5 & 84.97 & 161.6 & 50.00 & 3,271.57 \\
GPT-5.5 & 78.06 & 224.9 & 30.00 & 2,732.09 \\
GPT-5.6 Sol & 81.68 & 205.9 & 30.00 & 2,501.58 \\
Claude Sonnet 4.6 & 76.16 & 331.0 & 15.00 & 2,011.07 \\
GPT-5.6 Terra & 80.06 & 182.6 & 15.00 & 1,109.03 \\
Claude Haiku 4.5 & 72.23 & 477.0 & 5.00 & 966.02 \\
GPT-4o & 39.23 & 127.4 & 10.00 & 515.92 \\
GPT-5.6 Luna & 35.66 & 156.9 & 6.00 & 381.31 \\
DeepSeek V4 Flash & 73.54 & 389.7 & 0.28 & 44.19 \\
Gemini 3.1 Flash Lite & 59.66 & 70.1 & 1.50 & 42.61 \\
\bottomrule
\end{tabular}
\end{table}

Quality and cost do not form a simple ladder. The strongest frontier evaluations exceed \$2,500, while GPT-5.6 Terra retains about 98\% of GPT-5.6 Sol's Core at 44\% of its cost. The two Flash systems remain below \$45 but differ by almost 14 Core points. A long trace from a low-price model may still cost less than a short frontier run, so quality and cost should be reported together when planning replication.

\FloatBarrier
\subsection{Tool, memory, and interaction ablations}
\label{app:system-diagnostics}

Table~\ref{tab:system-design-ablation} reports outcome and capability scores for each ablation; Table~\ref{tab:system-explore} shows how the same runs searched, computed, and consumed context. Token use is descriptive and does not affect scoring.

\subsubsection{Research behavior and context consumption}

The trajectory diagnostics expose two different kinds of false productivity. Under aggressive compression, the agent lists files more than four times as often as with long retention and records more evidence references, yet receives the lowest process score. It is rebuilding context that the controller discarded, not broadening the analysis. The expert-only suite looks efficient for the opposite reason: it has the highest call success while inspecting the fewest distinct files. Its tools run cleanly, but the investigation is narrow. Activity counts become meaningful only when tied to state changes: a new source should resolve a claim, a computation should test a mechanism, and a repeated read should have a reason.

\begin{table}[!t]
\centering
\caption{\textbf{System-ablation trajectories.} Per-task means for the panels in Table~\ref{tab:system-design-ablation}.}
\label{tab:system-explore}
\tabfontdense
\setlength{\tabcolsep}{0.95pt}
\renewcommand{\arraystretch}{1.03}
\resizebox{\textwidth}{!}{%
\begin{tabular}{@{}l *{13}{r}@{}}
\toprule
& \multicolumn{3}{c}{Tools} & \multicolumn{2}{c}{Evidence} & \multicolumn{3}{c}{Actions} & Process & \multicolumn{4}{c}{Tokens} \\
\cmidrule(lr){2-4}\cmidrule(lr){5-6}\cmidrule(lr){7-9}\cmidrule(lr){10-10}\cmidrule(l){11-14}
Condition & \makecell{Calls\\OK} & \makecell{Calls\\failed} & \makecell{Success\\(\%)} & Files & \makecell{Evidence\\refs.} & Lists & Searches & Python & \makecell{Process\\(\%)} & Input & Output & Total & Cached \\
\midrule
\tblpanel{14}{(a) Tool-catalog scale}
Fixed tools (6) & 15.80 & 2.15 & 92.52 & 12.75 & 19.80 & 1.95 & 4.45 & 5.45 & 65.62 & 219,639 & 9,918 & 229,557 & 25,882 \\
Fixed tools (10) & 15.25 & 1.75 & 92.22 & 11.05 & 17.40 & 1.30 & 3.30 & 3.35 & 57.10 & 185,888 & 8,159 & 194,047 & 25,408 \\
Fixed tools (20) & 15.55 & 2.05 & 91.74 & 12.05 & 17.15 & 1.45 & 3.10 & 4.60 & 59.38 & 223,666 & 9,246 & 232,912 & 25,267 \\
Fixed tools (40) & 15.75 & 1.95 & 92.81 & 11.95 & 19.70 & 1.55 & 4.15 & 4.10 & 61.00 & 224,175 & 8,818 & 232,993 & 24,346 \\
Fixed tools (70) & 15.00 & 1.90 & 92.39 & 11.40 & 18.35 & 1.25 & 3.35 & 2.85 & 61.25 & 224,853 & 8,906 & 233,759 & 26,445 \\
Fixed tools (100) & 16.35 & 2.00 & 92.54 & 12.00 & 20.25 & 1.10 & 3.90 & 3.60 & 63.12 & 268,527 & 10,409 & 278,936 & 32,294 \\
Full tool catalog (170) & 13.90 & 1.50 & 95.04 & 9.55 & 13.15 & 1.05 & 1.85 & 2.55 & 55.88 & 264,983 & 8,233 & 273,216 & 33,331 \\
\addlinespace[2pt]
\tblpanel{14}{(b) Semantic routing and tool composition}
Top-10 dynamic router & 14.65 & 2.15 & 93.16 & 12.40 & 18.65 & 1.40 & 3.50 & 4.70 & 66.25 & 205,908 & 10,319 & 216,227 & 29,286 \\
Top-20 dynamic router & 12.75 & 1.65 & 92.98 & 11.30 & 16.15 & 1.35 & 2.60 & 4.05 & 65.88 & 170,972 & 9,038 & 180,010 & 21,696 \\
Workflow-aware router & 15.20 & 1.95 & 93.38 & 12.25 & 21.65 & 1.40 & 4.65 & 3.45 & 56.75 & 210,928 & 9,304 & 220,232 & 26,381 \\
Expert-only suite & 17.70 & 0.60 & 97.59 & 7.35 & 12.60 & 0.00 & 0.00 & 0.00 & 60.88 & 283,502 & 11,988 & 295,490 & 24,192 \\
General + expert suite & 14.05 & 1.55 & 94.43 & 10.40 & 15.40 & 1.40 & 3.05 & 3.45 & 63.00 & 186,216 & 8,875 & 195,092 & 31,437 \\
Atmospheric Python suite & 14.60 & 1.35 & 94.78 & 10.35 & 18.40 & 1.45 & 4.10 & 2.85 & 60.12 & 191,208 & 7,593 & 198,802 & 25,344 \\
Atmospheric Fortran suite & 14.55 & 1.85 & 93.68 & 11.25 & 17.10 & 1.55 & 3.35 & 3.30 & 57.88 & 195,439 & 9,823 & 205,262 & 31,078 \\
\addlinespace[2pt]
\tblpanel{14}{(c) Context retention}
Aggressive compression & 24.60 & 1.95 & 93.88 & 12.50 & 34.90 & 5.40 & 6.25 & 4.00 & 51.62 & 187,492 & 9,545 & 197,038 & 10,765 \\
Short retained context & 20.20 & 2.25 & 93.64 & 13.75 & 26.80 & 2.70 & 6.40 & 5.10 & 56.38 & 198,062 & 11,765 & 209,827 & 14,682 \\
Long retained context & 13.35 & 1.85 & 92.77 & 11.80 & 16.95 & 1.20 & 3.70 & 4.45 & 63.38 & 218,347 & 9,189 & 227,536 & 48,998 \\
Full retained context & 12.15 & 1.30 & 93.84 & 11.50 & 18.00 & 1.05 & 3.90 & 3.40 & 58.25 & 180,517 & 7,182 & 187,700 & 80,819 \\
\addlinespace[2pt]
\tblpanel{14}{(d) Interaction budget}
12-turn budget & 9.95 & 0.30 & 97.36 & 8.50 & 13.05 & 1.15 & 2.70 & 1.30 & 49.50 & 101,826 & 4,437 & 106,263 & 23,002 \\
24-turn budget & 12.90 & 1.65 & 93.11 & 10.80 & 16.90 & 1.35 & 4.20 & 3.30 & 52.88 & 153,261 & 8,143 & 161,404 & 32,346 \\
96-turn budget & 13.60 & 1.30 & 94.12 & 11.90 & 18.45 & 1.60 & 3.85 & 4.55 & 60.12 & 173,925 & 8,873 & 182,798 & 27,763 \\
Unbounded turn budget & 15.20 & 1.80 & 94.22 & 12.30 & 21.40 & 1.60 & 5.15 & 4.35 & 64.50 & 211,751 & 9,429 & 221,180 & 30,131 \\
\bottomrule
\end{tabular}%
}
\end{table}

\subsubsection{Capability-conditioned effects}

The largest capability losses occur when evidence must remain connected across several steps. With 170 visible tools, multi-source performance falls to 47.50 even though most other catalog configurations stay near or above 88. Aggressive compression loses 20.07 points in causal-chain reasoning relative to long retention. The two interventions damage continuity in different places. A large catalog makes the next action unstable; compression makes the meaning of earlier actions unstable. A single calculation can survive either disturbance because its inputs are local. A physical account cannot: its validity depends on remembering why several observations belong together.

\FloatBarrier
\subsection{Oracle, interface, and evidence robustness}
\label{app:oracle-diagnostics}

Table~\ref{tab:bottleneck-robustness} consolidates outcome scores, paired uncertainty, and capability slices for each intervention. Table~\ref{tab:oracle-process} separately reports trajectory and resource effects. Confidence intervals resample paired task-level Core differences. Oracles provide guidance without answers, while robustness conditions alter only package evidence.

\subsubsection{Paired intervention uncertainty}

Only the two evidence-localization oracles have intervals that exclude zero: relevant files [+5.48, +24.69] and the evidence map [+2.34, +21.25]. The stable gain comes from knowing where evidence is, not from receiving a generic plan. A plan names familiar analytical moves; localization resolves an event-specific uncertainty that scientific knowledge alone cannot settle. The other interventions remain suggestive because their intervals cross zero. On a diagnostic manifest, a few packages whose structure favors one interface can still move the mean substantially.

\subsubsection{Trajectory and resource effects}

Oracle guidance shortens search without necessarily reducing cost. The evidence map lowers rounds and file reads, but the injected structure itself consumes context and takes time to interpret. Distractors produce the reverse pattern: the agent performs more actions and records more evidence references with almost no quality loss. Obvious clutter is therefore mostly an efficiency tax. A compact evidence description poses a different problem. The agent must decide whether to trust the summary or reopen the underlying source, and the latter is often necessary to recover metadata, aggregation choices, or conflicts. Search length and evidential control are related, but they are not interchangeable.

\begin{table}[!htbp]
\centering
\caption{\textbf{Intervention trajectories.} Panels match Table~\ref{tab:bottleneck-robustness}; values are per-task means.}
\label{tab:oracle-process}
\tabfontdense
\setlength{\tabcolsep}{0.95pt}
\renewcommand{\arraystretch}{1.03}
\resizebox{\textwidth}{!}{%
\begin{tabular}{@{}l *{12}{r}@{}}
\toprule
& \multicolumn{3}{c}{Execution} & \multicolumn{4}{c}{Evidence and compute} & Latency & \multicolumn{4}{c}{Tokens} \\
\cmidrule(lr){2-4}\cmidrule(lr){5-8}\cmidrule(lr){9-9}\cmidrule(l){10-13}
Intervention & Rounds & \makecell{Tool\\calls} & \makecell{Success\\(\%)} & Reads & Files & \makecell{Evidence\\refs.} & Python & Seconds & Input & Output & Total & Cached \\
\midrule
\tblpanel{13}{(a) Answer-free information oracles}
Relevant-file oracle & 15.30 & 14.30 & 92.17 & 5.15 & 11.30 & 16.10 & 4.15 & 217.26 & 181,939 & 5,572 & 187,512 & 19,507 \\
Evidence-map oracle & 11.10 & 10.10 & 93.28 & 3.10 & 8.55 & 11.10 & 2.75 & 436.56 & 290,368 & 5,050 & 295,418 & 62,950 \\
Tool-subset oracle & 17.80 & 16.80 & 93.18 & 6.60 & 13.35 & 21.60 & 3.90 & 212.74 & 205,252 & 5,628 & 210,880 & 27,738 \\
Planning oracle & 16.95 & 15.95 & 91.49 & 5.45 & 12.65 & 18.30 & 4.05 & 185.18 & 196,489 & 5,817 & 202,306 & 25,190 \\
\addlinespace[2pt]
\tblpanel{13}{(b) Tool-interface presentation}
Hierarchical tool catalog & 16.60 & 15.60 & 91.97 & 5.75 & 12.45 & 17.75 & 3.85 & 197.77 & 182,350 & 6,000 & 188,351 & 20,416 \\
Descriptive tool names & 18.25 & 17.25 & 93.42 & 5.40 & 12.25 & 19.65 & 3.65 & 150.14 & 202,390 & 5,462 & 207,853 & 28,749 \\
Dynamic tool descriptions & 17.55 & 16.55 & 92.12 & 4.00 & 12.85 & 19.95 & 3.65 & 182.42 & 199,673 & 5,036 & 204,709 & 23,053 \\
Opaque tool names & 17.90 & 16.90 & 92.44 & 7.00 & 10.85 & 16.50 & 4.30 & 234.10 & 215,650 & 6,294 & 221,943 & 23,027 \\
\addlinespace[2pt]
\tblpanel{13}{(c) Evidence perturbations}
Equivalent unit change & 16.45 & 15.45 & 90.70 & 5.20 & 13.25 & 19.25 & 3.90 & 176.16 & 184,679 & 5,153 & 189,832 & 19,904 \\
Added distractor files & 19.60 & 18.60 & 89.97 & 6.20 & 13.10 & 22.50 & 4.15 & 162.54 & 220,098 & 5,658 & 225,756 & 25,792 \\
Missing evidence & 17.40 & 16.40 & 89.12 & 5.65 & 12.05 & 21.20 & 3.55 & 181.47 & 198,173 & 5,219 & 203,392 & 19,546 \\
Truncated evidence & 16.10 & 15.10 & 91.65 & 4.25 & 12.75 & 22.10 & 3.55 & 265.79 & 184,244 & 5,827 & 190,071 & 16,781 \\
Conflicting evidence & 16.50 & 15.50 & 92.04 & 5.65 & 12.55 & 19.70 & 3.25 & 155.59 & 182,815 & 5,875 & 188,689 & 18,624 \\
\addlinespace[2pt]
\tblpanel{13}{(d) Optional meteorological access}
Compact meteorology suite & 17.95 & 16.95 & 94.90 & 5.85 & 12.75 & 20.60 & 4.05 & 223.20 & 201,057 & 5,822 & 206,879 & 27,571 \\
Cue-routed meteorology suite & 15.65 & 14.65 & 92.50 & 5.65 & 11.50 & 17.85 & 3.15 & 238.22 & 170,431 & 5,875 & 176,306 & 25,882 \\
\bottomrule
\end{tabular}%
}
\end{table}

\subsubsection{Capability-conditioned intervention effects}

The effective oracles help most with spatiotemporal reconstruction: both exceed 90, compared with 64.40 at baseline. File localization directly removes uncertainty about which records contain the event window and region. Ranking and decision improve less because the oracle does not say how competing quantities should be weighted. Opaque tool names damage the same capability most strongly. Reconstructing a window rarely depends on one call; it requires a sequence of filtering, reading, aggregation, and comparison. Losing the semantic role of any operation makes the resulting timeline harder to recover and harder to audit.

\FloatBarrier
\subsection{Professional meteorological environments}
\label{app:meteo-diagnostics}

The professional-environment study separates method availability from execution. Compliance records whether the required interactions occurred; call success records whether they completed without adapter, argument, or runtime failure.

\subsubsection{Compliance, execution reliability, and implemented scope}

Nearly every run attempts the required specialist interaction, so willingness to call the tool is not the bottleneck. Execution is. Specialist-call success ranges from 20.29\% for PyET to 97.56\% for R climate, and high-success environments usually need fewer repair rounds and smaller contexts. The ranking should not be read as a comparison of languages. It reflects the overlap between the audited operations and the data actually present in the packages. R climate, Julia, and CDO often accept the available time series or grids directly; other adapters require variables, record lengths, or sampling structures that many event packages do not contain.

\begin{table}[H]
\centering
\caption{\textbf{Scientific-environment outcomes and execution.} Final Answer, Process, and Core scores are reported with per-task compliance, successful specialist calls, resource use, and implemented operations.}
\label{tab:meteo-process}
\tabfontdense
\setlength{\tabcolsep}{1.5pt}
\renewcommand{\arraystretch}{1.00}
\begin{adjustbox}{max width=\textwidth,center}
\begin{tabular}{@{}l l *{8}{r}@{}}
\toprule
& & \multicolumn{3}{c}{Outcome} & \multicolumn{2}{c}{Execution} & \multicolumn{3}{c}{Resources} \\
\cmidrule(lr){3-5}\cmidrule(lr){6-7}\cmidrule(l){8-10}
Environment & Operations & Answer & Process & Core & \makecell{Comply\\(\%)} & \makecell{Success\\(\%)} & Rounds & Tokens & \makecell{Latency\\(s)} \\
\midrule
\tblpanel{10}{(a) Python scientific libraries}
MetPy & rain summary; profiles; kinematics & 68.69 & 70.38 & 69.53 & 100 & 38.83 & 25.00 & 311,043 & 255.60 \\
xclim & heat/rain extremes; wet/dry spells & 67.77 & 69.62 & 68.70 & 95 & 38.78 & 24.50 & 299,074 & 268.44 \\
climate-indices & SPI; SPEI; PET & 64.50 & 66.75 & 65.62 & 95 & 22.60 & 30.55 & 374,199 & 306.69 \\
PyET & Penman--Monteith; Hargreaves & 65.34 & 67.38 & 66.36 & 95 & 20.29 & 32.30 & 387,917 & 315.96 \\
thermofeel & UTCI; heat index; wind chill & 66.78 & 68.75 & 67.76 & 100 & 26.85 & 28.55 & 346,924 & 357.88 \\
pyextremes & block maxima; POT; return levels & 59.02 & 60.00 & 59.51 & 100 & 34.19 & 28.05 & 363,839 & 289.16 \\
xskillscore & RMSE; correlation; Brier; CRPS & 63.37 & 64.12 & 63.75 & 95 & 57.14 & 23.70 & 297,724 & 231.47 \\
\addlinespace[2pt]
\tblpanel{10}{(b) Compiled and domain-language environments}
Julia & summary; integration; sensitivity & \best{71.77} & 70.25 & \best{71.01} & 100 & 91.11 & 21.50 & 256,852 & 204.47 \\
R climate & summary; SPEI; GEV; trend & 68.55 & \best{71.38} & 69.96 & 100 & \best{97.56} & 21.70 & 259,984 & 190.20 \\
CDO & temporal and field aggregates & 66.91 & 69.50 & 68.20 & 100 & 81.63 & 22.30 & 268,906 & 197.90 \\
NCO & inspect; weighted mean; derive field & 67.39 & 65.00 & 66.19 & 100 & 35.59 & 26.65 & 321,773 & 230.95 \\
NCL & summary only & 65.03 & 65.00 & 65.01 & 95 & 40.62 & 25.70 & 308,559 & 270.94 \\
Fortran & rain; kinematics; moisture transport & 64.46 & 64.25 & 64.35 & 95 & 43.48 & 25.70 & 301,236 & 222.60 \\
\bottomrule
\end{tabular}
\end{adjustbox}
\end{table}

\paragraph{Applicability and reproducibility.}
Environment names refer to the audited adapters in Table~\ref{tab:meteo-process}, not to every feature of the upstream packages. The outcome columns make the distinction explicit: R climate has the highest call success, while Julia has the highest Core. A successful call proves only that code ran. It does not establish that the estimator fits the record, that the sampling geometry supports the comparison, or that the returned value answers the question. Those checks sit between execution and scientific use, and the gap between execution reliability and final quality shows why they cannot be collapsed into one metric.

Each adapter therefore records normalized arguments, status, output, and errors. It also declares the operations it implements and validates package-scoped inputs before execution. This makes a justified failure reproducible: another reviewer can see that a method was rejected because the record was too short or a required variable was absent. Merely rerunning a command is weaker. Scientific reproducibility requires reconstructing why the operation was admissible and how its output entered the claim.

\FloatBarrier
\section{Prompts and execution contracts}
\label{app:casebook}
\label{app:prompts}

\label{app:casebook-prompts}

The prompts are grouped by role: package-scoped solving and review, answer--process evaluation, and benchmark construction. Placeholders mark values supplied by the harness. The solver sees no demonstration, hidden reference, or recommended file. Appendix~\ref{app:casebook-trajectories} presents representative runs.

\paragraph{Runtime visibility.}
The solver sees the public question, event-package root, exposed tool schemas, and package-relative observations. Ground truth, reference solutions, rubrics, computation programs, judge records, and other model answers remain hidden. The trace retains both successful and failed tool calls. Depending on the condition, the controller may request two format repairs, run a bounded evidence review, require a specialist operation, or reserve the final call for a schema-valid submission. None of these transitions adds scientific evidence. Appendix~\ref{app:inference-config} records their limits.

\begin{promptbox}{\promptkind{CaseBlue}{RUNTIME}\quad Package-scoped research suite}
(*@\promptdivider{CaseBlue}{SOLVER}{EarthVerse interactive solver}@*)
SYSTEM
You are an autonomous extreme-event benchmark solver. You must solve the task
by planning, selecting local files, executing tools, reading observations, and
then producing a structured final answer. Use only local event-package evidence
returned by tools. Do not use web knowledge. Never read computed_gt.json,
solution_en.md, review.json, or compute_gt.py. Prefer deterministic Python
calculations when the task asks for metrics. You must make real tool calls
before finalizing. Every assistant message must follow one of the two response
contracts below.

TASK MESSAGE
Task id: <<TASK_ID>>
Event id: <<EVENT_ID>>
Question path: <<QUESTION_PATH>>
Event package path: <<PACKAGE_ROOT>>

Question:
<<PUBLIC_QUESTION>>

Evidence boundary:
Strict package-only mode: only the current question and event package evidence
are visible.

Tools:
<<REGISTERED_TOOL_CATALOGUE>>

TOOL ACTION CONTRACT
Action        tool_call
Tool          one registered tool name
Arguments     only the arguments required by that tool
Reason        why this call is needed for the current analysis

FINAL ANSWER CONTRACT
Action        final_answer
Answer        the exact structure requested by the public question
Confidence    one numeric confidence value
Evidence      for each claim: package-relative source file and support type
              (direct metric, report claim, metadata, image context, table
              value, derived metric, or unsupported boundary)
Research      a brief ordered summary of the reasoning
Limitations   important unsupported or unresolved elements

Start by inspecting the package inventory, then choose only the files needed
for this event-specific disaster analysis.

(*@\promptdivider{CaseBlue}{REVIEW}{Final evidence review}@*)
Begin a final evidence-review phase before submitting the answer. Treat the
prior answer as a draft: re-check its evidence provenance, formulas, units,
intermediate values, competing physical explanations, and every required
output field against the package. Use additional tools when verification is
useful. Return either one tool action or a revised final answer; do not merely
endorse the draft. You have at most <<MAX_REVIEW_TURNS>> review turns.
\end{promptbox}

\begin{promptbox}[colback=CasePaleTeal,colframe=CaseTeal,colbacktitle=CaseTeal]{\promptkind{CaseTeal}{EVALUATION}\quad Joint answer and process judge}
(*@\promptdivider{CaseTeal}{JUDGE}{Answer and research-process evaluation}@*)
SYSTEM
You are a strict but fair evaluator for a local extreme-event benchmark. Your
job is to judge answer correctness and task-rubric quality. Use partial 0-100
scores and avoid old deterministic exact matching.

CALIBRATION CORE
Evaluate this model response for an extreme-event evidence-and-reasoning
benchmark. Make one combined evaluation and produce two core 0-100 scores.

- answer_correctness_score is the direct partial-credit comparison between the
  model's final answer and the ground truth. Treat the answer as a set of units:
  required fields, labels, decisions, values, units, orderings, alternatives,
  stage boundaries, and required conclusions. Missing required fields are
  incorrect; correct prose cannot replace a missing structured field.
- llm_rubric_score must come from the task's own 20-point rubric. Compare the
  answer and visible trace against every criterion, assign earned points out of
  20, and convert the result to percent.
- Do not invent extra credit. A plausible narrative, many tool calls, or reading
  the right files is not by itself a correct solution process. Evidence,
  intermediate values, transformations, and conclusions must agree.
- Keep the two scores distinct. A correct-looking final answer cannot substitute
  for missing process evidence, and an attempted method cannot receive full
  rubric credit when its result is wrong.
- Use only the supplied package evidence, ground truth, task solution, rubric,
  final answer, and model trace. Do not rely on web knowledge.

Runtime payload:
<<TASK_METADATA>>
<<QUESTION>>
<<GROUND_TRUTH>>
<<SOLUTION_AND_20_POINT_RUBRIC>>
<<MODEL_FINAL_ANSWER>>
<<MODEL_TOOL_TRACE>>
<<MODEL_RUN_NOTES>>

EVALUATION RECORD
Scores        answer correctness and rubric-based process quality
Answer units  each required unit and whether it was satisfied
Rubric ledger criterion-level points earned out of 20
Diagnosis     verdict, critical errors, missing elements, unsupported claims
Rationale     one sentence explaining the score pair
\end{promptbox}

\begin{promptbox}[colback=CasePalePurple,colframe=CasePurple,colbacktitle=CasePurple]{\promptkind{CasePurple}{CONSTRUCTION}\quad Benchmark construction suite}
(*@\promptdivider{CasePurple}{QUALITY}{Task construction acceptance review}@*)
ROLE
You are the final quality reviewer for one extreme-event benchmark task.

INPUTS
- The public question (question_en.md)
- The expert solution (solution_en.md)
- The executable ground-truth program (compute_gt.py)
- The generated scoring contract (computed_gt.json)
- The optional review record (review.json)

TASK
Verify that the five artifacts describe one event-specific scientific target.
Do not approve a rewritten public question when the solution or GT still scores
an older package-inspection objective.

ACCEPTANCE CHECKS
1. The public question reads like a real disaster-analysis assignment.
2. It contains no local path, visible CSX identifier, package inventory,
   available-material wording, AOI, quality-gate, evidence-boundary, or benchmark
   phrasing, and it does not reveal the useful files.
3. The solution contains Final Answer, Key Computations, Reasoning Path,
   Computed Interpretation, and Scoring Rubric.
4. computed_gt.json declares schema_version 1.0, task and event identifiers,
   task type, evaluation mode, primary GT, key values, and a top-level rubric.
5. Deterministic tasks contain compact machine-checkable targets. Open expert
   tasks contain concrete judge anchors rather than a fabricated exact essay.
6. The rubric has 5-8 criteria whose points sum exactly to 20.
7. Former distractors are rejected through numeric tests, inequalities, process
   timing, or competing-mechanism checks rather than option prose.
8. compute_gt.py runs successfully and regenerates the scoring contract.

OUTPUT
Return changed or reviewed file paths, event and hazard type, task type,
evaluation mode, acceptance status for every check, and any remaining ambiguity,
missing-data risk, temporal/spatial mismatch, or unsupported overclaim.

(*@\promptdivider{CasePurple}{RUBRIC}{Event-specific rubric construction}@*)
ROLE
You are constructing the hidden scoring contract for one benchmark task.

TASK
Create a 20-point rubric that can be applied without guessing the author's
intent. Use 5-8 small, independently judgeable criteria. Every criterion must
contain name, points, description, and partial_credit. The point values must sum
exactly to 20.

REQUIRED COVERAGE
- 3-4 points: final conclusion, numeric target, threshold state, proof result,
  ranking, or concise canonical stance.
- 3-5 points: required quantitative extraction or calculation, including units,
  tolerances, and windows where relevant.
- 2-4 points: formula, index, process-window, causal-chain, or score-ledger logic.
- 2-4 points: fusion of reports, time series, rasters, imagery, exposure,
  infrastructure, or trajectory evidence into checkable anchors.
- 1-3 points: a compact event-specific consequence that follows from the
  computed evidence.
- 1-3 points: rejection of plausible competing mechanisms or overclaims.
- 1-2 points: complete, concise output in the requested structure.

HARD RULES
Do not use vague criteria such as "good reasoning" or "uses evidence well."
Do not award points for choosing an option after a converted task has become a
numeric or structured analysis. A criterion that depends on an incorrect
scientific result cannot receive full process credit. For llm_judge tasks,
include canonical_stance, required_claims, required_metrics, forbidden_claims,
and acceptable_variants in primary_gt. For hybrid tasks, preserve both the
deterministic core and the reasoning anchors.

RUBRIC RECORD
Total points   exactly 20
Criteria       5-8 independently judgeable items
Each item      event-specific name; points; full-credit requirement;
               partial-credit and error rule

(*@\promptdivider{CasePurple}{SOURCE}{Event authenticity and source audit}@*)
ROLE
You are auditing the real-event anchor of one event package.

QUESTIONS
1. Did the event occur? Require at least one traceable source confirming the
   event name, time window, or principal affected region.
2. Is the source reliable? Prefer government agencies, international
   organizations, peer-reviewed work, and authoritative scientific institutions;
   then disaster catalogues and major news. Encyclopedias and generic search
   pages are supporting evidence only.
3. Does the source actually correspond to this event? Check event name, time,
   place, and hazard type. A source about the same hazard family but a different
   time or place is not a strong match.

OUTPUT FIELDS
- existence_verdict: confirmed | likely | weak | not_confirmed
- temporal_match: exact | near | broad | mismatch | unknown
- spatial_match: exact | regional | broad | mismatch | unknown
- hazard_match: exact | compound_related | broad_family | mismatch | unknown
- source_reliability: high | medium | low | mixed
- release_decision: keep | keep_with_minor_fix | needs_stronger_source |
  replace_event
- concise evidence and repair notes

DECISION RULES
Use keep when the event is confirmed, at least two of temporal/spatial/hazard
matching are exact or regional, and the preferred evidence is authoritative.
Use keep_with_minor_fix when the event is real but URL, naming, or scale needs a
small correction. Use needs_stronger_source when current support is generic,
secondary, or overly broad. Use replace_event when the event cannot be confirmed
or the cited evidence clearly belongs to another event.
\end{promptbox}

\FloatBarrier
\section{Representative case set}
\label{app:casebook-overview}

The six cards below index remote-sensing reconciliation, urban pluvial
causality, evidence substitution, multi-agent quantitative decomposition,
integrated heat research, and a partial scientific failure. Together they cover
five system families without repeating the full records that follow.

\begin{overviewbox}
\summarycard{CaseBlue}{FRONTIER HOSTED}{GPT-5.6 Sol}{CSX-189\_Q1}
{\scorechip{99.50}}{Uljin wildfire consistency.}
{Report claims, wind and precipitation, Sentinel-2 dNBR, annual change, smoke
transport, and bounded impact claims.}
{Sixteen successful calls, including a failed semantic search recovered by a
filename query and a final provenance check.}

\summarycard{CaseTeal}{GENERAL HOSTED}{Tencent Hy3}{CSX-059\_Q2}
{\scorechip{100.00}}{NYC Ida pluvial overload.}
{An hourly rainfall record inside a two-day rain shield, population load,
surface change, and a controlled future perturbation.}
{Ten evidence calls followed by one deterministic calculation of every ratio,
normalization term, and scenario score.}

\summarycard{CaseAmber}{OPEN-WEIGHT LARGE}{Qwen3-235B}{CSX-006\_Q2}
{\scorechip{93.93}}{Heat-source substitution audit.}
{Variable-level admissibility of an ERA5-Land aggregate substituted for local
hourly apparent-heat and warm-night data.}
{Eight calls covering inventory, source comparison, schema checks, corrected
heat metrics, and numerical source impact.}

\summarycard{CasePurple}{MULTI-AGENT FRAMEWORK}{GeoMMAgent}{CSX-063\_Q1}
{\scorechip{83.50}}{Zhengzhou threshold ledger.}
{Reported burst and multiday rainfall, product ordering, exposure thresholds,
and a conjunctive flood diagnosis.}
{Four coordinator roles and nine package-tool calls for retrieval, calculation,
scientific synthesis, and schema assembly.}

\summarycard{CaseGreen}{RESEARCH FRAMEWORK}{OpenResearcher}{CSX-018\_Q2}
{\scorechip{86.92}}{Sahel heat-response model.}
{Hourly heat, wet bulb, warm nights, exposure, services, roads, surface
stability, two indices, and a +1 C counterfactual.}
{Twenty-one calls, including one failed heat script, an explicit repair, and
independent checks of peaks, duration, density, scores, and deltas.}

\summarycard{CaseRose}{GENERAL HOSTED}{GPT-4o}{CSX-312\_Q2}
{\failchip{Core 41.25}}{Partial scientific failure.}
{The model identifies the coupled Kilauea mechanism but reports lava effusion
rate instead of the required summit-collapse-to-flow volume ratio.}
{Nineteen model rounds and nine tool calls: seven succeed, two guessed paths
fail, and the final answer earns Answer 40.00 and Rubric 42.50.}
\end{overviewbox}

\FloatBarrier
\section{Detailed research trajectories}
\label{app:casebook-trajectories}

Appendix~\ref{app:casebook-overview} identifies the six selected systems and
tasks. This section expands each card into its scientific target, evidence
sequence, calculations, and decision. The Uljin case retains the complete task
and action order; the remaining records focus on the steps that determine the
result.

\caseheading{CaseBlue}{Full case: Uljin report-to-burn consistency}

\begin{questionbox}{CSX-189\_Q1}
(*@\questionrole{CaseBlue}{SCIENTIFIC TARGET}@*)
(*@\questionvalue{CaseBlue}{Uljin wildfire report-to-burn consistency ledger}@*)

A technical review team is checking whether the March 4-13, 2022 Uljin wildfire
is numerically consistent with a dry, wind-assisted burn-scar diagnosis rather
than a late-rain-control or image-baseline alternative.

(*@\questionrole{CaseTeal}{EVIDENCE SCOPE}@*)
Use only package-local evidence. Do not use point-weather products for this task;
build the ledger from the event report, package gridded precipitation and wind
summaries, Sentinel-2 dNBR, and annual embedding-change context.

(*@\questionrole{CaseAmber}{REQUIRED COMPUTATION}@*)
Compute these ledger values:

- report_dry_wind_flag: 1 if the report links the event to strong winds and dry
  weather, otherwise 0.
- smoke_transport_flag: 1 if the report states that smoke moved toward southern
  Japan, otherwise 0.
- era5_mean_wind_speed_mps: vector speed from mean 10 m u and v components.
- era5_wind_bearing_to_deg: direction toward which the mean wind vector points,
  degrees clockwise from north.
- precip_mean_spread_mm: spread between ERA5-Land, GPM IMERG, and CHIRPS event
  precipitation means.
- dnbr_mean: mean dNBR over the event burn-change window.
- dnbr_max_to_mean_ratio: dNBR maximum divided by dNBR mean.
- dnbr_to_annual_change_mean_ratio: dNBR mean divided by annual 1-minus-cosine
  embedding-change mean.
- reported_charred_area_km2: reported hectares converted to square kilometers.

(*@\questionrole{CasePurple}{MECHANISM CHECKS}@*) Apply these tests:
- report_dry_wind_mechanism_test: pass if report_dry_wind_flag = 1.
- smoke_transport_context_test: pass if smoke_transport_flag = 1.
- positive_heterogeneous_burn_test: pass if dnbr_mean > 0 and
  dnbr_max_to_mean_ratio >= 2.0.
- late_rain_control_test: pass only if rainfall is the dominant control;
  otherwise fail.
- nuclear_damage_overclaim_test: pass if the answer avoids inferring nuclear-
  plant damage or radiation release.

(*@\questionrole{CaseRose}{OUTPUT}@*)
Evidence ledger
- Report/dry-wind flag and smoke-transport flag
- Mean ERA5 wind speed and bearing
- Precipitation-mean spread across the three gridded products
- Mean dNBR, dNBR maximum-to-mean ratio, and dNBR/annual-change ratio
- Reported charred area in square kilometres

Decision checks
- Report mechanism; smoke context; positive heterogeneous burn
- Late-rain control; bounded nuclear-impact claim

(*@\questionrole{CaseGreen}{INTERPRETATION}@*)
- One or two sentences tied directly to the computed ledger
\end{questionbox}

\begin{casepanel}{CaseBlue}{Research record: CSX-189\_Q1}
\panelhead{CaseBlue}{Complete action order; observation bodies abridged}
\tracestep[CaseBlue]{1}{list_package_files}{Discover}{Enumerated all 39 package files.}
\tracestep[CaseTeal]{2}{read_text_file}{Anchor}{Confirmed Uljin, South Korea, 4--13 March 2022.}
\tracestep[CaseBlue]{3}{search_package_files}{Mechanism}{Located report language linking strong winds and dry weather.}
\tracestep[CaseBlue]{4}{search_package_files}{Transport}{Located the statement that smoke moved toward southern Japan.}
\tracestep[CaseBlue]{5}{search_package_files}{Impact quantity}{Recovered nearly 17,000 charred hectares.}
\tracestep[CaseTeal]{6}{read_text_file}{ERA5-Land}{Read mean precipitation and mean 10 m wind components.}
\tracestep[CaseTeal]{7}{read_text_file}{GPM IMERG}{Read event precipitation mean of 2.884 mm.}
\tracestep[CaseTeal]{8}{read_text_file}{CHIRPS}{Read event precipitation mean of 4.456 mm.}
\tracestep[CaseTeal]{9}{read_text_file}{Sentinel-2 dNBR}{Read mean 0.303255 and maximum 0.708527.}
\tracestep[CaseBlue]{10}{search_package_files}{Embedding search}{Initial semantic query returned no match.}
\tracestep[CaseBlue]{11}{search_package_files}{Recovery}{A filename-oriented query located annual cosine-change statistics.}
\tracestep[CaseTeal]{12}{read_text_file}{Annual context}{Read mean 1-minus-cosine change of 0.013761.}
\tracestep[CaseBlue]{13}{search_package_files}{Smoke check}{Confirmed the report-side smoke passage.}
\tracestep[CaseTeal]{14}{read_text_file}{Full report}{Read the authoritative HTML surrounding all extracted claims.}
\tracestep[CasePurple]{15}{python_exec}{Deterministic ledger}{Computed wind speed 1.392 m/s, bearing 331.253 degrees, precipitation spread 6.389 mm, dNBR heterogeneity 2.336, dNBR/annual-change ratio 22.037, and 170 km2 charred area.}
\tracestep[CaseBlue]{16}{search_package_files}{Final provenance check}{Re-queried the most easily misquoted smoke-transport claim before finalization.}
\panelhead{CaseBlue}{Computed ledger}
\noindent
\metric{DRY-WIND FLAG}{1}\quad
\metric{SMOKE FLAG}{1}\quad
\metric{MEAN WIND}{1.392043 m/s}\quad
\metric{BEARING}{331.252581 deg}\par
\metric{PRECIPITATION SPREAD}{6.389107 mm}\quad
\metric{MEAN dNBR}{0.303255}\quad
\metric{dNBR MAX/MEAN}{2.336407}\par
\metric{dNBR/ANNUAL CHANGE}{22.037202}\quad
\metric{CHARRED AREA}{170.0 km$^2$}\par
\smallskip
\verdict{PASS}\enspace Report mechanism\quad
\verdict{PASS}\enspace Smoke context\quad
\verdict{PASS}\enspace Heterogeneous burn\par
\verdict[CaseRose]{FAIL}\enspace Late-rain control\quad
\verdict{PASS}\enspace Bounded nuclear-impact claim

\panelhead{CaseBlue}{Evidence reconciliation}
The burn scar is not accepted on appearance alone. The trace anchors two report
claims, reconciles three precipitation products, and keeps event-window dNBR
separate from annual embedding change. Step 10 also preserves the failed
semantic search and the filename-based recovery that follows.
\end{casepanel}

\begin{casepanel}{CaseTeal}{NYC Ida: urban pluvial overload}
\panelhead{CaseTeal}{Task}
\questionitem{CaseBlue}{SCIENTIFIC TARGET}{Reconstruct the New York City phase
of post-tropical Cyclone Ida as a record hourly burst embedded in a broader
two-day rain shield.}
\questionitem{CaseTeal}{EVIDENCE SCOPE}{Keep station-scale hourly rainfall
distinct from the wider gridded rainfall field, then add population exposure
and surface-change context.}
\questionitem{CaseAmber}{REQUIRED COMPUTATION}{Compute the baseline response
stress and repeat it under the prescribed future perturbation.}

\panelhead{CaseTeal}{Research trajectory: ten evidence calls and one calculation}
\tracestep[CaseBlue]{1}{list + anchor}{Window}{Enumerated 39 files and fixed the inclusive 1--2 September 2021 event window.}
\tracestep[CaseTeal]{2}{read report}{Local forcing}{Recovered the Central Park hourly record and the conservative 10-inch regional reference.}
\tracestep[CaseTeal]{3}{read 3 products}{Rain shield}{Read GPM IMERG, CHIRPS, and ERA5-Land event means and maxima.}
\tracestep[CaseTeal]{4}{read exposure}{Context}{Read 5.013 million exposed people, Sentinel-1 mean change of 0.234 dB, and annual embedding change of 0.022.}
\tracestep[CasePurple]{5}{python_exec}{Ledger + scenario}{Converted units, reconciled products, computed ratios and normalized terms, and reran the weighted index under the prescribed perturbation.}

\panelhead{CaseTeal}{Evidence, metrics, and decision}
\noindent
\metric{WINDOW}{2 days}\quad
\metric{RECORD HOUR}{88.138 mm}\quad
\metric{REGIONAL REFERENCE}{254.000 mm}\quad
\metric{GRID PEAK}{136.888 mm}\par
\metric{GRID MEAN}{53.427 mm}\quad
\metric{HOUR/PEAK}{0.644}\quad
\metric{REGIONAL/PEAK}{1.856}\quad
\metric{BURST/AREAL DAY}{3.299}\par
\metric{POPULATION}{5.013 million}\quad
\metric{RAIN LOAD}{267.812 million-person-mm}\quad
\metric{SURFACE CHANGE}{0.337}\par
\medskip
\verdict{DECISION}\enspace The baseline response-stress index is
\textbf{72.6}; it rises to \textbf{80.1} under the future perturbation
(\textbf{+7.5}).  The local hourly record is not treated as a substitute for
the wider gridded rain shield.
\end{casepanel}

\begin{casepanel}{CaseAmber}{Heat-source substitution audit}
\panelhead{CaseAmber}{Task}
\questionitem{CaseBlue}{SCIENTIFIC TARGET}{Determine whether a same-window
ERA5-Land aggregate can replace the local hourly heat source.}
\questionitem{CaseTeal}{EVIDENCE SCOPE}{Audit variable availability for
apparent-temperature duration and warm-night exposure rather than judging a
source by filename.}
\questionitem{CaseGreen}{DECISION RULE}{Accept the substitution only when the
replacement preserves every variable needed for the requested metrics.}

\panelhead{CaseAmber}{Research trajectory: eight successful calls}
\tracestep[CaseBlue]{1}{list_package_files}{Discover}{Enumerated 37 files and identified the two competing heat products.}
\tracestep[CaseTeal]{2}{read anchor}{Window}{Fixed 31 March--4 April 2024.}
\tracestep[CaseTeal]{3}{read sources}{Compare}{The local source contains hourly temperature, humidity, apparent temperature, and daily minima; the aggregate lacks the last two required roles.}
\tracestep[CasePurple]{4}{python_exec}{Schema audit}{Confirmed that apparent temperature is the first missing field and daily minimum temperature is also absent.}
\tracestep[CasePurple]{5}{python_exec}{Correct + compare}{Recomputed the target heat metrics and measured the peak-temperature error introduced by the substitute.}

\panelhead{CaseAmber}{Evidence, metrics, and decision}
\noindent
\metric{MAX AIR}{44.6 C}\quad
\metric{MAX APPARENT}{43.2 C}\quad
\metric{HOURS APPARENT $\geq 40$ C}{16}\quad
\metric{WARM NIGHTS $\geq 27$ C}{5}\par
\metric{AGGREGATE MAX}{35.3 C}\quad
\metric{PEAK DIFFERENCE}{9.3 C}\par
\metric{MISSING FIELDS}{apparent temperature; daily minimum}\par
\medskip
\verdict[CaseRose]{SOURCE SWAP FAILS}\enspace The aggregate cannot compute
apparent-heat duration or warm nights.  The corrected label is
\texttt{sustained\_apparent\_heat\_with\_warm\_night\_exposure}.
\end{casepanel}

\begin{casepanel}{CasePurple}{Zhengzhou: coordinated threshold ledger}
\panelhead{CasePurple}{Task}
\questionitem{CaseBlue}{SCIENTIFIC TARGET}{Test whether the 17--23 July 2021
Zhengzhou flood combined an exceptional hourly burst, multiday extreme
rainfall, and dense urban exposure.}
\questionitem{CaseAmber}{REQUIRED COMPUTATION}{Build the rainfall-ratio,
product-order, population, and infrastructure threshold ledger.}
\questionitem{CaseGreen}{DECISION RULE}{Assign the final label only when all
four numerical thresholds pass.}

\panelhead{CasePurple}{Research trajectory: four roles and nine package-tool calls}
\tracestep[CaseBlue]{1}{Agent 0: retrieve}{Evidence}{Read the event report, three precipitation products, WorldPop, and OSM with strict provenance.}
\tracestep[CasePurple]{2}{Agent 1: compute}{Ledger}{Calculated rainfall ratios, product order, and the four threshold flags.}
\tracestep[CaseAmber]{3}{Agent 2: synthesize}{Mechanism}{Joined physical rainfall evidence with population and road exposure.}
\tracestep[CaseGreen]{4}{Agent 3: assemble}{Output}{Built the requested schema without changing the shared evidence boundary.}
\tracestep[CasePurple]{5}{python_exec}{Verify}{Recovered hour share 0.3272, annual ratio 0.963, and four true threshold flags.}

\panelhead{CasePurple}{Evidence, metrics, and decision}
\noindent
\metric{THREE-DAY RAIN}{617.1 mm}\quad
\metric{ONE-HOUR RAIN}{201.9 mm}\quad
\metric{HOUR SHARE}{0.3272}\quad
\metric{ANNUAL RATIO}{0.963}\par
\metric{MEAN ORDER}{GPM $>$ CHIRPS $>$ ERA5-Land}\quad
\metric{MAX GRID MEAN}{460.5514 mm}\quad
\metric{MAX GRID}{484.255 mm}\par
\metric{POPULATION}{5.2625 million}\quad
\metric{HIGHWAYS}{618}\quad
\metric{AMENITIES}{50}\quad
\metric{TUNNELS}{2}\par
\medskip
\verdict{ALL FOUR THRESHOLDS PASS}\enspace Hour share $\geq 0.30$, annual
ratio $\geq 0.90$, population $\geq 5$ million, and highway features
$\geq 200$.  Final label:
\texttt{short\_burst\_multiday\_urban\_exposure}.
\end{casepanel}

\begin{casepanel}{CaseGreen}{Sahel heat: response load and warming scenario}
\panelhead{CaseGreen}{Task}
\questionitem{CaseBlue}{SCIENTIFIC TARGET}{Model a five-day Sahel heat episode
and its response burden.}
\questionitem{CaseTeal}{EVIDENCE SCOPE}{Combine hourly heat and Stull wet-bulb
temperature, warm nights, population, services, roads, gridded context, and
surface stability.}
\questionitem{CaseAmber}{REQUIRED COMPUTATION}{Calculate both response indices
at baseline and under a +1 C counterfactual.}

\panelhead{CaseGreen}{Research trajectory: twenty-one calls}
\tracestep[CaseBlue]{1}{list + open}{Core inputs}{Read hourly heat, WorldPop, AOI geometry, and OSM infrastructure.}
\tracestep[CasePurple]{2}{python: exposure}{Counts}{Computed schools, health facilities, shelters, major roads, AOI area, and population density.}
\tracestep[CaseRose]{3}{python: heat}{Failed attempt}{The first heat script failed because it assumed the wrong hourly field names.}
\tracestep[CasePurple]{4}{python: repair}{Schema recovery}{Re-read the actual fields and computed Stull wet-bulb temperature and warm nights.}
\tracestep[CaseTeal]{5}{open products}{Context}{Read ERA5-Land temperature, GPM and CHIRPS precipitation, and Sentinel-2 dNBR.}
\tracestep[CasePurple]{6}{python checks}{Heat metrics}{Independently verified air and apparent peaks, duration, degree-hours, and density.}
\tracestep[CasePurple]{7}{python scenario}{Indices}{Rechecked baseline and +1 C scores, threshold counts, and deltas.}

\panelhead{CaseGreen}{Heat, exposure, and surface context}
\noindent
\metric{MAX AIR}{44.6 C}\quad
\metric{MAX APPARENT}{43.2 C}\quad
\metric{MAX WET BULB}{16.3605 C}\quad
\metric{HOURS $\geq 40$ C}{16}\quad
\metric{DEGREE-HOURS}{24.4}\quad
\metric{WARM NIGHTS}{5}\par
\metric{POPULATION}{170,311.46}\quad
\metric{DENSITY}{3.46 km$^{-2}$}\quad
\metric{SCHOOLS}{6}\quad
\metric{HEALTH FACILITIES}{2}\quad
\metric{SHELTERS}{1}\quad
\metric{MAJOR ROADS}{57}\par
\metric{SURFACE CHANGE}{0.024472}\quad
\metric{SURFACE STABILITY}{0.97553}

\panelhead{CaseTeal}{Indices and +1 C scenario}
\noindent
\metric{BASE HEAT RESPONSE}{77.344912}\quad
\metric{BASE PRIORITY}{86.374648}\par
\metric{FUTURE MAX APPARENT}{44.2 C}\quad
\metric{FUTURE HOURS $\geq 40$ C}{25}\quad
\metric{FUTURE DEGREE-HOURS}{40.4}\par
\metric{FUTURE HEAT RESPONSE}{84.444912}\quad
\metric{FUTURE PRIORITY}{93.094648}\quad
\metric{DELTAS}{+7.1; +6.72}\par
\medskip
\verdict{INTERPRETATION}\enspace The explicit failed computation and repair
remain in the record; the final scenario is supported by independently
rechecked heat, exposure, and response terms.
\end{casepanel}

\label{app:casebook-failure}

\begin{casepanel}{CaseRose}{Partial failure: correct mechanism, wrong physical index}
\panelhead{CaseRose}{Task}
\questionitem{CaseBlue}{SCIENTIFIC TARGET}{Diagnose the 2018 Kilauea eruption
as isolated lower-rift, isolated summit, rainfall-triggered, or coupled
summit-drainage/lower-rift activity.}
\questionitem{CasePurple}{MECHANISM CHECK}{Use the evidence to distinguish the
coupled explanation from all three alternatives.}
\questionitem{CaseRose}{OUTPUT}{Return one physical index and value, three
evidence clues, and an operational implication.}

\smallskip
\noindent
\metric{MODEL ROUNDS}{19}\quad
\metric{TOOL CALLS}{9}\quad
\metric{SUCCESS}{7}\quad
\metric{ERRORS}{2}\quad
\metric{UNIQUE EVIDENCE FILES}{4}

\panelhead{CaseRose}{Observed research sequence}
\tracestep[CaseBlue]{1}{list_package_files}{Discover}{Enumerated all 34 package files.}
\tracestep[CaseTeal]{2}{read USGS report}{Primary evidence}{Opened the authoritative summit-collapse and Lower East Rift Zone report.}
\tracestep[CaseAmber]{3}{search package}{Weak recovery}{Two semantic searches returned no direct match for the requested coupling language.}
\tracestep[CaseTeal]{4}{read anchor}{Event bounds}{Confirmed the 3 May--4 August 2018 event window and Kilauea location.}
\tracestep[CaseRose]{5}{read guessed paths}{Tool errors}{Two nonexistent guessed files produced the run's only tool errors.}
\tracestep[CasePurple]{6}{search + reread}{Final extraction}{Recovered downrift propagation and effusion above 100 m$^3$/s, then finalized without computing the 0.8/0.8 volume ratio.}

\panelhead{CaseRose}{Returned answer}
\noindent
\metric{DIAGNOSIS}{coupled summit drainage and lower-rift effusion}\par
\metric{PHYSICAL INDEX}{Lava Effusion Rate}\quad
\metric{VALUE}{100 m$^3$/s}\par
\textcolor{CaseMuted}{\sffamily\scriptsize\bfseries CLUES}\enspace
Downrift magma propagation after summit collapse; lava discharge above
100 m$^3$/s; coordinated summit and rift monitoring.

\panelhead{CaseGreen}{Required answer}
\noindent
\metric{DIAGNOSIS}{coupled summit drainage and LERZ effusion}\par
\metric{PHYSICAL INDEX}{summit-collapse-to-lava-flow volume ratio}\quad
\metric{VALUE}{1.0}\par
The index follows from approximately \textbf{0.8 km$^3$} of summit collapse and
\textbf{0.8 km$^3$} of lava flow.  A complete answer also uses partial summit
drainage, near-daily collapses, and $M_w$ 4.7--5.4 energy release, then rejects
isolated-rift, isolated-summit, and rainfall-triggered alternatives.

\panelhead{CaseRose}{Assessment}
\noindent
\metric{ANSWER}{40.00}\quad
\metric{RUBRIC}{42.50}\quad
\metric{CORE}{41.25}\par
\medskip
\verdict[CaseRose]{PARTIAL FAILURE}\enspace The model identifies the correct
coupled mechanism and gives a useful operational implication, but substitutes
a real yet irrelevant rate for the required volume ratio.  It also omits the
summit evidence and explicit alternative-mechanism rejection.  The failure is
scientific and evidential, not merely a formatting breakdown.
\end{casepanel}

\end{document}